%% file: main.tex
\documentclass[journal]{IEEEtran}

\usepackage{fancyhdr}
\usepackage[normalem]{ulem}
\usepackage[hyphens]{url}
\usepackage{hyperref}
\usepackage{microtype}
\usepackage{amsmath}
\usepackage{graphicx}
\usepackage{epsfig}
\usepackage{subfigure}
\usepackage{multirow}
\usepackage{array}
\usepackage{enumitem}
\usepackage{cleveref}
\usepackage[numbers,sort&compress]{natbib}
\usepackage{tablefootnote}
\usepackage{url}
\usepackage{cleveref}
\usepackage[usenames,dvipsnames]{xcolor}
\let\oldmarginpar\marginpar
\renewcommand\marginpar[1]{\-\oldmarginpar[\raggedleft\footnotesize #1]%
{\raggedright\footnotesize #1}}

\begin{document}
%\bstctlcite{IEEEexample:BSTcontrol}

% JSE - I changed 'Efficient' to 'Accelerated', since I think that better characterizes what the article is about
\title{Efficient Processing of Deep Neural Networks: \\A Tutorial and Survey}

\author{\IEEEauthorblockN{Vivienne Sze,~\IEEEmembership{Senior Member,~IEEE,} Yu-Hsin Chen,~\IEEEmembership{Student Member,~IEEE,} Tien-Ju Yang,~\IEEEmembership{Student Member,~IEEE,} Joel Emer,~\IEEEmembership{Fellow,~IEEE}}
\thanks{V. Sze, Y.-H. Chen and T.-J. Yang are with the Department of Electrical Engineering and Computer Science, Massachusetts Institute of Technology, Cambridge, MA 02139 USA. (e-mail:
sze@mit.edu; yhchen@mit.edu, tjy@mit.edu)}
\thanks{J. S. Emer is with the Department of Electrical Engineering and Computer Science, Massachusetts Institute of Technology, Cambridge, MA 02139 USA, and also with Nvidia Corporation, Westford, MA 01886 USA. (e-mail:
jsemer@mit.edu)}% <-this % stops a space
}


%Telephone: (800) 555--1212, Fax: (888) 555--1212}


\maketitle

\input{abstract}


% \fi
%\IEEEpeerreviewmaketitle

\input{introduction}
\input{background}

\input{training}
\input{development_history}
\input{applications}

\input{embedded}
\input{overview}
\input{resources}
%\input{challenges}
\input{architecture}
\input{technology}
\input{algorithms}
\input{benchmarking}

\input{summary}



%\bibliographystyle{IEEEtran}

%\bibliographystyle{ieeetr}

\small
\bibliographystyle{IEEEtran}
\bibliography{IEEEabrv,references}

\end{document}

%% file: abstract.tex
\begin{abstract}
Deep neural networks (DNNs) are currently widely used for many artificial intelligence (AI) applications including computer vision, speech recognition, and robotics. While DNNs deliver state-of-the-art accuracy on many AI tasks, it comes at the cost of high computational complexity. Accordingly, techniques that enable efficient processing of DNNs to improve \emph{energy efficiency} and \emph{throughput} without sacrificing application accuracy or increasing hardware cost are critical to the wide deployment of DNNs in AI systems. 

This article aims to provide a comprehensive tutorial and survey about the recent advances towards the goal of enabling efficient processing of DNNs. Specifically, it will provide an overview of DNNs, discuss various hardware platforms and architectures that support DNNs, and highlight key trends in reducing the computation cost of DNNs either solely via hardware design changes or via joint hardware design and DNN algorithm changes. It will also summarize various development resources that enable researchers and practitioners to quickly get started in this field, and highlight important benchmarking metrics and design considerations that should be used for evaluating the rapidly growing number of DNN hardware designs, optionally including algorithmic co-designs, being proposed in academia and industry.

The reader will take away the following concepts from this article: understand the key design considerations for DNNs; be able to evaluate different DNN hardware implementations with benchmarks and comparison metrics; understand the trade-offs between various hardware architectures and platforms; be able to evaluate the utility of various DNN design techniques for efficient processing; and understand recent implementation trends and opportunities.
\end{abstract}

%% file: introduction.tex
\section{Introduction}
%This is the era of \emph{big data}.  More data has been created in the past two years than the entire history of the human race~\cite{forbes}.  This is primarily driven by the exponential increase in the use of sensors (10 billion per year in 2013, expected to reach 1 trillion by 2020~\cite{sensors}) and connected devices (6.4 billion in 2016, expected to reach 20.8 billion by 2020~\cite{connected}). These sensors and devices generate hundreds of zetabytes ($10^{21}$ bytes) of data per year ---  petabytes ($10^{15}$ bytes) per second~\cite{ciscoGCI}.

%\emph{Machine learning} is needed to extract meaningful, and ideally actionable, information from this data.  A significant amount of computation is required in order to analyze this data, which often happens in the cloud. However, given the sheer volume and rate at which data is being generated, and the high energy cost of communication and often limited bandwidth, there is an increasing need to perform the analysis locally near the sensor rather than sending the raw data to the cloud. Embedding machine learning at the edge also addresses important concerns related to privacy, latency and security. 

Deep neural networks (DNNs) are currently the foundation for many modern artificial intelligence (AI) applications~\cite{nature2015-lecun}. Since the breakthrough application of DNNs to speech recognition~\cite{deng2013recent} and image recognition~\cite{nips2012-krizhevsky}, the number of applications that use DNNs has exploded. These DNNs are employed in a myriad of applications from self-driving cars~\cite{chen2015deepdriving}, to detecting cancer~\cite{esteva2017dermatologist} to playing complex games~\cite{nature2016-silver}. In many of these domains, DNNs are now able to exceed human accuracy.  The superior performance of DNNs comes from its ability to extract high-level features from raw sensory data after using statistical learning over a large amount of data to obtain an effective representation of an input space.  This is different from earlier approaches that use hand-crafted features or rules designed by experts.

% The last two sentences have way too many undefined terms (features, statistical learning).
% It would be nice to make it more accessible

The superior accuracy of DNNs, however, comes at the cost of high computational complexity. While general-purpose compute engines, especially graphics processing units (GPUs), have been the mainstay for much DNN processing, increasingly there is interest in providing more specialized acceleration of the DNN computation. This article aims to provide an overview of DNNs, the various tools for understanding their behavior, and the techniques being explored to efficiently accelerate their computation.

This paper is organized as follows:
\begin{itemize}
\item Section~\ref{sec:background} provides background on the context of why DNNs are important, their history and applications.  
\item Section~\ref{sec:overview} gives an overview of the basic components of DNNs and popular DNN models currently in use. 
\item Section~\ref{sec:resources} describes the various resources used for DNN research and development.
\item Section~\ref{sec:architecture} describes the various hardware platforms used to process DNNs and the various optimizations used to improve throughput and energy efficiency without impacting application accuracy (i.e., produce bit-wise identical results).
\item Section~\ref{sec:technology} discusses how mixed-signal circuits and new memory technologies can be used for near-data processing to address the expensive data movement that dominates throughput and energy consumption of DNNs. 
\item Section~\ref{sec:algorithms} describes various joint algorithm and hardware optimizations that can be performed on DNNs to improve both throughput and energy efficiency while trying to minimize impact on accuracy. 
\item Section~\ref{sec:benchmarking} describes the key metrics that should be considered when comparing various DNN designs.
\end{itemize}

%% file: background.tex
\section{Background on Deep Neural Networks (DNN)}
\label{sec:background}

In this section, we describe the position of DNNs in the context of AI in general and some of the concepts that motivated its development. We will also present a brief chronology of the major steps in its history, and some current domains to which it is being applied.

\subsection{Artificial Intelligence and DNNs}

\begin{figure}
    \begin{center}
        \includegraphics[width=0.8\linewidth]{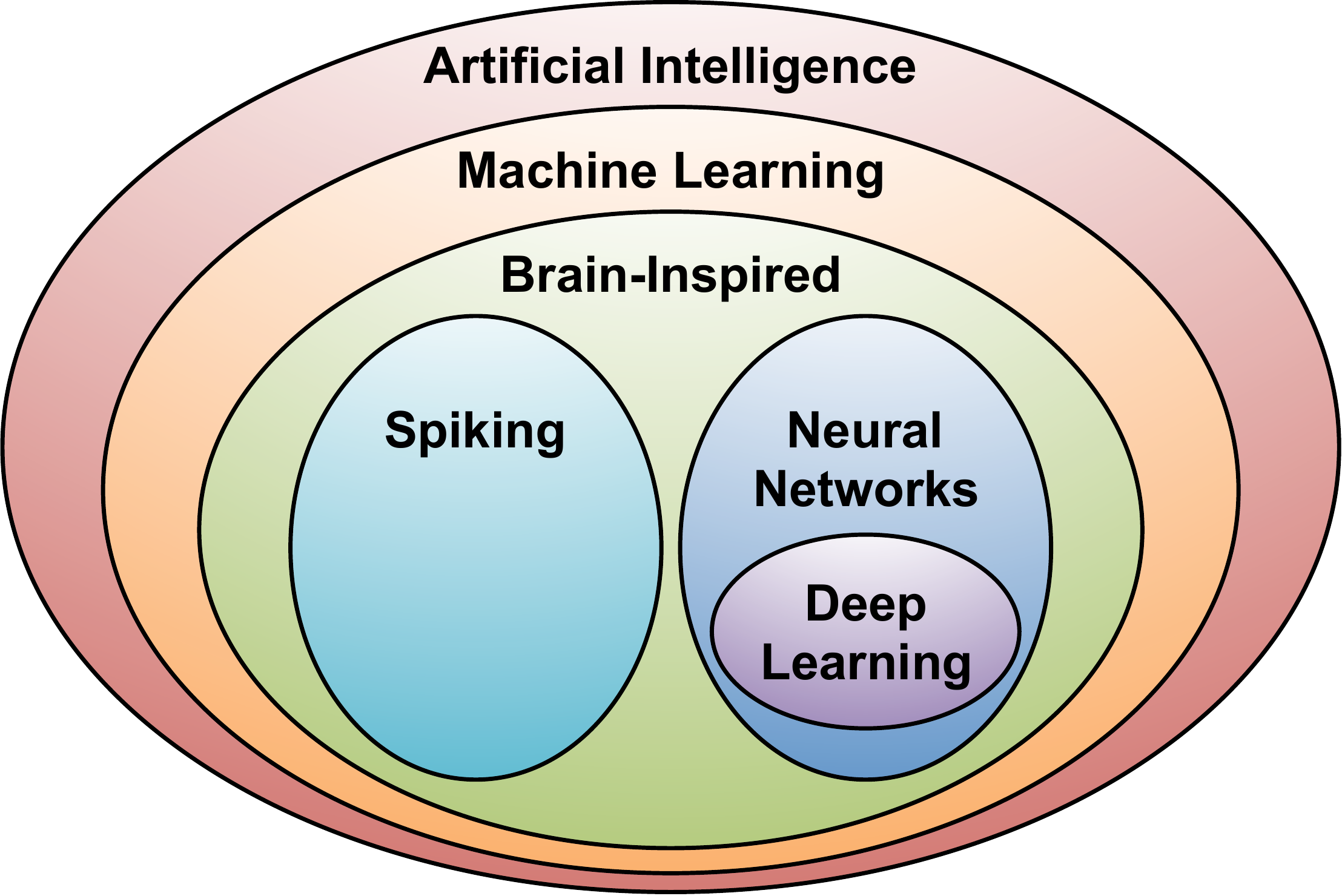}
        \caption{    Deep Learning in the context of Artificial Intelligence.
                }       
        \label{fig:venn}
    \end{center}
\end{figure}

DNNs, also referred to as deep learning, are a part of the broad field of AI, which is the science and engineering of creating intelligent machines that have the ability to achieve goals like humans do, according to John McCarthy, the computer scientist who coined the term in the 1950s. The relationship of deep learning to the whole of artificial intelligence is illustrated in Fig.~\ref{fig:venn}.

Within artificial intelligence is a large sub-field called machine learning, which was defined in 1959 by Arthur Samuel as the field of study that gives computers the ability to learn without being explicitly programmed. That means a single program, once created, will be able to learn how to do some intelligent activities outside the notion of programming. This is in contrast to purpose-built programs whose behavior is defined by hand-crafted heuristics that explicitly and statically define their behavior.

The advantage of an effective machine learning algorithm is clear. Instead of the laborious and hit-or-miss approach of creating a distinct, custom program to solve each individual problem in a domain, the single machine learning algorithm simply needs to learn, via a processes called \emph{training}, to handle each new problem. 

Within the machine learning field, there is an area that is often referred to as brain-inspired computation. Since the brain is currently the best `machine' we know for learning and solving problems, it is a natural place to look for a machine learning approach. Therefore, a brain-inspired computation is a program or algorithm that takes some aspects of its basic form or functionality from the way the brain works. This is in contrast to attempts to create a brain, but rather the program aims to emulate some aspects of how we understand the brain to operate.

\begin{figure}
    \begin{center}
        \includegraphics[width=0.8\linewidth]{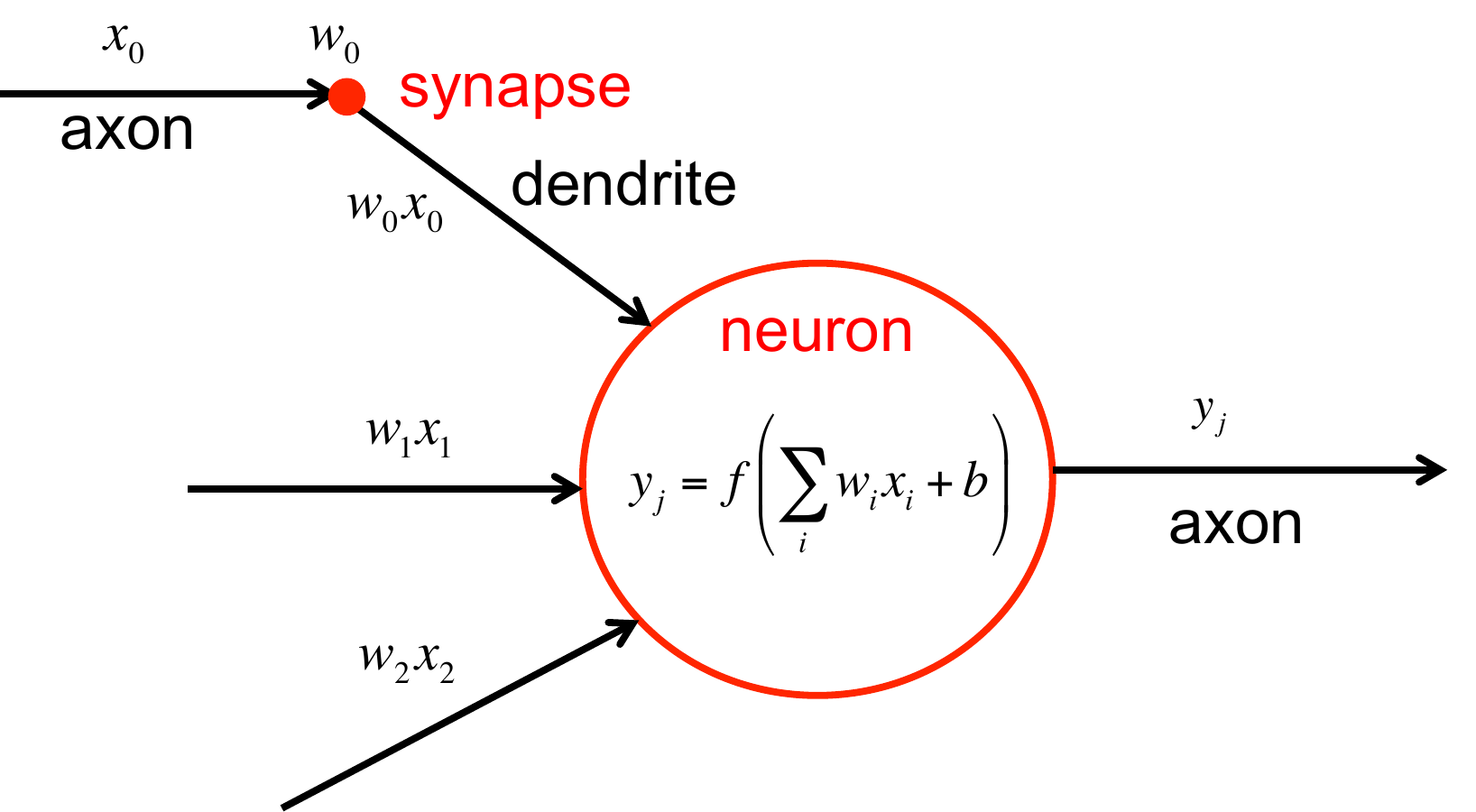}
        \caption{    Connections to a neuron in the brain. $x_{i}$, $w_{i}$, $f(\cdot)$, and $b$ are the activations, weights, non-linear function and bias, respectively.
                (Figure adopted from~\cite{cs231n}.)}       
        \label{fig:neuron}
    \end{center}
\end{figure}

Although scientists are still exploring the details of how the brain works, it is generally believed that the main computational element of the brain is the \emph{neuron}. There are approximately 86 billion neurons in the average human brain. The neurons themselves are connected together with a number of elements entering them called dendrites and an element leaving them called an axon as shown in Fig.~\ref{fig:neuron}. The neuron accepts the signals entering it via the dendrites, performs a computation on those signals, and generates a signal on the axon. These input and output signals are referred to as \emph{activations}. The axon of one neuron branches out and is connected to the dendrites of many other neurons. The connections between a branch of the axon and a dendrite is called a \emph{synapse}. There are estimated to be $10^{14}$ to $10^{15}$ synapses in the average human brain. 

% Breaking news I just saw today (3/13/2017) is that there is also some computation in the dendrites!!!
% oo lala! 
% but I haven't read the article yet...

A key characteristic of the synapse is that it can scale the signal ($x_{i}$) crossing it as shown in Fig.~\ref{fig:neuron}. That scaling factor can be referred to as a \emph{weight} ($w_{i}$), and the way the brain is believed to learn is through changes to the weights associated with the synapses. Thus, different weights result in different responses to an input.  Note that learning is the adjustment of the weights in response to a learning stimulus, while the organization (what might be thought of as the program) of the brain does not change. This characteristic makes the brain an excellent inspiration for a machine-learning-style algorithm.

Within the brain-inspired computing paradigm there is a subarea called spiking computing. In this subarea, inspiration is taken from the fact that the communication on the dendrites and axons are spike-like pulses and that the information being conveyed is not just based on a spike's amplitude. Instead, it also depends on the time the pulse arrives and that the computation that happens in the neuron is a function of not just a single value but the width of pulse and the timing relationship between different pulses. An example of a project that was inspired by the spiking of the brain is the IBM TrueNorth~\cite{merolla2014million}. In contrast to spiking computing, another subarea of brain-inspired computing is called neural networks, which is the focus of this article.\footnote{Note: Recent work using TrueNorth in a stylized fashion allows it to be used to compute reduced precision neural networks~\cite{esser2016convolutional}.  These types of neural networks are discussed in Section~\ref{ssec:precision}.}

\subsection{Neural Networks and Deep Neural Networks (DNNs)}

Neural networks take their inspiration from the notion that a neuron's computation involves a weighted sum of the input values. These weighted sums correspond to the value scaling performed by the synapses and the combining of those values in the neuron. Furthermore, the neuron doesn't just output that weighted sum, since  the computation associated with a cascade of neurons would then be a simple linear algebra operation. Instead there is a functional operation within the neuron that is performed on the combined inputs. This operation appears to be a non-linear function that causes a neuron to generate an output only if the inputs cross some threshold. Thus by analogy, neural networks apply a non-linear function to the weighted sum of the input values. We look at what some of those non-linear functions are in Section~\ref{sssec:non-linearity}.

\begin{figure}
\centering{	
    \subfigure[Neurons and synapses]{
		\includegraphics[width=0.4\linewidth]{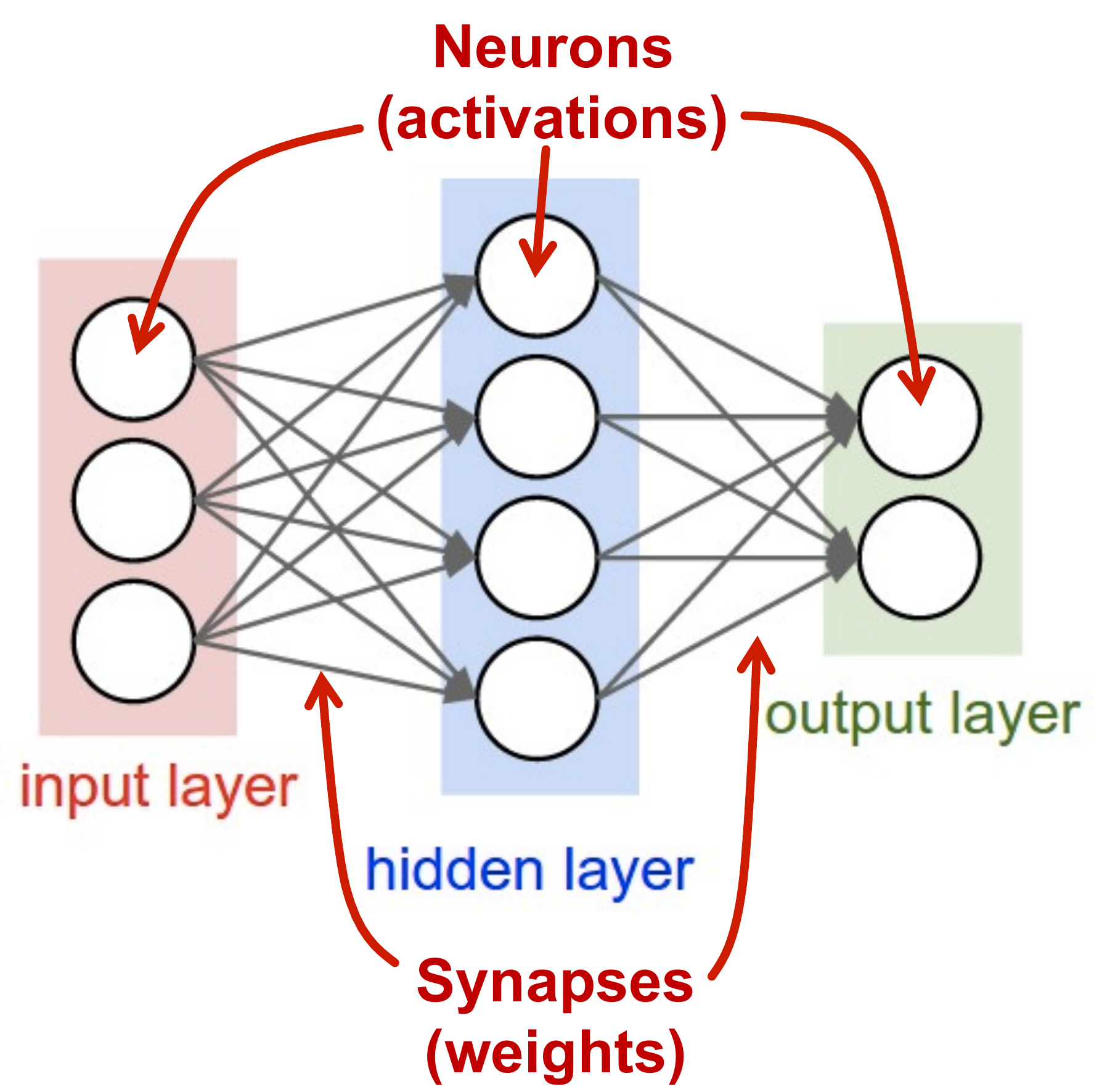}
		\label{fig:DNN_101_ns}
	}\hfill
	    \subfigure[Compute weighted sum for each layer]{
		\includegraphics[width=0.52\linewidth]{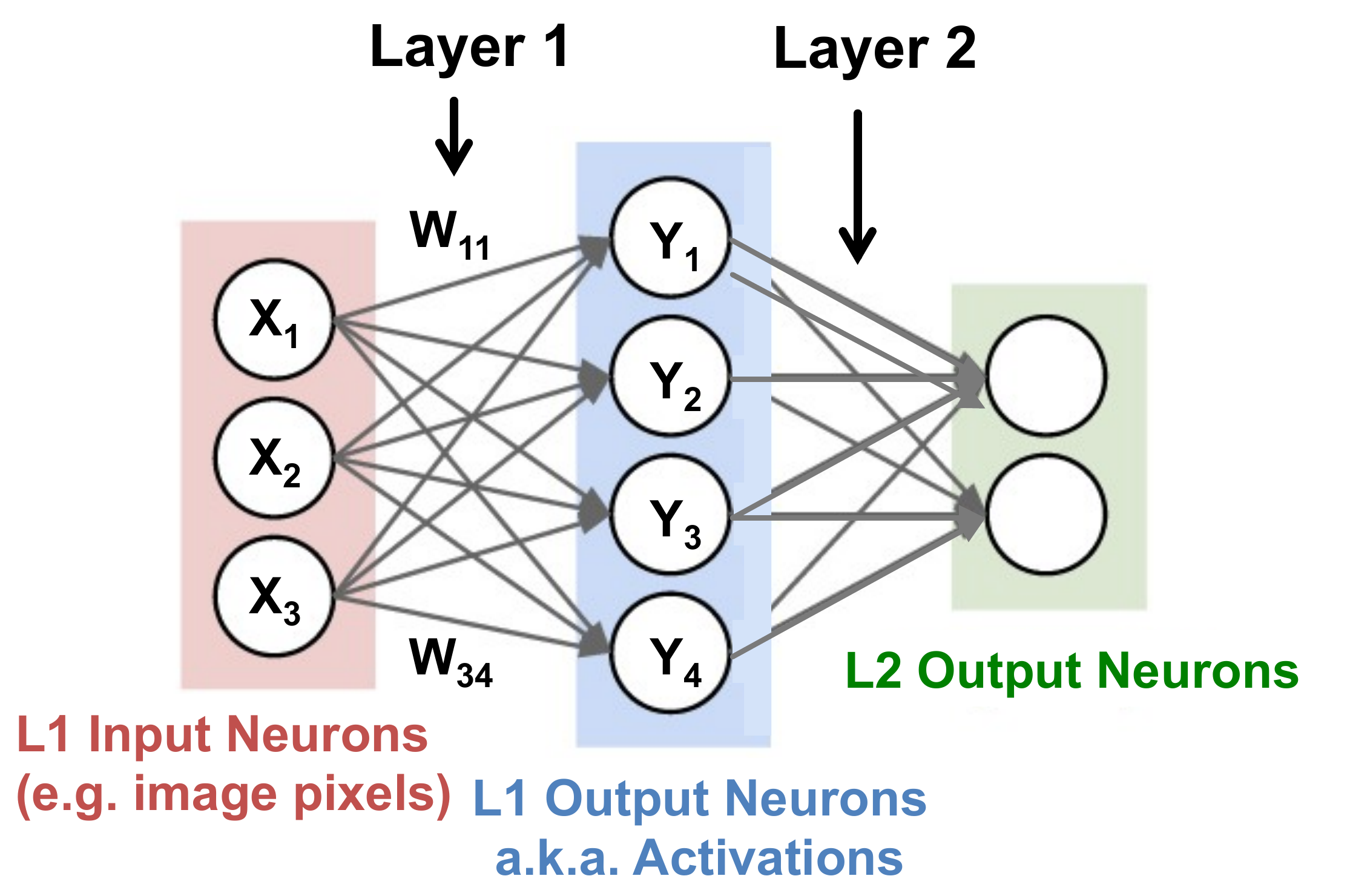}
		\label{fig:DNN_101_weights}
	}\\
}
        \caption{    Simple neural network example and terminology (Figure adopted from~\cite{cs231n}).
                }               
        \label{fig:neuralnetwork}
\end{figure}

Fig.~\ref{fig:DNN_101_ns} shows a diagrammatic picture of a computational neural network. The neurons in the input layer receive some values and propagate them to the neurons in the middle layer of the network, which is also frequently called a `hidden layer'. The weighted sums from one or more hidden layers are ultimately propagated to the output layer, which presents the final outputs of the network to the user. To align brain-inspired terminology with neural networks, the outputs of the neurons are often referred to as \emph{activations}, and the synapses are often referred to as \emph{weights} as shown in Fig.~\ref{fig:DNN_101_ns}. We will use the activation/weight nomenclature in this article.

Fig.~\ref{fig:DNN_101_weights} shows an example of the computation at each layer:
$y_{j}=f(\sum\limits_{i=1}^{3} W_{ij} \times x_{i} + b)$, where $W_{ij}$, $x_{i}$ and $y_{j}$ are the weights, input activations and output activations, respectively, and \emph{$f(\cdot)$} is a non-linear function described in Section~\ref{sssec:non-linearity}.  The bias term $b$ is omitted from Fig.~\ref{fig:DNN_101_weights} for simplicity.

% We really don't describe all the nuances of this figure yet, 
% and it uses the neuron synapse nomenclature.

Within the domain of neural networks, there is an area called \emph{deep learning}, in which the neural networks have more than three layers, i.e., more than one hidden layer. Today, the typical numbers of network layers used in deep learning range from five to more than a thousand. In this article, we will generally use the terminology \emph{deep neural networks (DNNs)} to refer to the neural networks used in deep learning.

DNNs are capable of learning high-level features with more complexity and abstraction than shallower neural networks. An example that demonstrates this point is using DNNs to process visual data. In these applications, pixels of an image are fed into the first layer of a DNN, and the outputs of that layer can be interpreted as representing the presence of different low-level features in the image, such as lines and edges. At subsequent layers, these features are then combined into a measure of the likely presence of higher level features, e.g., lines are combined into shapes, which are further combined into sets of shapes. And finally, given all this information, the network provides a probability that these high-level features comprise a particular object or scene. This deep feature hierarchy enables DNNs to achieve superior performance in many tasks.

%% file: training.tex
\subsection{Inference versus Training}
\label{ssec:training}

Since DNNs are an instance of a machine learning algorithm, the basic program does not change as it learns to perform its given tasks. In the specific case of DNNs, this learning involves determining the value of the weights (and bias) in the network, and is referred to as \emph{training} the network. Once trained, the program can perform its task by computing the output of the network using the weights determined during the training process. Running the program with these weights is referred to as \emph{inference}. 

In this section, we will use image classification, as shown in Fig.~\ref{fig:image_classification}, as a driving example for training and using a DNN.  When we perform inference using a DNN, we give an input image and the output of the DNN is a vector of scores, one for each object class; the class with the highest score indicates the most likely class of object in the image. The overarching goal for training a DNN is to determine the weights that maximize the score of the correct class and minimize the scores of the  incorrect classes. When training the network the correct class is often known because it is given for the images used for training (i.e., the training set of the network). The gap between the ideal correct scores and the scores computed by the DNN based on its current weights is referred to as the \emph{loss} ($L$). Thus the goal of training DNNs is to find a set of weights to minimize the average loss over a large training set.  

When training a network, the weights ($w_{ij}$) are usually updated using a hill-climbing optimization process called gradient descent. A multiple of the gradient of the loss relative to each weight, which is the partial derivative of the loss with respect to the weight, is used to update the weight (i.e., updated $w_{ij}^{t+1} = w_{ij}^{t} - \alpha\frac{\partial L}{\partial w_{ij}}$, where $\alpha$ is called the learning rate). Note that this gradient indicates how the weights should change in order to reduce the loss. The process is repeated iteratively to reduce the overall loss.

An efficient way to compute the partial derivatives of the gradient is through a process called \emph{backpropagation}. Backpropagation, which is a computation derived from the \emph{chain rule} of calculus, operates by passing values backwards through the network to compute how the loss is affected by each weight. 

\begin{figure}
\centering{	
    \subfigure[Compute the gradient of the loss relative to the filter inputs]{
		\includegraphics[width=0.43\linewidth]{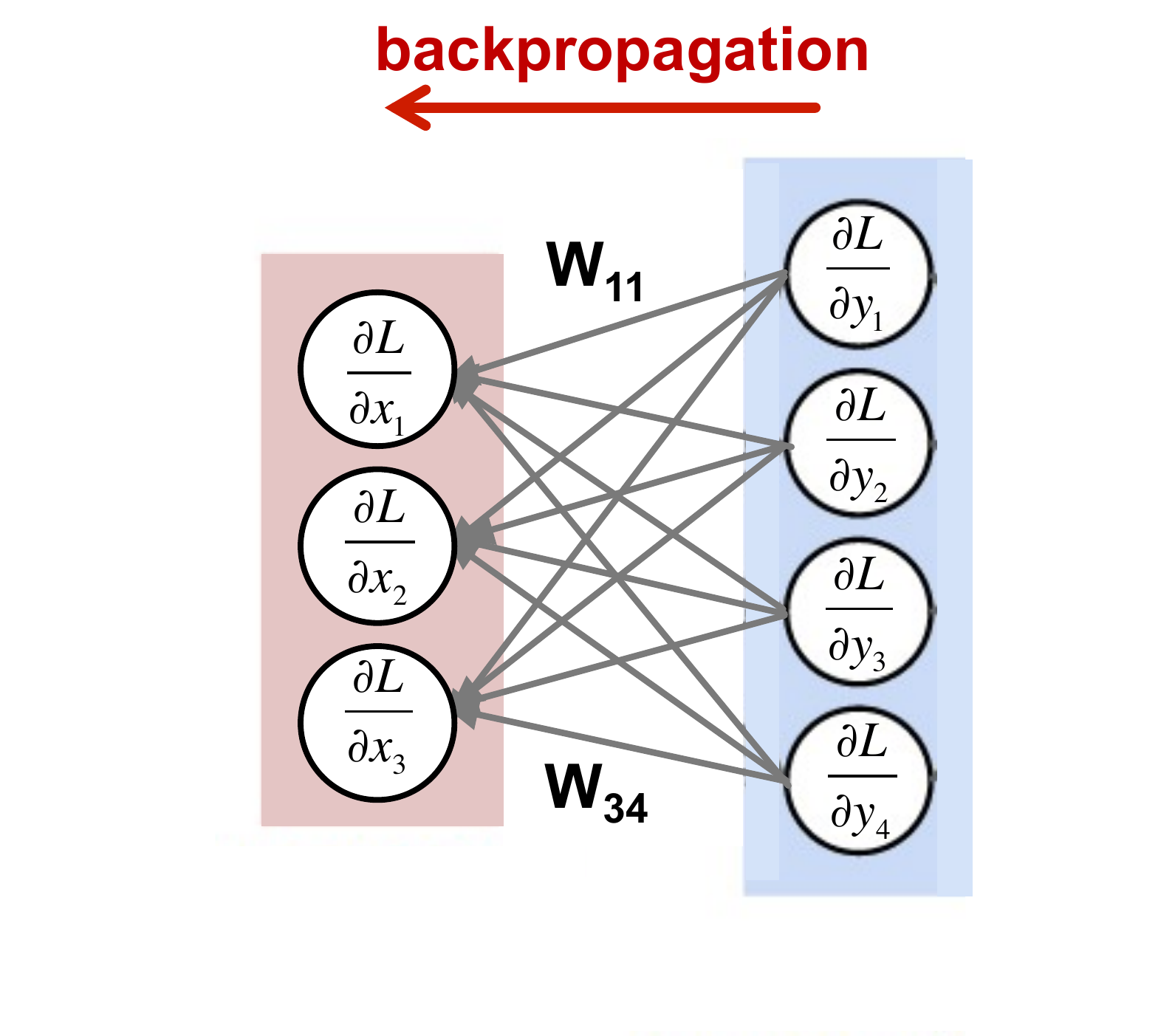}
		\label{fig:backprop_act}
	}\hfill
	    \subfigure[Compute the gradient of the loss relative to the weights]{
		\includegraphics[width=0.47\linewidth]{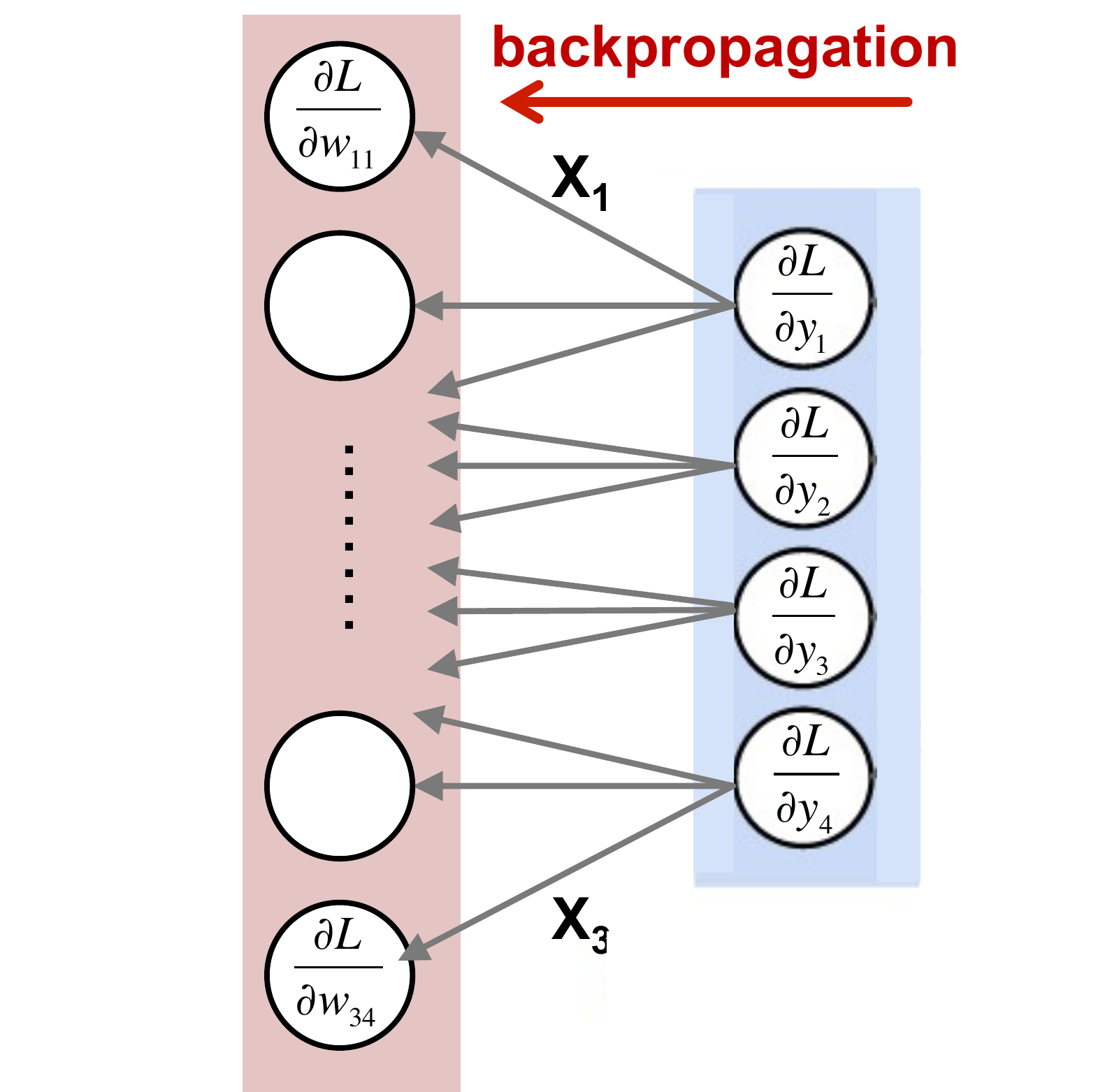}
		\label{fig:backprop_weights}
	}\\
}
        \caption{    An example of backpropagation through a neural network.
                }               
        \label{fig:backprop}
\end{figure}

This backpropagation computation is, in fact, very similar in form to the computation used for inference as shown in Fig.~\ref{fig:backprop}~\cite{mathieu2013fast}.\footnote{To backpropagate through each filter: (1) compute the gradient of the loss relative to the weights from the filter inputs (i.e., the forward activations) and the gradients of the loss relative to the filter outputs; (2) compute the gradient of the loss relative to the filter inputs from the filter weights and the gradients of the loss relative to the filter outputs.} Thus, techniques for efficiently performing inference can sometimes be useful for performing training. It is, however, important to note a couple of points. First, backpropagation requires intermediate outputs of the network to be preserved for the backwards computation, thus training has increased storage requirements. Second, due to the gradients use for hill-climbing, the precision requirement for training is generally higher than inference. Thus many of the reduced precision techniques discussed in Section~\ref{sec:algorithms} are limited to inference only. 

A variety of techniques are used to improve the efficiency and robustness of training. For example, often the loss from multiple sets of input data, i.e., a \emph{batch}, are collected before a single pass of weight update is performed; this helps to speed up and stabilize the training process.  
 
There are multiple ways to train the weights. The most common approach, as described above, is called \emph{supervised learning}, where all the training samples are labeled (e.g.,  with the correct class). \emph{Unsupervised learning} is another approach where all the training samples are not labeled and essentially the goal is to find the structure or clusters in the data. \emph{Semi-supervised learning} falls in between the two approaches where only a small subset of the training data is labeled (e.g., use unlabeled data to define the cluster boundaries, and use the small amount of labeled data to label the clusters). Finally, \emph{reinforcement learning} can be used to the train weights such that given the state of the current environment, the DNN can output what action the agent should take next to maximize expected rewards; however, the rewards might not be available immediately after an action, but instead only after a series of actions.

%to train a DNN to be a policy network such that given an input, it can output a decision on what action to take next and receive the corresponding reward; the process of training this network is to make decisions that maximize the received rewards (i.e., a reward function), and the training process must balance exploration (trying new actions) and exploitation (using actions that are known to give high rewards).

%train policy networks that is implemented as DNNs where the training is done by having an agent interact with an environment to maximize future rewards (i.e., reward function); the reward is often not available immediately after an action/decision, but instead provided at the end of a series of actions/decisions.

%such that given an input, it can output a decision on what action to take next and receive the corresponding reward; the process of 
%,and the training process must balance exploration (trying new actions) and exploitation (using actions that are known to give high rewards).  

Another commonly used approach to determine weights is \emph{fine-tuning}, where previously-trained weights are available and are used as a starting point and then those weights are adjusted for a new dataset (e.g., transfer learning) or for a new constraint (e.g., reduced precision). This results in faster training than starting from a random starting point, and can sometimes result in better accuracy.

This article will focus on the efficient processing of DNN inference rather than training, since DNN inference is often performed on embedded devices (rather than the cloud) where resources are limited as discussed in more details later.

%This section seems like it will either be very superficial or off-topic if more detailed. Let's discuss.

%\section{Hardware for Training}
%Operations (stochastic gradient decent, back prop (Fig.~\ref{ref:backprop}), forward prop)

%\begin{figure}
%    \begin{center}
%        \includegraphics[width=0.9\linewidth]{figures/backprop.pdf}
%        \caption{    Back propagation.
%                }
%        \vspace{-15pt}                
%        \label{fig:backprop}
%    \end{center}
%\end{figure}

%Hardware requirements (storage of activations (Fig.~\ref{fig:training_activation_storage}, bitwidth)
%\begin{figure}
%    \begin{center}
%        \includegraphics[width=0.9\linewidth]{figures/training_activation_storage.pdf}
%        \caption{    Storage cost of activation during training. Figure adopted from ~\cite{rhu2016vdnn}.
%                }
%        \vspace{-15pt}                
%        \label{fig:training_activation_storage}
%    \end{center}
%\end{figure}

%Synchronization across multiple compute engines (model vs. data parallelism %(Fig.~\ref{fig:batch_parameter_update}, Hogwild~\cite{recht2011hogwild}) 

%\begin{figure}
%   \begin{center}
%        \includegraphics[width=0.9\linewidth]{figures/batch_parameter_update.pdf}
%        \caption{    Batch parameter update~\cite{dean2012large}.
%                }
%        \vspace{-15pt}                
%        \label{fig:batch_parameter_update}
%    \end{center}
%\end{figure}

%% file: development_history.tex
\subsection{Development History}

Although neural nets were proposed in the 1940s, the first practical application employing multiple digital neurons didn't appear until the late 1980s with the LeNet network for hand-written digit recognition~\cite{cm1989-lecun}\footnote{In the early 1960s, single analog neuron systems were used for adaptive filtering~\cite{ire1960-widrow,IEEESignalProcessing2005-widrow}.}. Such systems are widely used by ATMs for digit recognition on checks. However, the early 2010s have seen a blossoming of DNN-based applications with highlights such as Microsoft's speech recognition system in 2011~\cite{deng2013recent} and the AlexNet system for image recognition in 2012~\cite{nips2012-krizhevsky}. A brief chronology of deep learning is shown in Fig.~\ref{fig:nnhistory}.

\begin{figure}
\begin{center}
\textbf{DNN Timeline}
\end{center}
\begin{itemize}
\item 1940s - Neural networks were proposed
\item 1960s - Deep neural networks were proposed
\item 1989 - Neural networks for recognizing digits (LeNet)
\item 1990s - Hardware for shallow neural nets (Intel ETANN)
\item 2011 - Breakthrough DNN-based speech recognition (Microsoft)
\item 2012 - DNNs for vision start supplanting hand-crafted approaches (AlexNet)
\item 2014+ - Rise of DNN accelerator research (Neuflow, DianNao...)
\end{itemize}
\caption{A concise history of neural networks. 'Deep' refers to the number of layers in the network. }
\label{fig:nnhistory}
\end{figure}

The deep learning successes of the early 2010s are believed to be a confluence of three factors. The first factor is the amount of available information to train the networks. To learn a powerful representation (rather than using a hand-crafted approach) requires a large amount of training data. For example, Facebook receives over 350 millions images per day, Walmart creates 2.5 Petabytes of customer data hourly and YouTube has 300 hours of video uploaded every minute. As a result, the cloud providers and many businesses have a huge amount of data to train their algorithms. 

The second factor is the amount of compute capacity available. Semiconductor device and computer architecture advances have continued to provide increased computing capability, and we appear to have crossed a threshold where the large amount of weighted sum computation in DNNs, which is required for both inference and training, can be performed in a reasonable amount of time. 

The successes of these early DNN applications opened the floodgates of algorithmic development. It has also inspired the development of several (largely open source) frameworks that make it even easier for researchers and practitioners to explore and use DNNs. Combining these efforts contributes to the third factor, which is the evolution of the algorithmic techniques that have improved application accuracy significantly and broadened the domains to which DNNs are being applied.

\begin{figure}
    \begin{center}
        \includegraphics[width=0.9\linewidth]{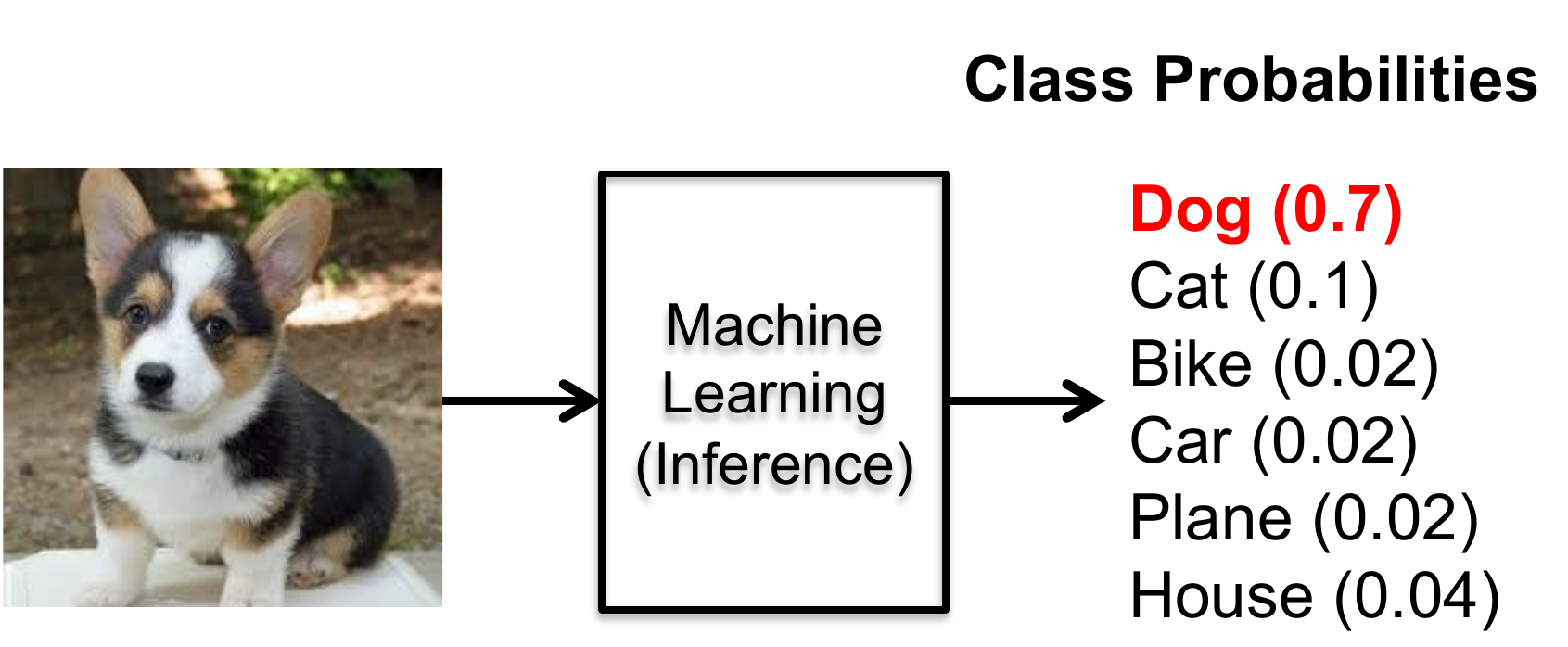}
        \caption{    Example of an image classification task. The machine learning platform takes in an image and outputs the confidence scores for a predefined set of classes.
                }               
        \label{fig:image_classification}
    \end{center}
\end{figure}

%This challenge is a contest, which had two components. 
An excellent example of the successes in deep learning can be illustrated with the ImageNet Challenge~\cite{ijcv2015-russakovsky}. This challenge is a contest involving several different components. One of the components is an image classification task where algorithms are given an image and they must identify what is in the image, as shown in Fig.~\ref{fig:image_classification}. The training set consists of 1.2 million images, each of which is labeled with one of 1000 object categories that the image contains. For the evaluation phase, the algorithm must accurately identify objects in a test set of images, which it hasn't previously seen.

%The other task is a object detection task, which is given an image with one or more objects in the image, the algorithm must identify the objects in the image. In this case, the contestants are provided with 450K images for training that are labeled with objects from 200 categories. The objective is to locate the objects in the image and identify what category the object comes from. 

\begin{figure}
    \begin{center}
        \includegraphics[width=0.9\linewidth]{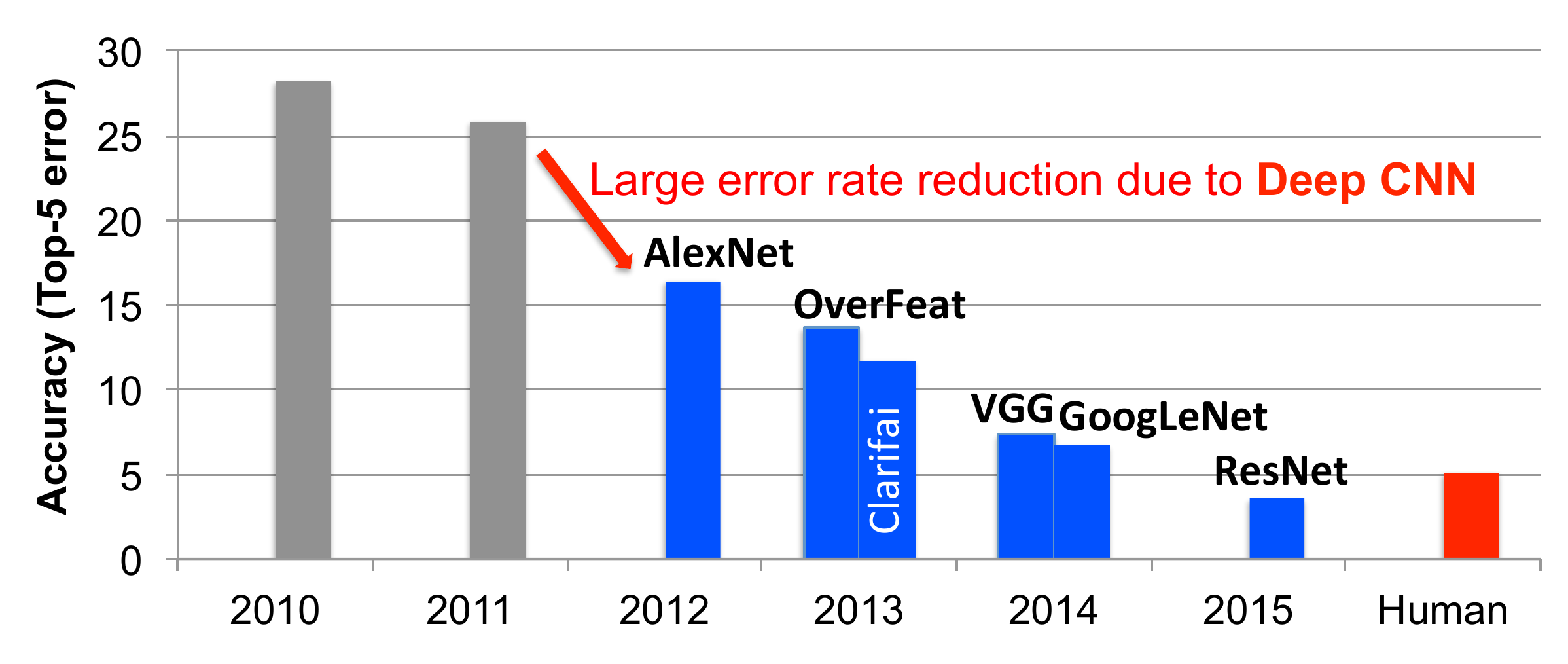}
        \caption{    Results from the ImageNet Challenge~\cite{ijcv2015-russakovsky}.
                }
        \label{fig:imagenet_challenge}
    \end{center}
\end{figure}

Fig.~\ref{fig:imagenet_challenge} shows the performance of the best entrants in the ImageNet contest over a number of years. One sees that the accuracy of the algorithms initially had an error rate of 25\% or more. In 2012, a group from the University of Toronto used graphics processing units (GPUs) for their high compute capability and a deep neural network approach, named AlexNet, and dropped the error rate by approximately 10\%~\cite{nips2012-krizhevsky}. Their accomplishment inspired an outpouring of deep learning style algorithms that have resulted in a steady stream of improvements. 

In conjunction with the trend to deep learning approaches for the ImageNet Challenge, there has been a corresponding increase in the number of entrants using GPUs. From 2012 when only 4 entrants used GPUs to 2014 when almost all the entrants (110) were using them. This reflects the almost complete switch from traditional computer vision approaches to deep learning-based approaches for the competition.

In 2015, the ImageNet winning entry, ResNet~\cite{cvpr2016-he}, exceeded human-level accuracy with a top-5 error rate\footnote{The top-5 error rate is measured based on whether the correct answer appears in one of the top 5 categories selected by the algorithm.} below 5\%.  Since then, the error rate has dropped below 3\% and more focus is now being placed on more challenging components of the competition, such as object detection and localization. These successes are clearly a contributing factor to the wide range of applications to which DNNs are being applied.

%% file: applications.tex
\subsection{Applications of DNN}
\label{sec:applications}
Many applications can benefit from DNNs ranging from multimedia to medical space. In this section, we will provide examples of areas where DNNs are currently making an impact and highlight emerging areas where DNNs hope to make an impact in the future.  

\begin{itemize}
    \item \textbf{Image and Video}
Video is arguably the biggest of the big data.  It accounts for over 70\% of today's Internet traffic~\cite{ciscoVNI}.  For instance, over 800 million hours of video is collected daily worldwide for video surveillance~\cite{video_surveillance}.  Computer vision is necessary to extract meaningful information from video.  DNNs have significantly improved the accuracy of many computer vision tasks such as image classification~\cite{ijcv2015-russakovsky}, object localization and detection~\cite{cvpr2014-girshick}, image segmentation~\cite{cvpr2015-long}, and action recognition~\cite{simonyan2014two}.   

% Lots of jargon in the tasks.

    \item \textbf{Speech and Language} 
DNNs have significantly improved the accuracy of speech recognition~\cite{hinton2012deep} as well as many related tasks such as machine translation~\cite{deng2013recent}, natural language processing~\cite{collobert2011natural}, and audio generation~\cite{van2016wavenet}. 
%(it's used for the acoustic model)
    \item \textbf{Medical} 
DNNs have played an important role in genomics to gain insight into the genetics of diseases such as autism, cancers, and spinal muscular atrophy~\cite{xiong2015human, zhou2015predicting,alipanahi2015predicting,zeng2016convolutional}.  They have also been used in medical imaging to detect skin cancer~\cite{esteva2017dermatologist}, brain cancer~\cite{jermyn2016neural} and breast cancer~\cite{wang2016deep}.

    \item \textbf{Game Play}
Recently, many of the grand AI challenges involving game play have been overcome using DNNs. These successes also required innovations in training techniques and many rely on reinforcement learning~\cite{kaelbling1996reinforcement}. DNNs have surpassed human level accuracy in playing Atari~\cite{mnih2013playing} as well as Go~\cite{nature2016-silver}, where an exhaustive search of all possibilities is not feasible due to the unimaginably huge number of possible moves.%; for instance, AlphaGo beat the world champion in 2016. In many of these games, it is not feasible to do an exhaustive search of all possibilities due to the complexity of the game; for instance, in the game of Go, the number of possible moves exceeds the number of atoms in the universe.  DNNs are able to learn a representation (specifically the policy and value functions trained with reinforcement learning) that can be used to effectively control agents in the game.  With its high dimension, DNNs can store information that is difficult to articulate directly with rules.

\item \textbf{Robotics}
DNNs have been successful in the domain of robotic tasks such as grasping with a robotic arm~\cite{levine2016end}, motion planning for ground robots~\cite{pfeiffer2016perception}, visual navigation~\cite{gupta2017cognitive, chen2015deepdriving}, control to stabilize a quadcopter~\cite{zhang2016learning} and driving strategies for autonomous vehicles~\cite{shalev2016safe}.  %Similar to game play, in most of these applications the DNNs are used to represent a policy network that is trained with reinforcement learning.   These policy networks take in sensor data (often in raw form) and the output actuation or control signals.

\end{itemize}

DNNs are already widely used in multimedia applications today (e.g., computer vision, speech recognition).  Looking forward, we expect that DNNs will likely play an increasingly important role in the medical and robotics fields, as discussed above, as well as finance (e.g., for trading, energy forecasting, and risk assessment), infrastructure (e.g., structural safety, and traffic control), weather forecasting and event detection~\cite{future_DNN_applications}. The myriad application domains pose new challenges to the efficient processing of DNNs; the solutions then have to be adaptive and scalable in order to handle the new and varied forms of DNNs that these applications may employ.

%% file: embedded.tex
\subsection{Embedded versus Cloud}
% vivienne to fix

The various applications and aspects of DNN processing (i.e., training versus inference) have different computational needs. Specifically, training often requires a large dataset\footnote{One of the major drawbacks of DNNs is their need for large datasets to prevent over-fitting during training.} and significant computational resources for multiple weight-update iterations. In many cases, training a DNN model still takes several hours to multiple days and thus is typically performed in the cloud. Inference, on the other hand, can happen either in the cloud or at the edge (e.g., IoT or mobile).

In many applications, it is desirable to have the DNN inference processing near the sensor. For instance, in computer vision applications, such as measuring wait times in stores or predicting traffic patterns, it would be desirable to extract meaningful information from the video right at the image sensor rather than in the cloud to reduce the communication cost. For other applications such as autonomous vehicles, drone navigation and robotics, local processing is desired since the latency and security risks of relying on the cloud are too high. However, video involves a large amount of data, which is computationally complex to process; thus, low cost hardware to analyze video is challenging yet critical to enabling these applications. Speech recognition enables us to seamlessly interact with electronic devices, such as smartphones. While currently most of the processing for applications such as Apple Siri and Amazon Alexa voice services is in the cloud, it is still desirable to perform the recognition on the device itself to reduce latency and dependency on connectivity, and to improve privacy and security.  %Speech recognition is the first step before many other AI tasks such as machine translation, natural language processing, etc. %Low power hardware for speech recognition is explored in~\cite{price20156,yazdaniultra}.

Many of the embedded platforms that perform DNN inference have stringent energy consumption, compute and memory cost limitations; efficient processing of DNNs have thus become of prime importance under these constraints. Therefore, in this article, we will focus on the compute requirements for inference rather than training.

%% file: overview.tex
\section{Overview of DNNs}
\label{sec:overview}
DNNs come in a wide variety of shapes and sizes depending on the application. The popular shapes and sizes are also evolving rapidly to improve accuracy and efficiency. In all cases, the input to a DNN is a set of values representing the information to be analyzed by the network. For instance, these values can be pixels of an image, sampled amplitudes of an audio wave or the numerical representation of the state of some system or game.

\begin{figure}
\centering{	
	    \subfigure[Feedforward versus feedback (recurrent) networks]{
		\includegraphics[width=0.45\linewidth]{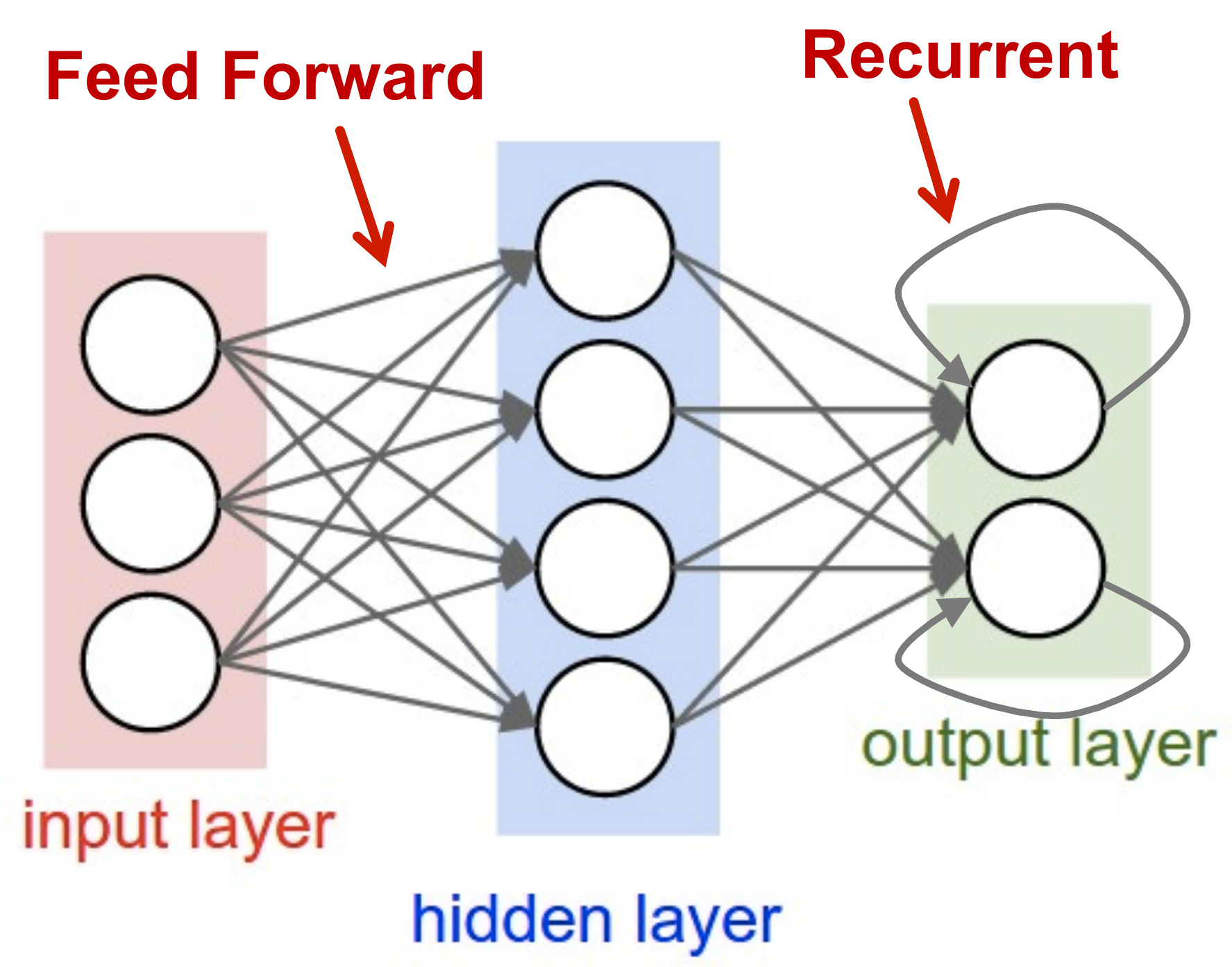}
		\label{fig:DNN_101_ff_fb}
	}\hfill		
    \subfigure[Fully connected versus sparse]{
		\includegraphics[width=0.45\linewidth]{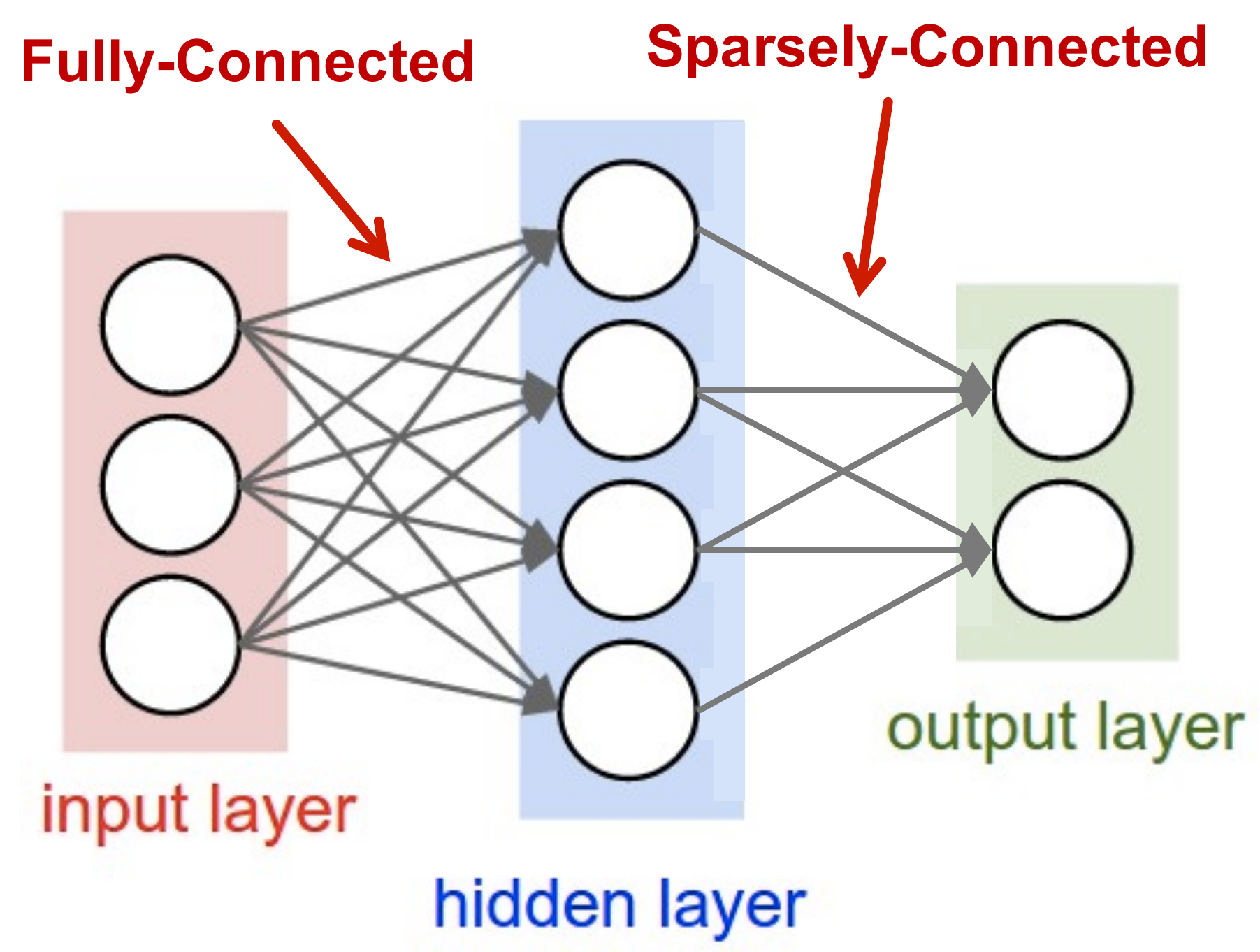}
		\label{fig:DNN_101_fc_sparse}
	}

}
        \caption{    Different types of neural networks (Figure adopted from~\cite{cs231n}).
                }               
        \label{fig:neuralnetworktypes}
\end{figure}

The networks that process the input come in two major forms: feed forward and recurrent as shown in Fig.~\ref{fig:DNN_101_ff_fb}. In feed-forward networks all of the computation is performed as a sequence of operations on the outputs of a previous layer. The final set of operations generates the output of the network, for example a probability that an image contains a particular object, the probability that an audio sequence contains a particular word, a bounding box in an image around an object or the proposed action that should be taken. In such DNNs, the network has no memory and the output for an input is always the same irrespective of the sequence of inputs previously given to the network.

In contrast, recurrent neural networks (RNNs), of which Long Short-Term Memory networks (LSTMs)~\cite{hochreiter1997long} are a popular variant, have internal memory to allow long-term dependencies to affect the output. In these networks, some intermediate operations generate values that are stored internally in the network and used as inputs to other operations in conjunction with the processing of a later input. In this article, we will focus on feed-forward networks since (1) the major computation in RNNs is still the weighted sum, which is covered by the feed-forward networks, and (2) to-date little attention has been given to hardware acceleration specifically for RNNs.

DNNs can be composed solely of \emph{fully-connected} (FC) layers (also referred to as multi-layer perceptrons, or MLP) as shown in the leftmost layer of~Fig.~\ref{fig:DNN_101_fc_sparse}. In a FC layer, all output activations are composed of a weighted sum of all input activations (i.e., all outputs are connected to all inputs). This requires a significant amount of storage and computation. Thankfully, in many applications, we can remove some connections between the activations by setting the weights to zero without affecting accuracy. This results in a \emph{sparsely-connected layer}. A sparsely connected layer is illustrated in the rightmost layer of~Fig.~\ref{fig:DNN_101_fc_sparse}.

We can also make the computation more efficient by limiting the number of weights that contribute to an output. This sort of structured sparsity can arise if each output is only a function of a fixed-size window of inputs. Even further efficiency can be gained if the same set of weights are used in the calculation of every output. This repeated use of the same weight values is called \emph{weight sharing} and can significantly reduce the storage requirements for weights. 

\begin{figure}
\centering{
    \subfigure[2-D convolution in traditional image processing]{
		\includegraphics[width=0.9\linewidth]{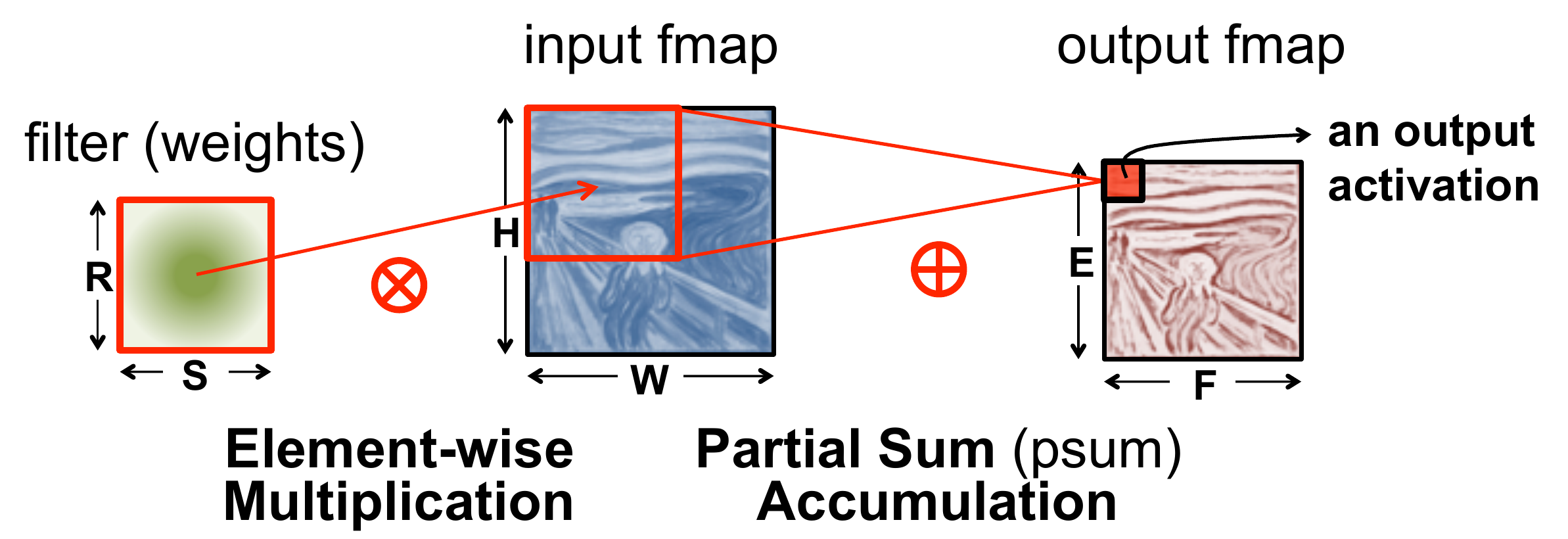}
		\label{fig:2D_conv}
	}	
    \subfigure[High dimensional convolutions in CNNs]{
		\includegraphics[width=0.9\linewidth]{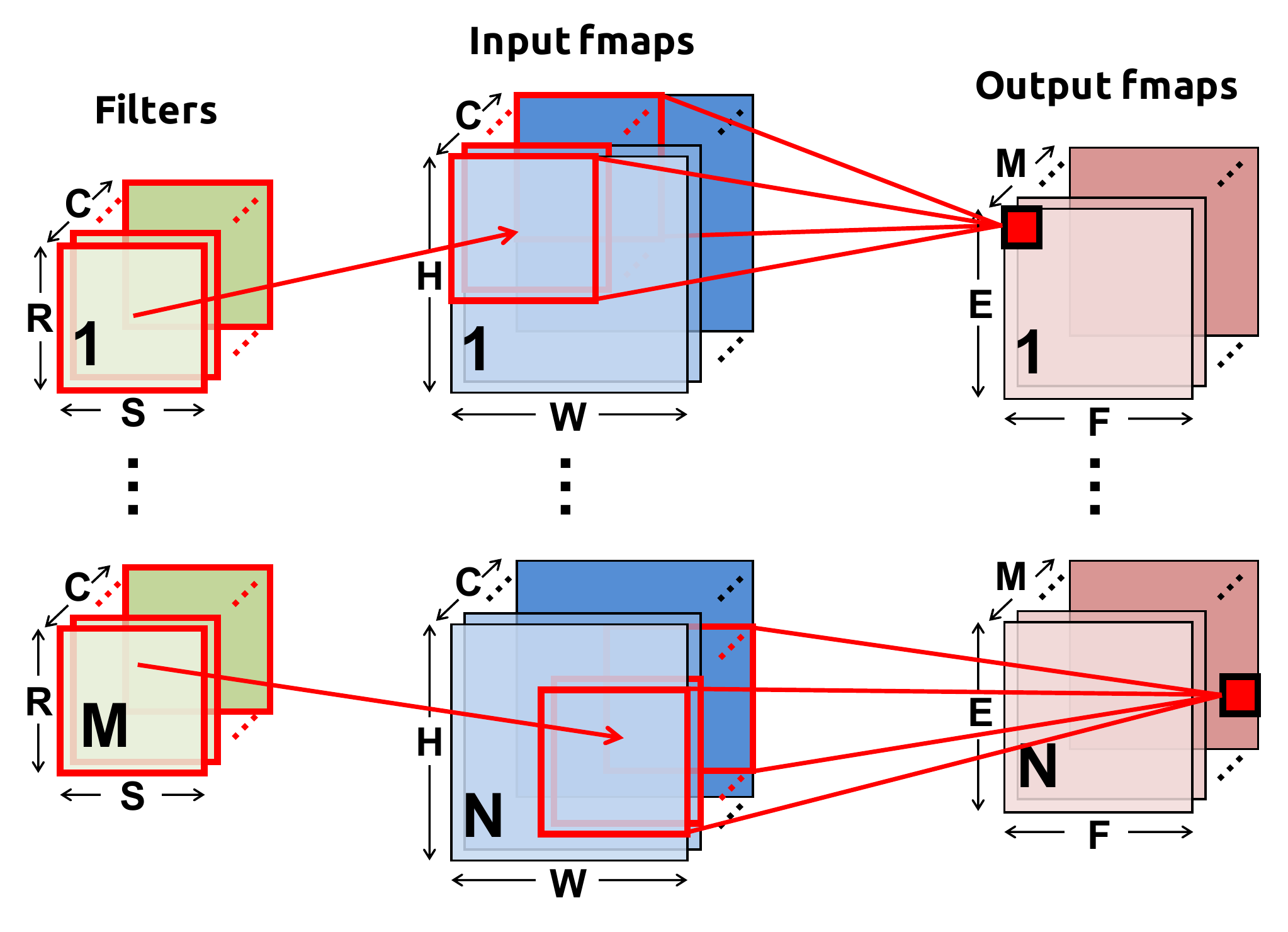}
				\label{fig:DNN_conv}
	}
}
\caption{Dimensionality of convolutions.}
\label{fig:convolutions}
\end{figure}

An extremely popular windowed and weight-shared DNN layer arises by structuring the computation as a convolution, as shown in Fig.~\ref{fig:2D_conv}, where the weighted sum for each output activation is computed using only a small neighborhood of input activations (i.e., all weights beyond beyond the neighborhood are set to zero), and where the same set of weights are shared for every output (i.e., the filter is space invariant). Such convolution-based layers are referred to as \emph{convolutional} (CONV) layers.~\footnote{Note: the structured sparsity in CONV layers is orthogonal to the sparsity that occurs from network pruning as described in Section~\ref{ssec:pruning}.} 

 % A good example of this is in image processing, where neighboring pixels in an image are correlated, and this correlation often holds regardless of the region within the image.  As a result, image processing often involves 

\subsection{Convolutional Neural Networks (CNNs)}

\begin{figure}
    \begin{center}
        \includegraphics[width=1\linewidth]{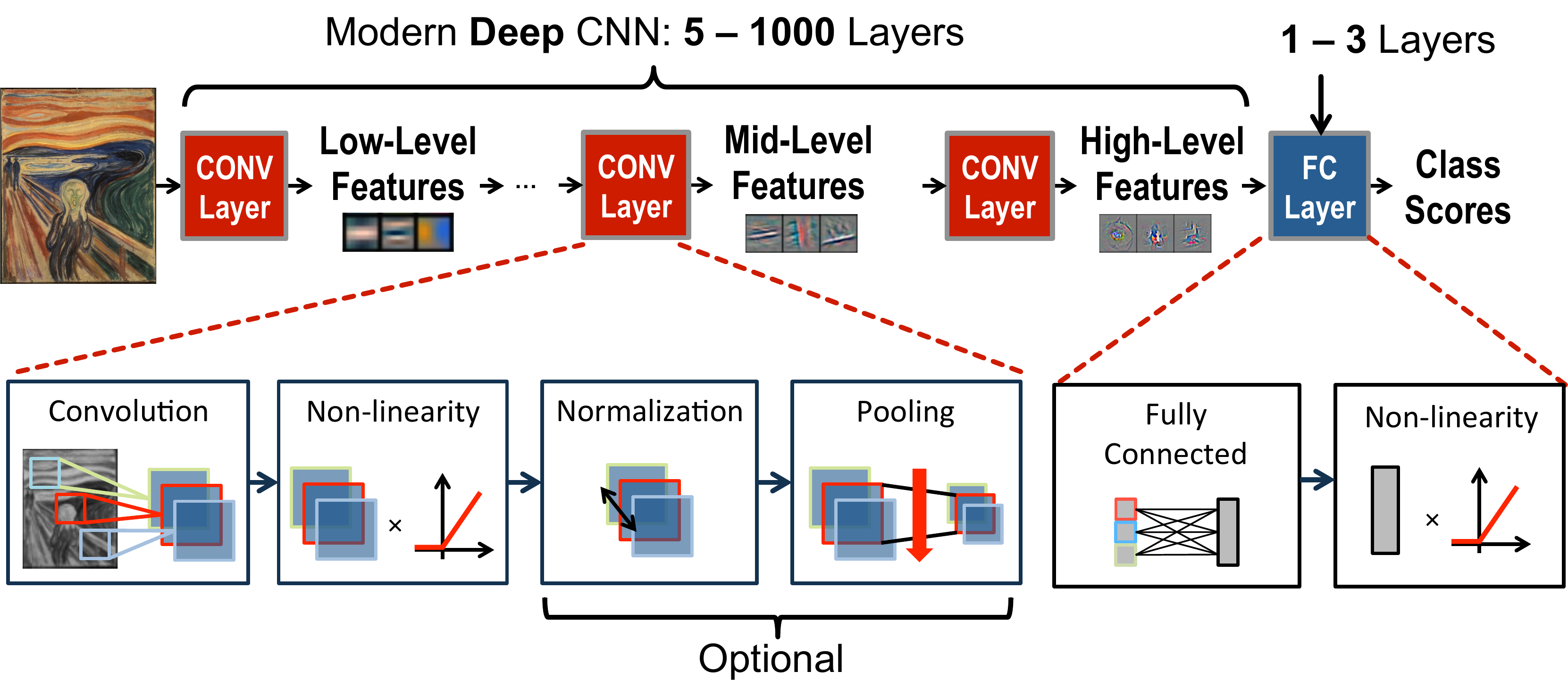}
        \caption{   Convolutional Neural Networks. 
                }      
        \label{fig:DNN_overview}
    \end{center}
\end{figure}

A common form of DNNs is \emph{Convolutional Neural Nets} (CNNs), which are composed of multiple CONV layers as shown in Fig.~\ref{fig:DNN_overview}. In such networks, each layer generates a successively higher-level abstraction of the input data, called a \emph{feature map} (fmap), which preserves essential yet unique information. Modern CNNs are able to achieve superior performance by employing a very deep hierarchy of layers. CNN are widely used in a variety of applications including image understanding~\cite{nips2012-krizhevsky}, speech recognition~\cite{sainath2013deep}, game play~\cite{nature2016-silver}, robotics~\cite{levine2016end}, etc. This paper will focus on its use in image processing, specifically for the task of image classification~\cite{nips2012-krizhevsky}.

Each of the CONV layers in CNN is primarily composed of high-dimensional convolutions as shown in Fig.~\ref{fig:DNN_conv}. In this computation, the input activations of a layer are structured as a set of 2-D \emph{input feature maps} (ifmaps), each of which is called a \emph{channel}. Each channel is convolved with a distinct 2-D filter from the stack of filters, one for each channel; this stack of 2-D filters is often referred to as a single 3-D filter. The results of the convolution at each point are summed across all the channels. In addition, a 1-D bias can be added to the filtering results, but some recent networks~\cite{cvpr2016-he} remove its usage from parts of the layers. The result of this computation is the output activations that comprise one channel of \emph{output feature map} (ofmap). Additional 3-D filters can be used on the same input to create additional output channels. Finally, multiple input feature maps may be processed together as a \emph{batch} to potentially improve reuse of the filter weights.

Given the shape parameters in Table~\ref{table:conv_layer_parameters}, the computation of a CONV layer is defined as

\begin{equation}
\label{eq:convolutional_layer_math}
\scriptsize
\begin{split}
& \mathbf{O}[z][u][x][y] = \mathbf{B}[u] + \sum_{k=0}^{C-1}\sum_{i=0}^{S-1}\sum_{j=0}^{R-1}\mathbf{I}[z][k][Ux+i][Uy+j]\times \mathbf{W}[u][k][i][j], \\ 
& 0\leq z <N, 0\leq u < M, 0\leq x < F, 0\leq y <E, \\
& E = (H-R+U)/U, F = (W-S+U)/U.
\end{split}
\end{equation}
$\mathbf{O}$, $\mathbf{I}$, $\mathbf{W}$ and $\mathbf{B}$ are the matrices of the ofmaps, ifmaps, filters and biases, respectively. $U$ is a given stride size. Fig.~\ref{fig:DNN_conv} shows a visualization of this computation (ignoring biases). 

\begin{table}[t]
    \centering
    \begin{tabular}{|c|l|}
        \hline
        \textbf{Shape Parameter}   & \textbf{Description}                 \\
        \hline
        $N$     & batch size of 3-D fmaps                        \\
        \hline
        $M$     & \# of 3-D filters / \# of ofmap channels       \\
        \hline
        $C$     & \# of ifmap/filter channels                   \\
        \hline
        $H/W$     & ifmap plane height/width                      \\
        \hline
        $R/S$     & filter plane height/width (= $H$ or $W$ in FC)       \\
        \hline
        $E/F$     & ofmap plane height/width (= 1 in FC)          \\
        \hline
    \end{tabular}
    % \vspace{-5pt}
    \caption{   Shape parameters of a CONV/FC layer.
            }
    \label{table:conv_layer_parameters}
\end{table}

To align the terminology of CNNs with the generic DNN, 
\begin{itemize}
    \item filters are composed of weights (i.e., synapses)
    \item input and output feature maps (ifmaps, ofmaps) are composed of activations (i.e., input and output neurons)
\end{itemize}

From five~\cite{nips2012-krizhevsky} to more than a thousand~\cite{cvpr2016-he} CONV layers are commonly used in recent CNN models. A small number, e.g., 1 to 3, of fully-connected (FC) layers are typically applied after the CONV layers for classification purposes. A FC layer also applies filters on the ifmaps as in the CONV layers, but the filters are of the same size as the ifmaps. Therefore, it does not have the weight sharing property of CONV layers. Eq.~(\ref{eq:convolutional_layer_math}) still holds for the computation of FC layers with a few additional constraints on the shape parameters: $H = R$, $F = S$, $E = F = 1$, and $U = 1$.  

In addition to CONV and FC layers, various optional layers can be found in a DNN such as the non-linearity, pooling, and normalization.  The function and computations for each of these layers are discussed next.

\subsubsection{Non-Linearity}
\label{sssec:non-linearity}
\begin{figure}
    \begin{center}
        \includegraphics[width=0.9\linewidth]{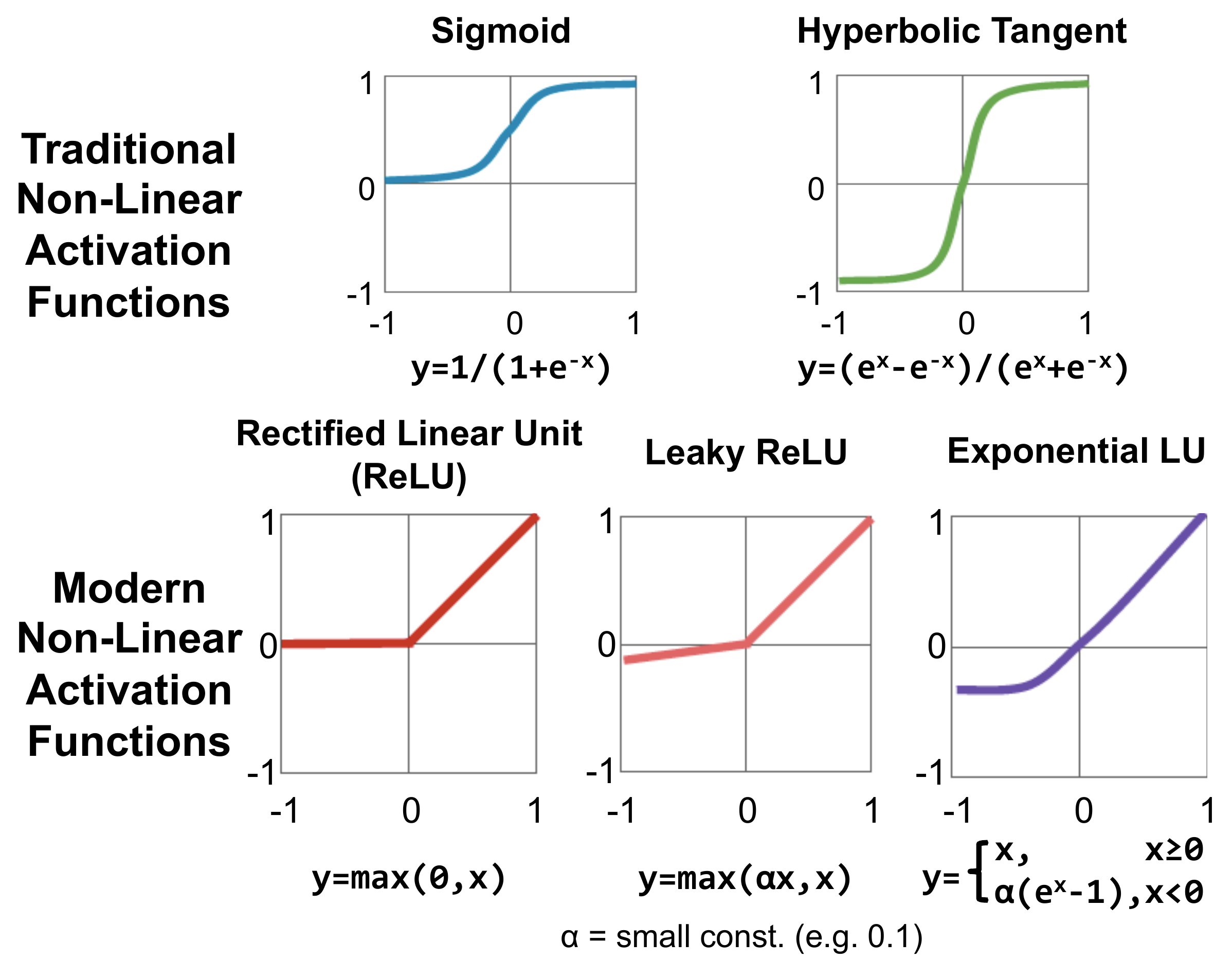}
        \caption{   Various forms of non-linear activation functions (Figure adopted from Caffe Tutorial~\cite{caffe}).
                }      
        \label{fig:activations}
    \end{center}
\end{figure}

A non-linear activation function is typically applied after each CONV or FC layer.  Various non-linear functions are used to introduce non-linearity into the DNN as shown in Fig.~\ref{fig:activations}.  These include historically conventional non-linear functions such as sigmoid or hyperbolic tangent as well as rectified linear unit (ReLU)~\cite{icml2014-nair}, which has become popular in recent years due to its simplicity and its ability to enable fast training. Variations of ReLU, such as leaky ReLU~\cite{maas2013rectifier}, parametric ReLU~\cite{He_2015_ICCV}, and exponential LU~\cite{clevert2015fast} have also been explored for improved accuracy. Finally, a non-linearity called maxout, which takes the max value of two intersecting linear functions, has shown to be effective in speech recognition tasks~\cite{zhang2014improving, zhang2017towards}.

% leaky reLU or maxout or exponential elu

\subsubsection{Pooling}

\begin{figure}
    \begin{center}
        \includegraphics[width=0.9\linewidth]{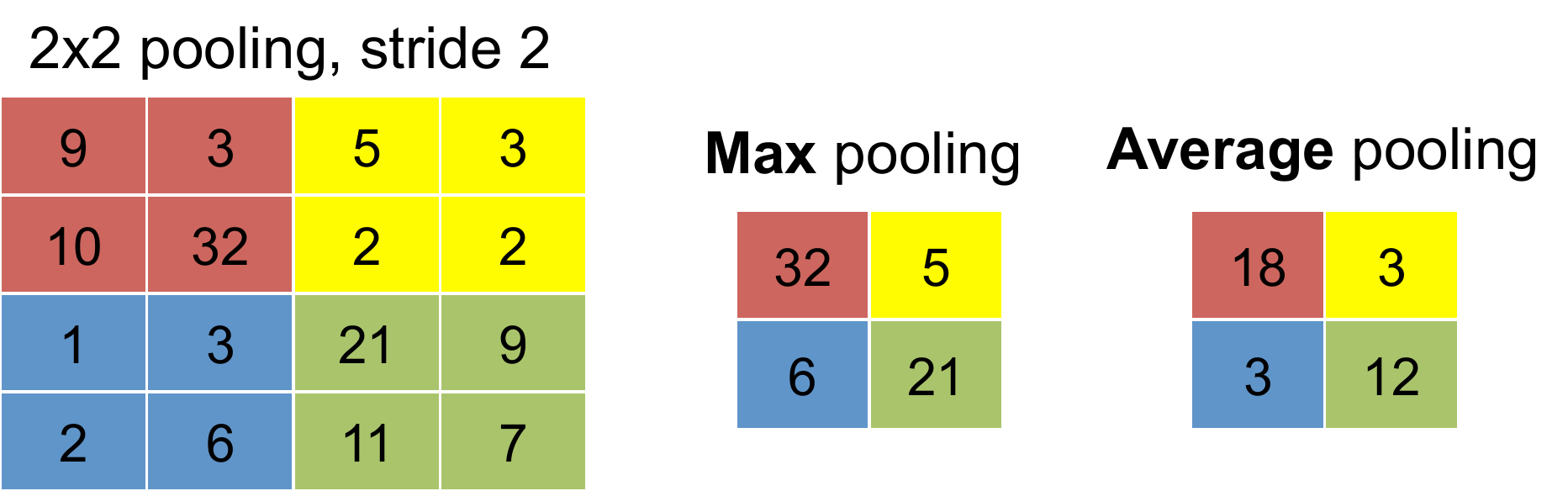}
        \caption{   Various forms of pooling (Figure adopted from Caffe Tutorial~\cite{caffe}).
                }      
        \label{fig:pooling}
    \end{center}
    \vspace{-10pt}
\end{figure}

A variety of computations that reduce the dimensionality of a feature map are referred to as \emph{pooling}. Pooling, which is applied to each channel separately, enables the network to be robust and invariant to small shifts and distortions. Pooling combines, or \emph{pools}, a set of values in its \emph{receptive field} into a smaller number of values. It can be configured based on the size of its receptive field (e.g., 2$\times$2) and pooling operation (e.g., max or average), as shown in Fig.~\ref{fig:pooling}.  Typically pooling occurs on non-overlapping blocks (i.e., the stride is equal to the size of the pooling).  Usually a stride of greater than one is used such that there is a reduction in the dimension of the representation (i.e., feature map). 

%Each of the CONV and FC layers is also immediately followed by an activation (ACT) layer, such as a rectified linear unit~\cite{icml2014-nair}.  These different fuctions will be discussed next.

\subsubsection{Normalization}
Controlling the input distribution across layers can help to significantly speed up training and improve accuracy.  Accordingly, the distribution of the layer input activations ($\sigma$, $\mu$) are normalized such that it has a zero mean and a unit standard deviation.  In batch normalization (BN), the normalized value is further scaled and shifted, as shown in Eq.~(\ref{eq:normalization}), where the parameters ($\gamma$, $\beta$)  are learned from training~\cite{ioffe2015batch}. $\epsilon$ is a small constant to avoid numerical problems. Prior to this, local response normalization (LRN)~\cite{nips2012-krizhevsky} was used, which was inspired by lateral inhibition in neurobiology where excited neurons (i.e., high value activations) should subdue its neighbors (i.e., cause low value activations); however, BN is now considered standard practice in the design of CNNs while LRN is mostly deprecated. Note that while LRN usually is performed after the non-linear function, BN is mostly performed between the CONV or FC layer and the non-linear function.
\begin{equation}
\label{eq:normalization}
\begin{split}
y=\frac{x-\mu}{\sqrt{\sigma^{2}+\epsilon}}\gamma+\beta
\end{split}
\end{equation}

% JSE - As per Vivienne's suggestion I moved Popular DNN models here

\subsection{Popular DNN Models}
\label{sec:popular}

Many DNN models have been developed over the past two decades.  Each of these models has a different `network architecture' in terms of number of layers, layer types, layer shapes (i.e., filter size, number of channels and filters), and connections between layers. Understanding these variations and trends is important for incorporating the right flexibility in any efficient DNN engine.

% jse - Some consequences on an accelerator for the attributes in each example would be useful.

%Although the first popular DNN, LeNet~\cite{ieee1998-lecun}, was published in the 1990s, it wasn't until 2012 that AlexNet~\cite{nips2012-krizhevsky} was used in the ImageNet Challenge~\cite{ijcv2015-russakovsky}. 
In this section, we will give an overview of various popular DNNs such as LeNet~\cite{ieee1998-lecun} as well as those that competed in and/or won the ImageNet Challenge~\cite{ijcv2015-russakovsky} as shown in Fig.~\ref{fig:imagenet_challenge}, most of whose models with pre-trained weights are publicly available for download; the DNN models are summarized in Table~\ref{tab:popular_dnns}. Two results for top-5 error results are reported.  In the first row, the accuracy is boosted by using multiple crops from the image and an ensemble of multiple trained models (i.e., the DNN needs to be run several times); these results were used to compete in the ImageNet Challenge. The second row reports the accuracy if only a single crop was used (i.e., the DNN is run only once), which is more consistent with what would likely be deployed in real-time and/or energy-constrained applications.

%The ImageNet Challenge involves a classification task into one of 1000 categories; the accuracy is measured based on whether the correct answer appears in one of the top 5 categories selected by the classifier. Since then, DNN-based approaches, specifically CNNs, have dominated the ImageNet Challenge.  In 2015, the winning entry, ResNet~\cite{cvpr2016-he}, exceeded human-level accuracy with a top-5 error rate below 5\%. Since then, the error rate has dropped below 3\% and more focus is now being placed on more challenging tasks such as object detection and localization.

\emph{LeNet}~\cite{cm1989-lecun} was one of the first CNN approaches introduced in 1989.  It was designed for the task of digit classification in grayscale images of size 28$\times$28. The most well known version, LeNet-5, contains two CONV layers and two FC layers~\cite{ieee1998-lecun}. Each CONV layer uses filters of size 5$\times$5 (1 channel per filter) with 6 filters in the first layer and 16 filters in the second layer. Average pooling of 2$\times$2 is used after each convolution and a sigmoid is used for the non-linearity.  In total, LeNet requires 60k weights and 341k multiply-and-accumulates (MACs) per image. LeNet led to CNNs' first commercial success, as it was deployed in ATMs to recognize digits for check deposits.%\footnote{Note: There are various versions of LeNet-5 available online.  For instance, in Model Zoo of Caffe, LeNet has 431k weights for the filters and requires 2.3M MACs per image, and uses ReLU rather than sigmoid~\cite{caffe_mnist}.} 
% specifically, the average is computed for the four inputs, scaled by a trained coefficient, and a bias is added.  The result is passed through a sigmoid function for the non-linearity, where the coefficient and bias control the effect of this non-linearity.

\emph{AlexNet}~\cite{nips2012-krizhevsky} was the first CNN to win the ImageNet Challenge in 2012. It consists of five CONV layers and three FC layers. Within each CONV layer, there are 96 to 384 filters and the filter size ranges from 3$\times$3 to 11$\times$11, with 3 to 256 channels each.  In the first layer, the 3 channels of the filter correspond to the red, green and blue components of the input image. A ReLU non-linearity is used in each layer.  Max pooling of 3$\times$3 is applied to the outputs of layers 1, 2 and 5.  To reduce computation, a stride of 4 is used at the first layer of the network. AlexNet introduced the use of LRN in layers 1 and 2 before the max pooling, though LRN is no longer popular in later CNN models. One important factor that differentiates AlexNet from LeNet is that the number of weights is much larger and the shapes vary from layer to layer. To reduce the amount of weights and computation in the second CONV layer, the 96 output channels of the first layer are split into two groups of 48 input channels for the second layer, such that the filters in the second layer only have 48 channels. Similarly, the weights in fourth and fifth layer are also split into two groups.  In total, AlexNet requires 61M weights and 724M MACs to process one 227$\times$227 input image.   %A modified version of AlexNet is also available in Caffe framework called CaffeNet with the order of LRN and pooling reversed (check!).

\emph{Overfeat}~\cite{arxiv2013-sermanet} has a very similar architecture to AlexNet with five CONV layers and three FC layers. The main differences are that the number of filters is increased for layers 3 (384 to 512), 4 (384 to 1024), and 5 (256 to 1024), layer 2 is not split into two groups, the first fully connected layer only has 3072 channels rather than 4096, and the input size is 231$\times$231 rather than 227$\times$227.  As a result, the number of weights grows to 146M and the number of MACs grows to 2.8G per image. Overfeat has two different models: fast (described here) and accurate. The accurate model used in the ImageNet Challenge gives a 0.65\% lower top-5 error rate than the fast model at the cost of 1.9$\times$ more MACs

\begin{figure}
\centering{
    \subfigure[Constructing a 5$\times$5 support from 3$\times$3 filters. Used in VGG-16.]{
		\includegraphics[width=0.9\linewidth]{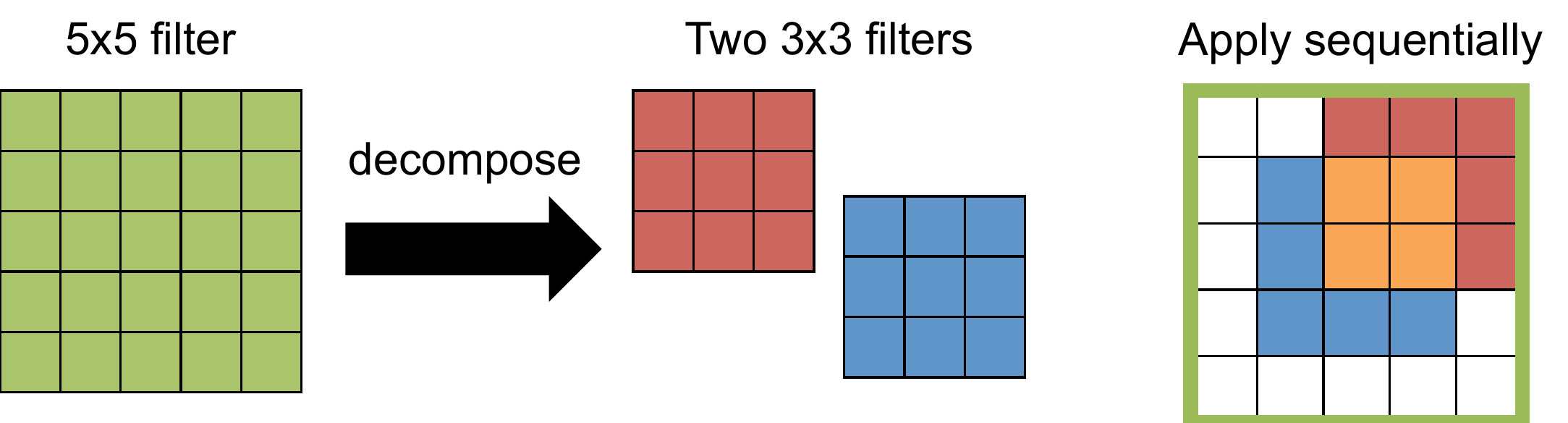}
		\label{fig:decompose_2d}
	}	
    \subfigure[Constructing a 5$\times$5 support from 1$\times$5 and 5$\times$1 filter. Used in GoogleNet/Inception v3 and v4.]{
		\includegraphics[width=0.9\linewidth]{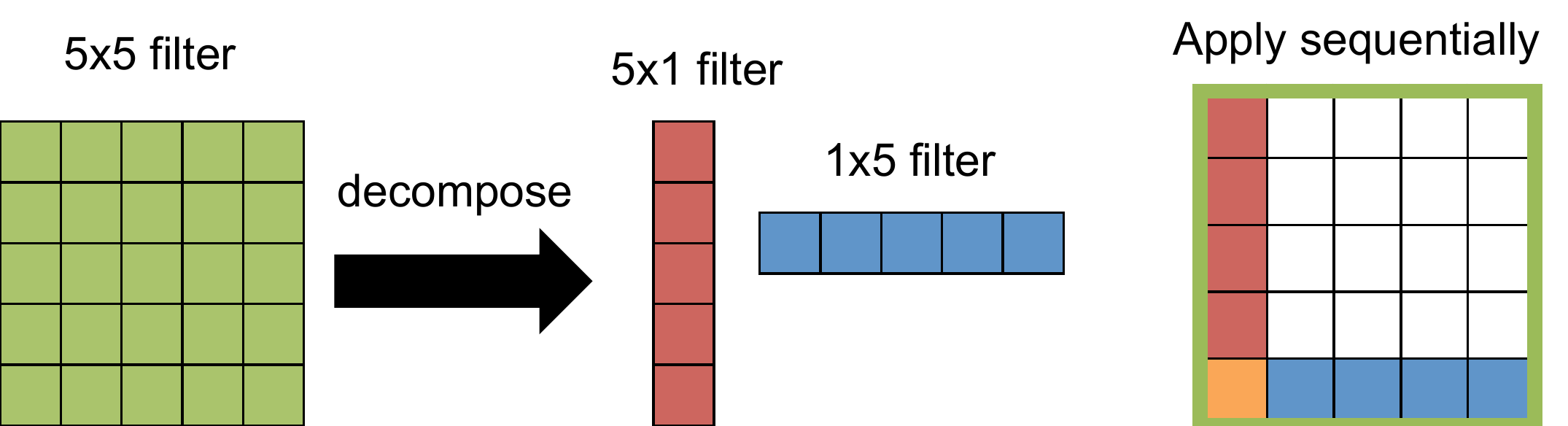}
				\label{fig:decompose_1d}
	}
}
\caption{Decomposing larger filters into smaller filters.}
\label{fig:decompose_filters}
\end{figure}

\emph{VGG-16}~\cite{iclr2015-simonyan} goes deeper to 16 layers consisting of 13 CONV layers and 3 FC layers. In order to balance out the cost of going deeper, larger filters (e.g., 5$\times$5) are built from multiple smaller filters (e.g., 3$\times$3), which have fewer weights, to achieve the same receptive fields as shown in Fig.~\ref{fig:decompose_2d}. As a result, all CONV layers have the same filter size of 3$\times$3.  In total, VGG-16 requires 138M weights and 15.5G MACs to process one 224$\times$224 input image. VGG has two different models: VGG-16 (described here) and VGG-19. VGG-19 gives a 0.1\%  lower top-5 error rate than VGG-16 at the cost of 1.27$\times$ more MACs.

\begin{figure}
    \begin{center}
        \includegraphics[width=0.9\linewidth]{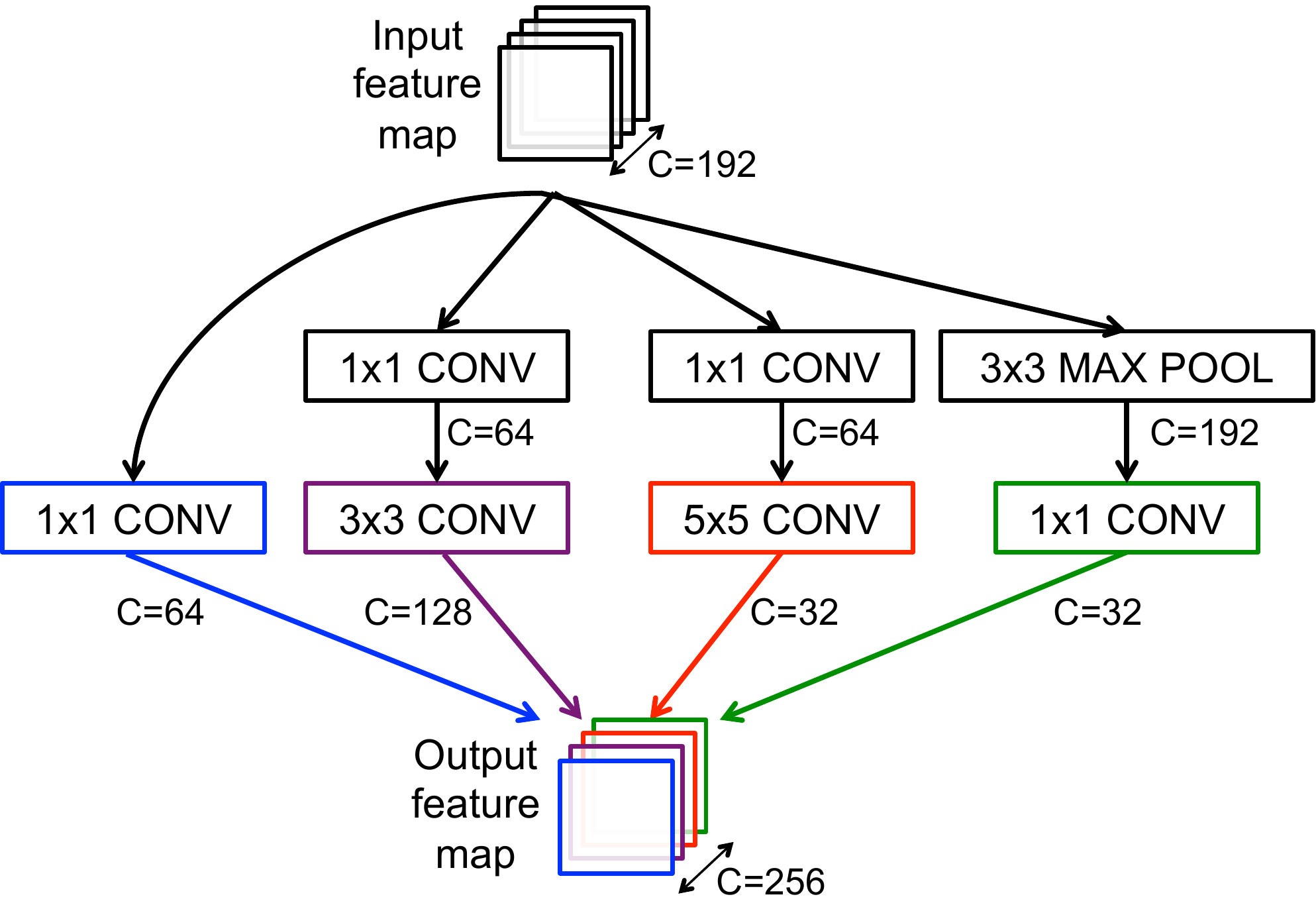}
        \caption{    Inception module from GoogleNet~\cite{cvpr2015-szegedy} with example channel lengths.  Note that each CONV layer is followed by a ReLU (not drawn).
                }       
        \label{fig:inception}
    \end{center}
\end{figure}

\emph{GoogLeNet}~\cite{cvpr2015-szegedy} goes even deeper with 22 layers. It introduced an inception module, shown in Fig.~\ref{fig:inception}, which is composed of parallel connections, whereas previously there was only a single serial connection. Different sized filters (i.e., 1$\times$1, 3$\times$3, 5$\times$5), along with 3$\times$3 max-pooling, are used for each parallel connection and their outputs are concatenated for the module output.  Using multiple filter sizes  has the effect of processing the input at multiple scales. For improved training speed, GoogLeNet is designed such that the weights and the activations, which are stored for backpropagation during training, could all fit into the GPU memory. In order to reduce the number of weights, 1$\times$1 filters are applied as a `bottleneck' to reduce the number of channels for each filter~\cite{lin2013network}.  The 22 layers consist of three CONV layers, followed by 9 inceptions layers (each of which are two CONV layers deep), and one FC layer. Since its introduction in 2014, GoogleNet (also referred to as Inception) has multiple versions: v1 (described here), v3~\footnote{v2 is very similar to v3.} and v4. Inception-v3 decomposes the convolutions by using smaller 1-D filters as shown in Fig.~\ref{fig:decompose_1d} to reduce number of MACs and weights in order to go deeper to 42 layers.  In conjunction with batch normalization~\cite{ioffe2015batch}, v3 achieves over 3\% lower top-5 error than v1 with 2.5$\times$ increase in computation~\cite{szegedy2016rethinking}. Inception-v4 uses residual connections~\cite{szegedy2016inception}, described in the next section, for a 0.4\% reduction in error.

\begin{figure}
\centering{
    \subfigure[Without bottleneck]{
		\includegraphics[width=0.4\linewidth]{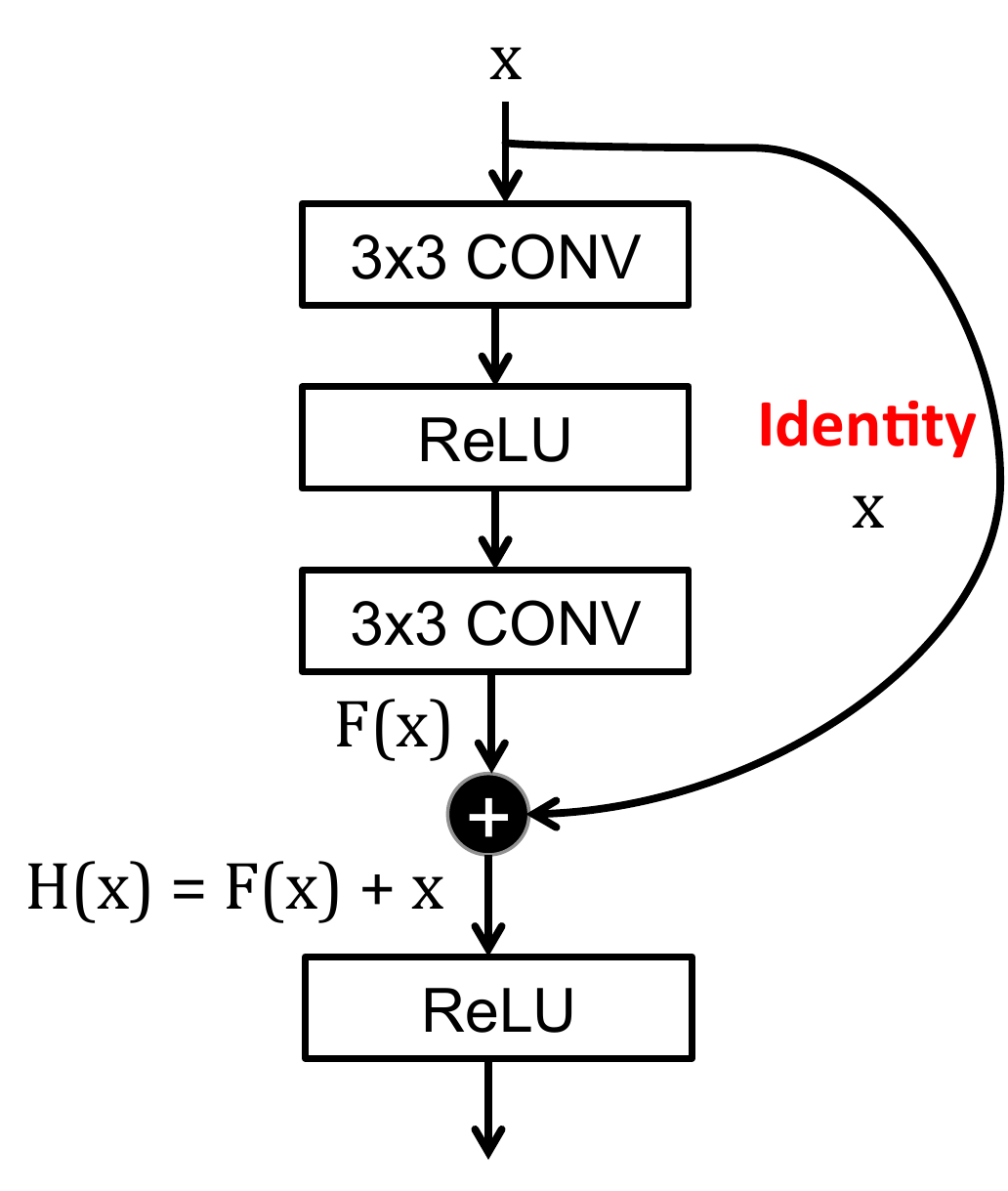}
		\label{fig:resnet_wo}
	}	
    \subfigure[With bottleneck]{
		\includegraphics[width=0.5\linewidth]{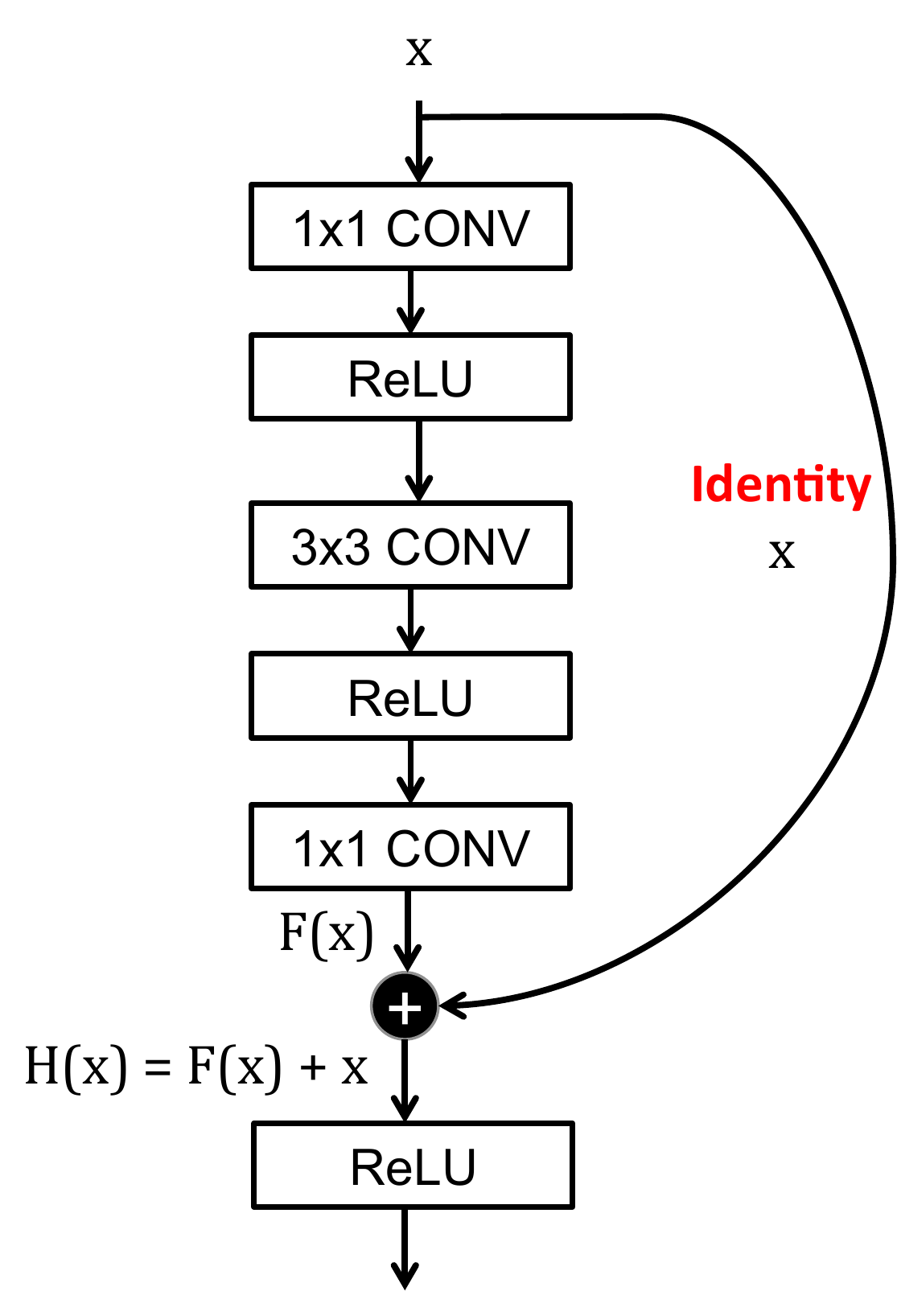}
				\label{fig:resnet_w}
	}
}
\caption{Shortcut module from ResNet~\cite{cvpr2016-he}. Note that ReLU following last CONV layer in short cut is \emph{after} the addition.}
\label{fig:resnet}
\end{figure}

\emph{ResNet}~\cite{cvpr2016-he}, also known as Residual Net, uses residual connections to go even deeper (34 layers or more). It was the first entry DNN in ImageNet Challenge that exceeded human-level accuracy with a top-5 error rate below 5\%. One of the challenges with deep networks is the vanishing gradient during training: as the error backpropagates through the network the gradient shrinks, which affects the ability to update the weights in the earlier layers for very deep networks. Residual net introduces a `shortcut' module which contains an identity connection such that the weight layers (i.e., CONV layers) can be skipped as shown in Fig.~\ref{fig:resnet}. Rather than learning the function for the weight layers $F(x)$, the shortcut module learns the residual mapping ($F(x)=H(x)-x$).  Initially, $F(x)$ is zero and the identity connection is taken; then gradually during training, the actual forward connection through the weight layer is used.  This is similar to the LSTM networks that are used for sequential data.  ResNet also uses the `bottleneck' approach of using 1$\times$1 filters to reduce the number of weight parameters.  As a result, the two layers in the shortcut module are replaced by three layers (1$\times$1, 3$\times$3, 1$\times$1) where the 1$\times$1 reduces and then increases (restores) the number of weights. ResNet-50 consists of one CONV layer, followed by 16 shortcut layers (each of which are three CONV layers deep), and one FC layer; it requires 25.5M weights and 3.9G MACs per image. There are various versions of ResNet with multiple depths (e.g., \emph{without bottleneck:} 18, 34; \emph{with bottleneck:} 50, 101, 152). The ResNet with 152 layers was the winner of the ImageNet Challenge requiring 11.3G MACs and 60M weights. Compared to ResNet-50, it reduces the top-5 error by around 1\% at the cost of 2.9$\times$ more MACs and 2.5$\times$ more weights.

Several trends can be observed in the popular DNNs shown in Table~\ref{tab:popular_dnns}.  Increasing the depth of the network tends to provide higher accuracy. Controlling for number of weights, a deeper network can support a wider range of non-linear functions that are more discriminative and also provides more levels of hierarchy in the learned representation~\cite{urban2016deep,iclr2015-simonyan, cvpr2015-szegedy, cvpr2016-he}. The number of filter shapes continues to vary across layers, thus flexibility is still important. Furthermore, most of the computation has been placed on CONV layers rather than FC layers. In addition, the number of weights in the FC layers is reduced and in most recent networks (since GoogLeNet) the CONV layers also dominate in terms of weights. Thus, the focus of hardware implementations should be on addressing the efficiency of the CONV layers, which in many domains are increasingly important.

\begin{table*}
\centering
\begin{tabular}{|c|c|c|c|c|c|c|}
\hline
\multirow{2}{*}{\textbf{Metrics}} & \textbf{LeNet} & \textbf{AlexNet} &\textbf{Overfeat} & \textbf{VGG}& \textbf{GoogLeNet}& \textbf{ResNet}\\
& \textbf{5} & \textbf{}& \textbf{fast} & \textbf{16}& \textbf{v1}& \textbf{50}\\\noalign{\hrule height 2pt}
\textbf{Top-5 error$^{\dagger}$}& {n/a} & {16.4} & {14.2} & {7.4}& {6.7}& {5.3}\\\hline
\textbf{Top-5 error (single crop)$^{\dagger}$} & {n/a} &{19.8} &{17.0} & {8.8}& {10.7}& {7.0}\\\hline
\textbf{Input Size} & {28$\times$28}& {227$\times$227} & {231$\times$231}& {224$\times$224}& {224$\times$224}& {224$\times$224}\\\noalign{\hrule height 1.5pt}
\textbf{\# of CONV Layers}& {2} & {5} & {5}& {13}& {57}& {53}\\\hline
\textbf{Depth in \# of CONV Layers}& {2} & {5} & {5}& {13}& {21}& {49}\\\hline
\textbf{Filter Sizes} &{5}& {3,5,11} &  {3,5,11}& {3}& {1,3,5,7}& {1,3,7}\\\hline
\textbf{\# of Channels} &{1, 20} & {3-256} & {3-1024}& {3-512}& {3-832}& {3-2048}\\\hline
\textbf{\# of Filters} &{20, 50}& {96-384} & {96-1024} & {64-512}& {16-384}& {64-2048}\\\hline
\textbf{Stride} &{1} & {1,4}& {1,4} & {1}& {1,2}& {1,2}\\\hline
\textbf{Weights}& {2.6k} & {2.3M}& {16M} & {14.7M}& {6.0M}& {23.5M}\\\hline
\textbf{MACs}& {283k} & {666M} & {2.67G} & {15.3G}& {1.43G}& {3.86G}\\\noalign{\hrule height 1.5pt}
\textbf{\# of FC Layers}& {2} & {3} & {3}& {3}& {1}& {1}\\\hline
\textbf{Filter Sizes} &{1,4}& {1,6} &  {1,6,12}& {1,7}& {1}& {1}\\\hline
\textbf{\# of Channels} &{50, 500} & {256-4096} & {1024-4096}& {512-4096}& {1024}& {2048}\\\hline
\textbf{\# of Filters} &{10, 500}& {1000-4096} & {1000-4096} & {1000-4096}& {1000}& {1000}\\\hline
\textbf{Weights}& {58k} & {58.6M} & {130M}& {124M}& {1M}& {2M}\\\hline
\textbf{MACs}& {58k} & {58.6M} & {130M}& {124M}& {1M}& {2M}\\\noalign{\hrule height 1.5pt}
\textbf{Total Weights}& {60k} & {61M} & {146M} & {138M}& {7M}& {25.5M}\\\hline
\textbf{Total MACs}& {341k} & {724M} & {2.8G}& {15.5G}& {1.43G}& {3.9G}\\\noalign{\hrule height 1.5pt}
\textbf{Pretrained Model Website}& {\cite{caffe_mnist}$^{\ddagger}$}& {\cite{caffe_model_zoo, matcovnet_models}} & {n/a}& {\cite{caffe_model_zoo, matcovnet_models, tensorflow_models}}&  {\cite{caffe_model_zoo, matcovnet_models, tensorflow_models}}& {\cite{caffe_model_zoo, matcovnet_models, tensorflow_models}}\\\hline
\end{tabular}
\caption{Summary of popular DNNs \cite{ieee1998-lecun,nips2012-krizhevsky,iclr2015-simonyan, cvpr2015-szegedy, cvpr2016-he}.  $^{\dagger}$Accuracy is measured based on Top-5 error on ImageNet~\cite{ijcv2015-russakovsky}.
$^{\ddagger}$This version of LeNet-5 has 431k weights for the filters and requires 2.3M MACs per image, and uses ReLU rather than sigmoid.}
\label{tab:popular_dnns}
\end{table*}

%\subsection{Recurrent Neural Networks (RNNs)}
%\label{ssec:RNN}
% VS -- to polish
%Used for sequential data (speech, NLP, video, image captioning); use sigmoid more than ReLU;

%bidirectional LSTM, GRU

%Typical sizes; currently more popular on cloud than CNN according to TPU paper

%Unlike CNN, not many regular structures.

%% file: resources.tex
\section{DNN development resources}
\label{sec:resources}

One of the key factors that has enabled the rapid development of DNNs is the set of development resources that have been made available by the research community and industry. These resources are also key to the development of DNN accelerators by providing characterizations of the workloads and facilitating the exploration of trade-offs in model complexity and accuracy. This section will describe these resources such that those who are interested in this field can quickly get started.

\subsection{Frameworks}

For ease of DNN development and to enable sharing of trained networks, several deep learning frameworks have been developed from various sources. These open source libraries contain software libraries for DNNs. Caffe was made available in 2014 from UC Berkeley~\cite{caffe}. It supports C, C++, Python and MATLAB. Tensorflow was released by Google in 2015, and supports C++ and python; it also supports multiple CPUs and GPUs and has more flexibility than Caffe, with the computation expressed as dataflow graphs to manage the “tensors” (multidimensional arrays).  Another popular framework is Torch, which was developed by Facebook and NYU and supports C, C++ and Lua. There are several other frameworks such as Theano, MXNet, CNTK, which are described in~\cite{nvidia_frameworks}. There are also higher-level libraries that can run on top of the aforementioned frameworks to provide a more universal experience and faster development. One example of such libraries is Keras, which is written in Python and supports Tensorflow, CNTK and Theano.

The existence of such frameworks are not only a convenient aid for DNN researchers and application designers, but they are also invaluable for engineering high performance or more efficient DNN computation engines. In particular, because the frameworks make heavy use of a set primitive operations, such processing of a CONV layer, they can incorporate use of optimized software or hardware accelerators. This acceleration is transparent to the user of the framework. Thus, for example, most frameworks can use Nvidia's cuDNN library for rapid execution on Nvidia GPUs. Similarly, transparent incorporation of dedicated hardware accelerators can be achieved as was done with the Eyeriss chip~\cite{jssc2017-chen}. 

Finally, these frameworks are a valuable source of workloads for hardware researchers. They can be used to drive experimental designs for different workloads, for profiling different workloads and for exploring hardware-software trade-offs.  

\subsection{Models}
Pretrained DNN models can be downloaded from various websites~\cite{caffe_mnist,caffe_model_zoo, matcovnet_models, tensorflow_models} for the various different frameworks. It should be noted that even for the same DNN (e.g., AlexNet) the accuracy of these models can vary by around 1\% to 2\% depending on how the model was trained, and thus the results do not always exactly match the original publication. 

\subsection{Popular Datasets for Classification}
\begin{figure}
    \begin{center}
        \includegraphics[width=0.9\linewidth]{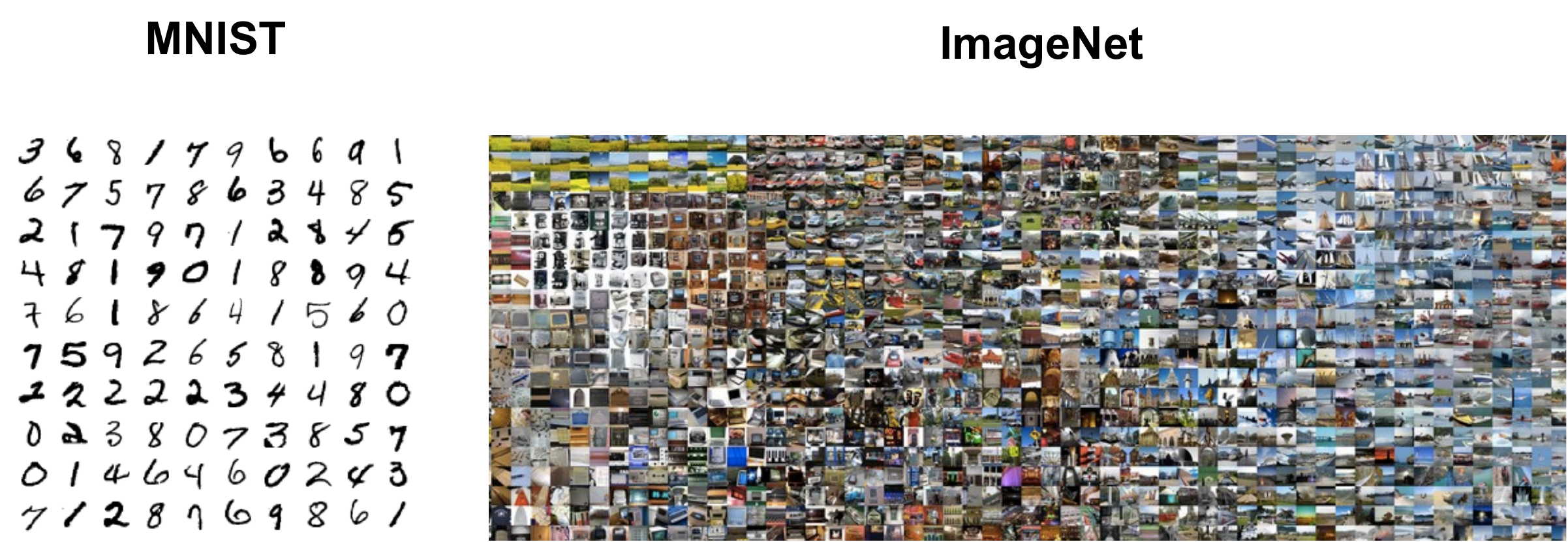}
        \caption{   MNIST (10 classes, 60k training, 10k testing)~\cite{mnist} vs. ImageNet (1000 classes, 1.3M training, 100k testing)\cite{ijcv2015-russakovsky} dataset.
                }
        \label{fig:datasets}
    \end{center}
    \vspace{-10pt}
\end{figure}

It is important to factor in the difficulty of the task when comparing  different DNN models. For instance, the task of classifying handwritten digits from the MNIST dataset~\cite{mnist} is much simpler than classifying an object into one of 1000 classes as is required for the ImageNet dataset~\cite{ijcv2015-russakovsky}(Fig.~\ref{fig:datasets}).  It is expected that the size of the DNNs (i.e., number of weights) and the number of MACs will be larger for the more difficult task than the simpler task and thus require more energy and have lower throughput. For instance, LeNet-5\cite{ieee1998-lecun} is designed for digit classification, while AlexNet\cite{nips2012-krizhevsky}, VGG-16\cite{iclr2015-simonyan}, GoogLeNet\cite{cvpr2015-szegedy}, and ResNet\cite{cvpr2016-he} are designed for the 1000-class image classification.   

There are many AI tasks that come with publicly available datasets in order to evaluate the accuracy of a given DNN. Public datasets are important for comparing the accuracy of different approaches. The simplest and most common task is image classification, which involves being given an entire image, and selecting 1 of N classes that the image most likely belongs to. There is no localization or detection.  
%(e.g. listed on http://rodrigob.github.io/are_we_there_yet/build and https://www.kaggle.com/datasets) 

\emph{MNIST} is a widely used dataset for digit classification that was introduced in 1998~\cite{mnist}. It consists of 28$\times$28 pixel grayscale images of handwritten digits. There are 10 classes (for 10 digits) and 60,000 training images and 10,000 test images. LeNet-5 was able to achieve an accuracy of 99.05\% when MNIST was first introduced.  Since then the accuracy has increased to 99.79\% using regularization of neural networks with dropconnect~\cite{wan2013regularization}. Thus, MNIST is now considered a fairly easy dataset.

\emph{CIFAR} is a dataset that consists of 32$\times$32 pixel colored images of of various objects, which was released in 2009~\cite{cifar}.  CIFAR is a subset of the 80 million Tiny Image dataset~\cite{torralba200880}.  CIFAR-10 is composed of 10 mutually exclusive classes. There are 50,000 training images (5000 per class) and 10,000 test images (1000 per class). A two-layer convolutional deep belief network was able to achieve 64.84\% accuracy on CIFAR-10 when it was first introduced~\cite{krizhevsky2010convolutional}. Since then the accuracy has increased to 96.53\% using fractional max pooling~\cite{graham2014fractional}. 

\emph{ImageNet} is a large scale image dataset that was first introduced in 2010; the dataset stabilized in 2012~\cite{ijcv2015-russakovsky}.  It contains images of 256$\times$256 pixel in color with 1000 classes.  The classes are defined using the WordNet as a backbone to handle ambiguous word meanings and to combine together synonyms into the same object category.  In otherwords, there is a hierarchy for the ImageNet categories. The 1000 classes were selected such that there is no overlap in the ImageNet hierarchy.  The ImageNet dataset contains many fine-grained categories including 120 different breeds of dogs. There are 1.3M training images (732 to 1300 per class), 100,000 testing images (100 per class)  and 50,000 validation images (50 per class).

The accuracy of the ImageNet Challenge are reported using two metrics: Top-5 and Top-1 error.  Top-5 error means that if any of the top five scoring categories are the correct category, it is counted as a correct classification.  The Top-1 requires that the top scoring category be correct. In 2012, the winner of the ImageNet Challenge (AlexNet) was able to achieve an accuracy of 83.6\% for the top-5 (which is substantially better than the 73.8\% which was second place that year that did not use DNNs); it achieved 61.9\% on the top-1 of the validation set. In 2017, the highest accuracy was 97.7\% for the top-5. 

In summary of the various image classification datasets, it is clear that MNIST is a fairly easy dataset, while ImageNet is a challenging one with a wider coverage of classes.  Thus in terms of evaluating the accuracy of a given DNN, it is important to consider that dataset upon which the accuracy is measured.

\subsection{Datasets for Other Tasks}
Since the accuracy of the state-of-the-art DNNs are performing better than human-level accuracy on image classification tasks, the ImageNet Challenge has started to focus on more difficult tasks such as single-object localization and object detection.  For single-object localization, the target object must be localized and classified (out of 1000 classes).  The DNN outputs the top five categories and top five bounding box locations.  There is no penalty for identifying an object that is in the image but not included in the ground truth. For object detection, all objects in the image must be localized and classified (out of 200 classes). The bounding box for all objects in these categories must be labeled. Objects that are not labeled are penalized as are duplicated detections.

Beyond ImageNet, there are also other popular image datasets for computer vision tasks. For object detection, there is the PASCAL VOC (2005-2012) dataset that contains 11k images representing 20 classes (27k object instances, 7k of which has detailed segmentation)~\cite{voc_dataset}.  For object detection, segmentation and recognition in context, there is the MS COCO dataset with 2.5M labeled instances in 328k images (91 object categories)~\cite{mscoco}; compared to ImageNet, COCO has fewer categories but more instances per category, which is useful for precise 2-D localization. COCO also has more labeled instances per image to potentially help with contextual information.

Most recently even larger scale datasets have been made available. For instance, Google has an Open Images dataset with over 9M images~\cite{openimages}, spanning 6000 categories. There is also a YouTube dataset with 8M videos (0.5M hours of video) covering 4800 classes~\cite{youtube8m}. Google also released an audio dataset comprised of 632 audio event classes and a collection of 2M human-labeled 10-second sound clips~\cite{audioset}. These large datasets will be evermore important as DNNs become deeper with more weight parameters to train.

Undoubtedly, both larger datasets and datasets for new domains will serve as important resources for profiling and exploring the efficiency of future DNN engines. 

% VS: decide if want to add (from updated tutorial)
%The development resources presented in this section enable us to evaluate hardware using the appropriate DNN model and dataset. In particular, it's important to realize that difficult tasks typically require larger models; for instance, LeNet would not apply to the ImageNet Challenge.  In addition, different datasets are required for different tasks; for instance, self-driving cars require high-definition video, and thus a network trained on the low resolution ImageNet dataset may not be sufficient.  To address these requirements, the number of datasets growing at a rapid pace.

%% file: architecture.tex
\section{Hardware for DNN Processing}
\label{sec:architecture}
Due to the popularity of DNNs, many recent hardware platforms have special features that target DNN processing. For instance, the Intel Knights Landing CPU features special vector instructions for deep learning; the Nvidia PASCAL GP100 GPU features 16-bit floating point (FP16) arithmetic support to perform two FP16 operations on a single precision core for faster deep learning computation. Systems have also been built specifically for DNN processing such as Nvidia DGX-1 and Facebook's Big Basin custom DNN server~\cite{facebook_system}.  DNN inference has also been demonstrated on various embedded System-on-Chips (SoC) such as Nvidia Tegra and Samsung Exynos as well as FPGAs.  Accordingly, it's important to have a good understanding of how the processing is being performed on these platforms, and how application-specific accelerators can be designed for DNNs for further improvement in throughput and energy efficiency.

\begin{figure}
    \begin{center}
        \includegraphics[width=0.9\linewidth]{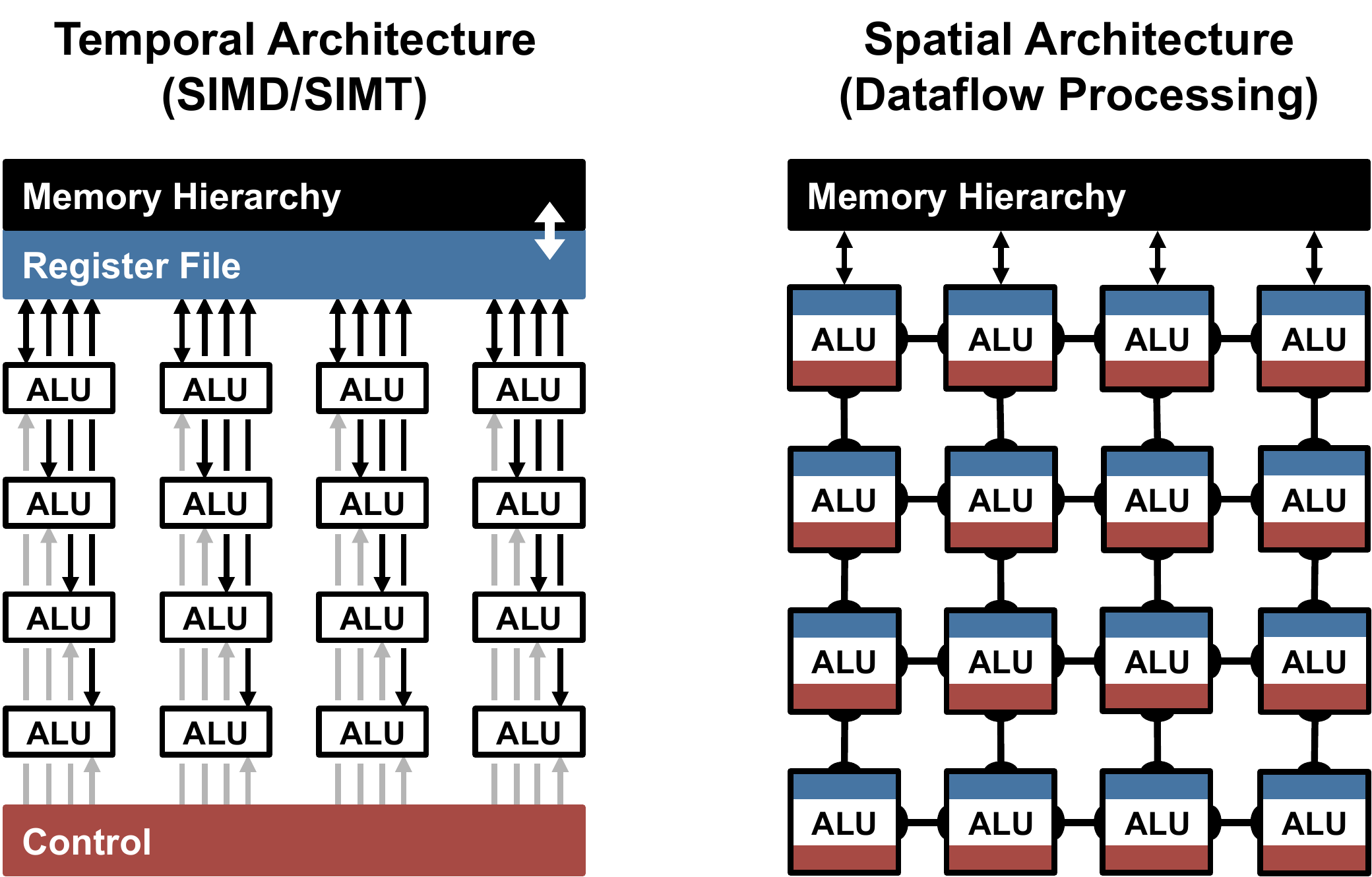}
        \caption{     Highly-parallel compute paradigms.
                }
        \label{fig:parallel_compute}
    \end{center}
\end{figure}

The fundamental component of both the CONV and FC layers are the multiply-and-accumulate (MAC) operations, which can be easily parallelized.  In order to achieve high performance, highly-parallel compute paradigms are very commonly used, including both temporal and spatial architectures as shown in Fig.~\ref{fig:parallel_compute}. The temporal architectures appear mostly in CPUs or GPUs, and employ a variety of techniques to improve parallelism such as vectors (SIMD) or parallel threads (SIMT). Such temporal architecture use a centralized control for a large number of ALUs. These ALUs can only fetch data from the memory hierarchy and cannot communicate directly with each other. In contrast, spatial architectures use dataflow processing, i.e., the ALUs form a processing chain so that they can pass data from one to another directly. Sometimes each ALU can have its own control logic and local memory, called a scratchpad or register file. We refer to the ALU with its own local memory as a processing engine (PE).  Spatial architectures are commonly used for DNNs in ASIC and FPGA-based designs. In this section, we will discuss the different design strategies for efficient processing on these different platforms, without any impact on accuracy (i.e., all approaches in this section produce bit-wise identical results); specifically, 

\begin{itemize}
    \item For \emph{temporal} architectures such as CPUs and GPUs, we will discuss how \emph{computational transforms} on the kernel can reduce the number of multiplications to \emph{increase throughput}.
    \item For \emph{spatial} architectures used in accelerators, we will discuss how \emph{dataflows} can increase data reuse from low cost memories in the memory hierarchy to \emph{reduce energy consumption}.
\end{itemize}

\subsection{Accelerate Kernel Computation on CPU and GPU Platforms}
\label{ssec:kernels}
\begin{figure}
\centering{
        \subfigure[Matrix Vector multiplication is used when computing a single output feature map from a single input feature map.]{
		\includegraphics[width=0.9\linewidth]{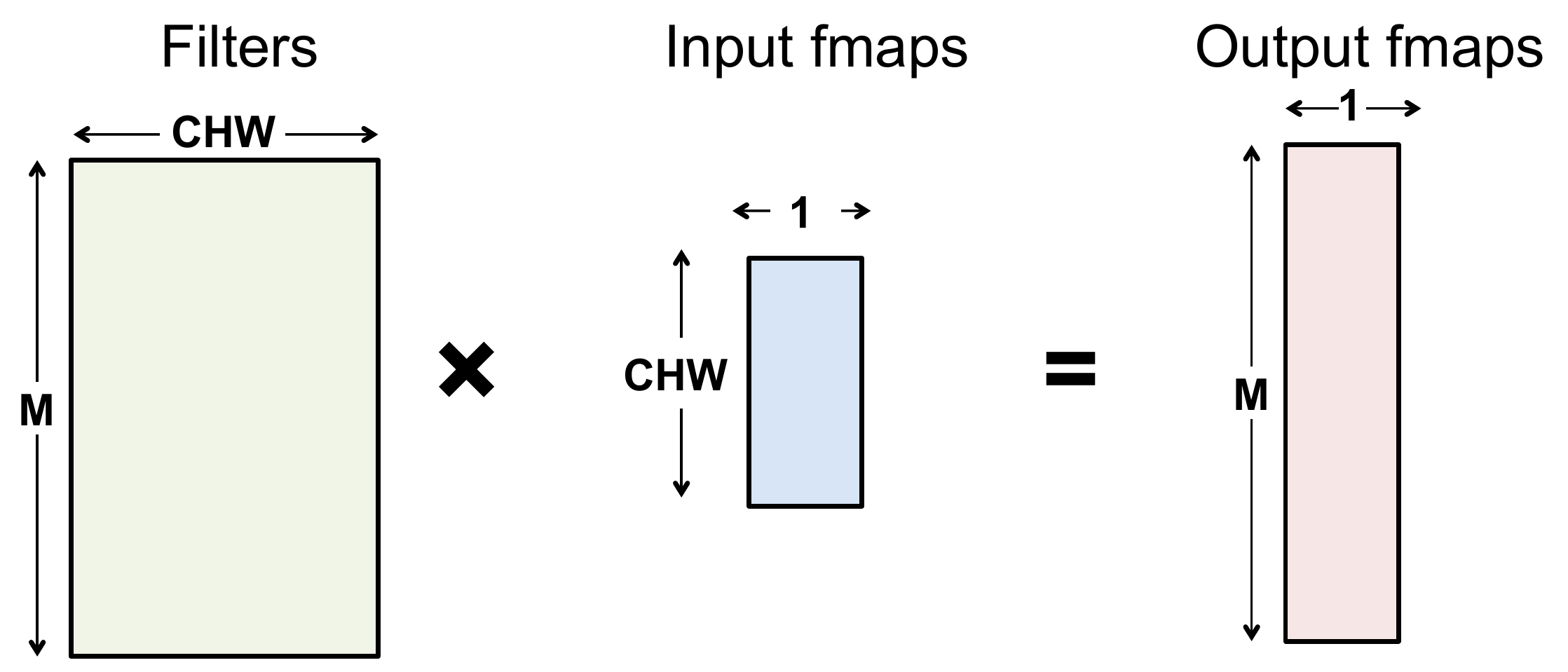}
		\label{fig:matrix_vec_FC}
	}	
    \subfigure[Matrix Multiplications is used when computing $N$ output feature maps from $N$ input feature maps.]{
		\includegraphics[width=0.9\linewidth]{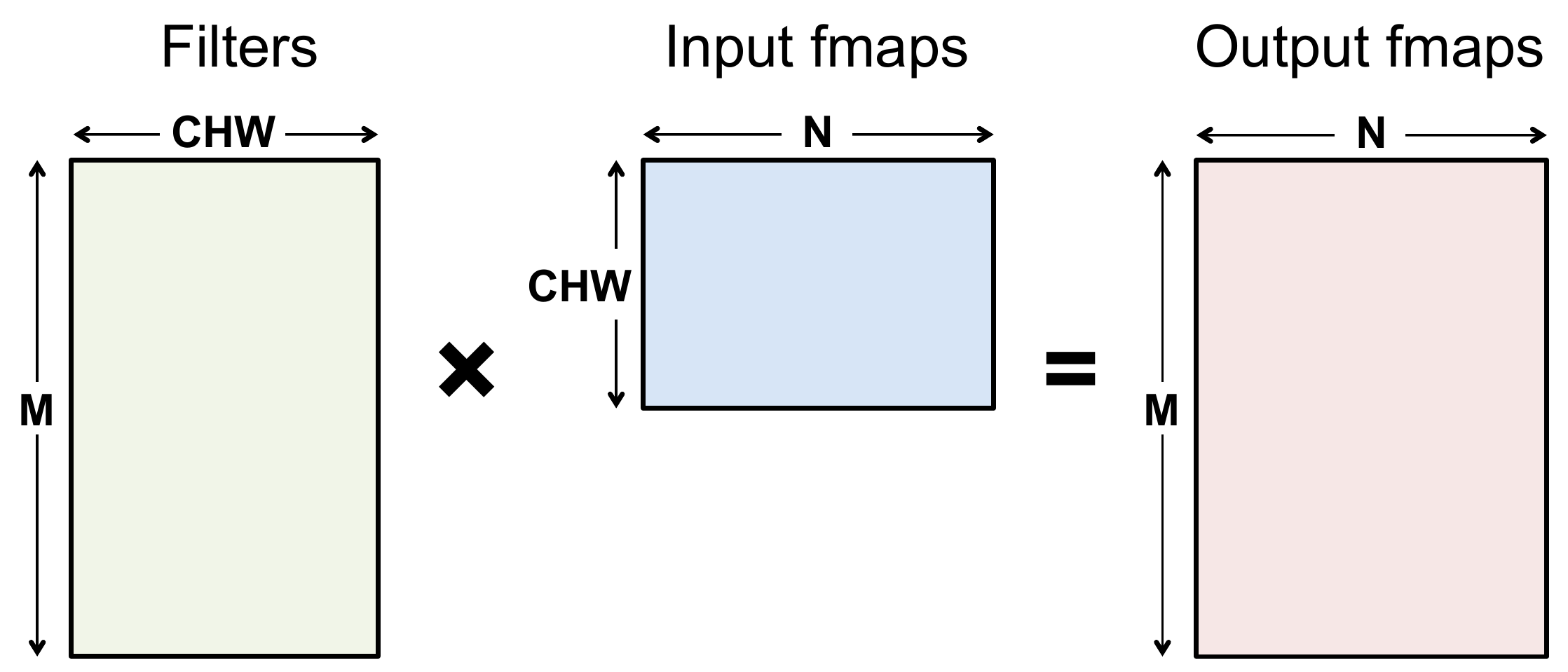}
				\label{fig:matrix_mult_FC}
	}    
}
\caption{Mapping to matrix multiplication for fully connected layers}
\label{fig:matrix}
\end{figure}

CPUs and GPUs use parallelizaton techniques such as SIMD or SIMT to perform the MACs in parallel.  All the ALUs share the same control and memory (register file).  On these platforms, both the FC and CONV layers are often mapped to a matrix multiplication (i.e., the kernel computation).  Fig.~\ref{fig:matrix} shows how a matrix multiplication is used for the FC layer.  The height of the filter matrix is the number of filters and the width is the number of weights per filter (input channels ($C$) $\times$ width ($W$) $\times$ height ($H$), since $R=W$ and $S=H$ in the FC layer); the height of the input feature maps matrix is the number of activations per input feature map ($C$ $\times$ $W$ $\times$ $H$), and the width is the number of input feature maps (one in Fig.~\ref{fig:matrix_vec_FC} and $N$ in Fig.~\ref{fig:matrix_mult_FC}); finally, the height of the output feature map matrix is the number of channels in the output feature maps ($M$), and the width is the number of output feature maps ($N$), where each output feature map of the FC layer has the dimension of 1$\times$1$\times$number of output channels ($M$).
 
\begin{figure}
\centering{
    \subfigure[Mapping convolution to Toeplitz matrix]{
		\includegraphics[width=0.9\linewidth]{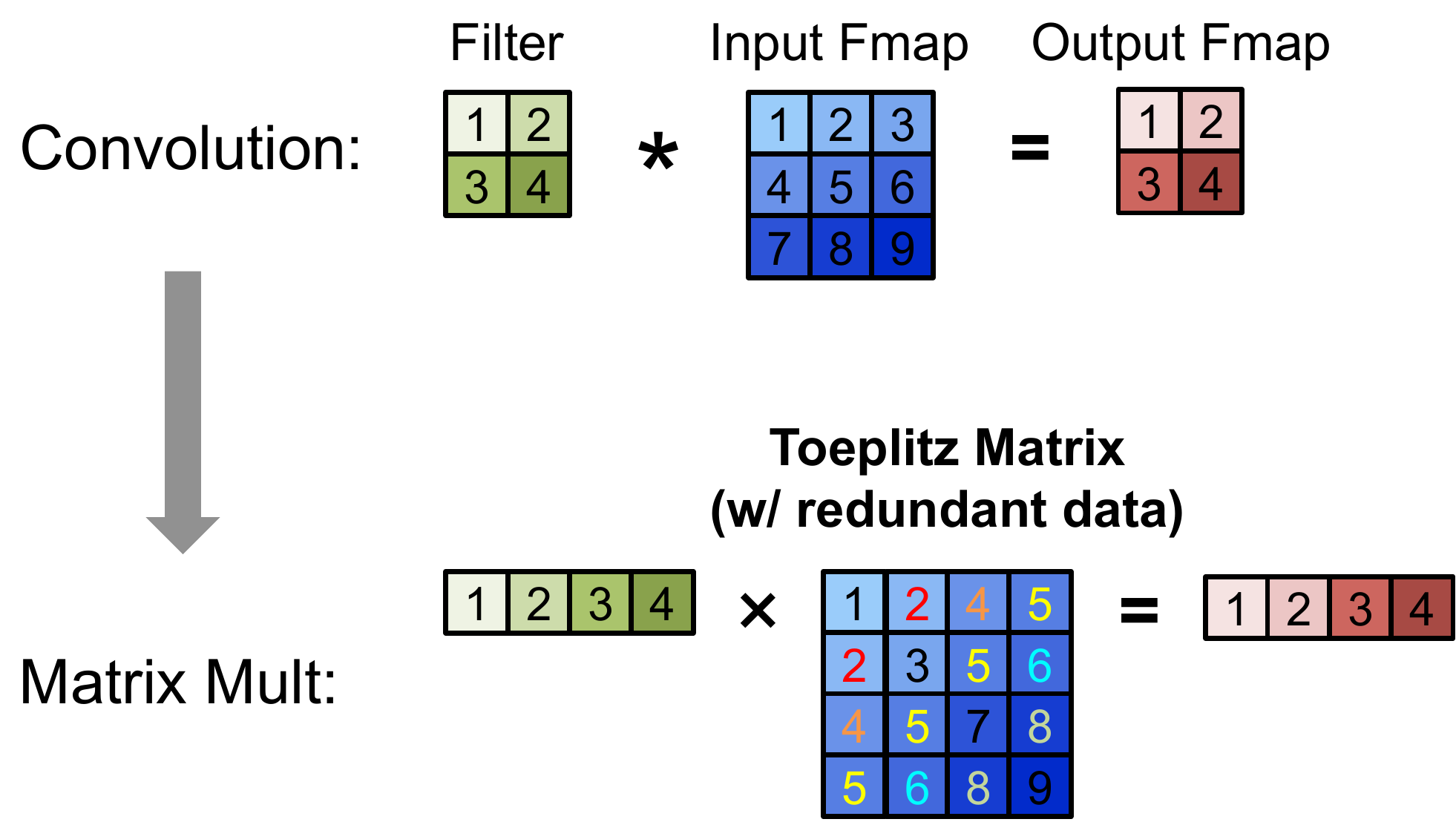}
		\label{fig:toeplitz}
	}	
    \subfigure[Extend Toeplitz matrix to multiple channels and filters]{
		\includegraphics[width=0.9\linewidth]{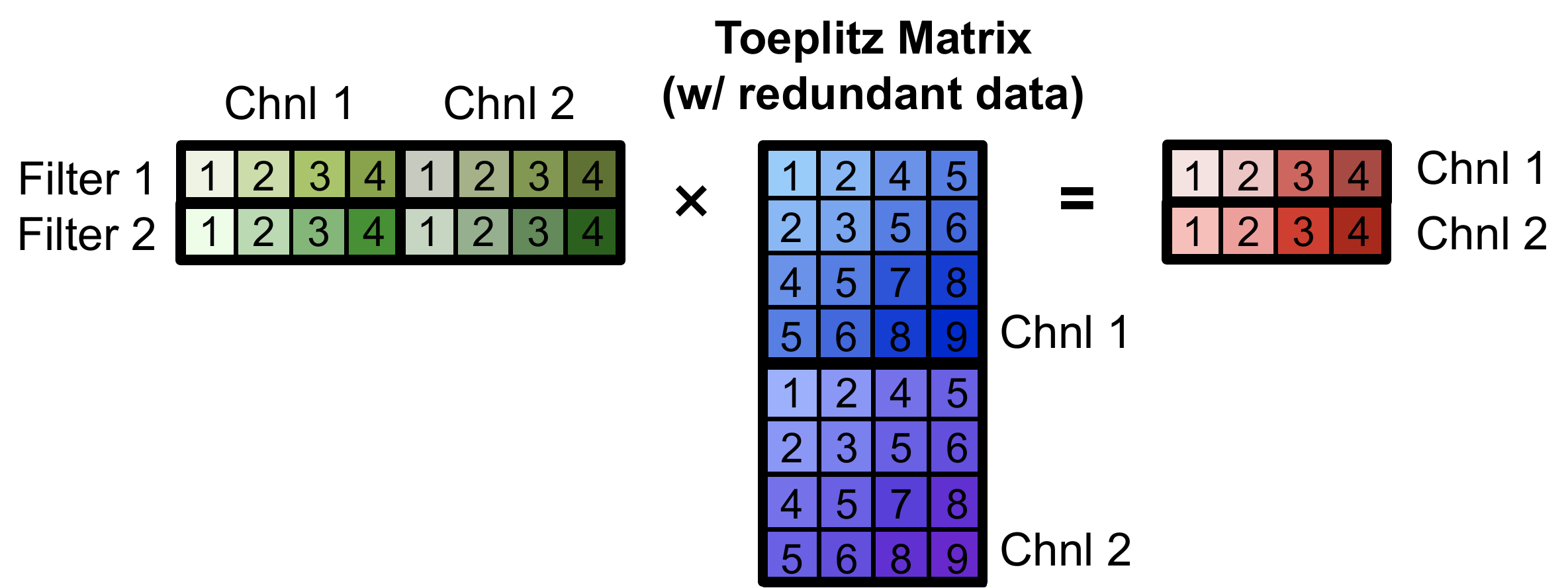}
				\label{fig:toeplitz_multi_channel}
	}
}
\caption{Mapping to matrix multiplication for convolutional layers.}
\label{fig:matrix_conv}
\end{figure}

The CONV layer in a DNN can also be mapped to a matrix multiplication using a relaxed form of the Toeplitz matrix as shown in Fig.~\ref{fig:matrix_conv}. The downside for using matrix multiplication for the CONV layers is that there is redundant data in the input feature map matrix as highlighted in Fig.~\ref{fig:toeplitz}.  This can lead to either inefficiency in storage, or a complex memory access pattern.
 
There are software libraries designed for CPUs (e.g., OpenBLAS, Intel MKL, etc.) and GPUs (e.g., cuBLAS, cuDNN, etc.) that optimize for matrix multiplications.  The matrix multiplication is tiled to the storage hierarchy of these platforms, which are on the order of a few megabytes at the higher levels.

% should we extend to multiple images
The matrix multiplications on these platforms can be further sped up by applying computational transforms to the data to reduce the number of multiplications, while still giving the same bit-wise result. Often this can come at a cost of increased number of additions and a more irregular data access pattern.

%They can be though of an extension of the Gauss Multiplication algorithm, where the computation $(a+bi)(c+di)=(ac-bd)+(bc-ad)i$, which would normally require 4 multiplications and 3 additions, can be transformed such that it can be reduced to 2 multi

\begin{figure}
    \begin{center}
        \includegraphics[width=0.9\linewidth]{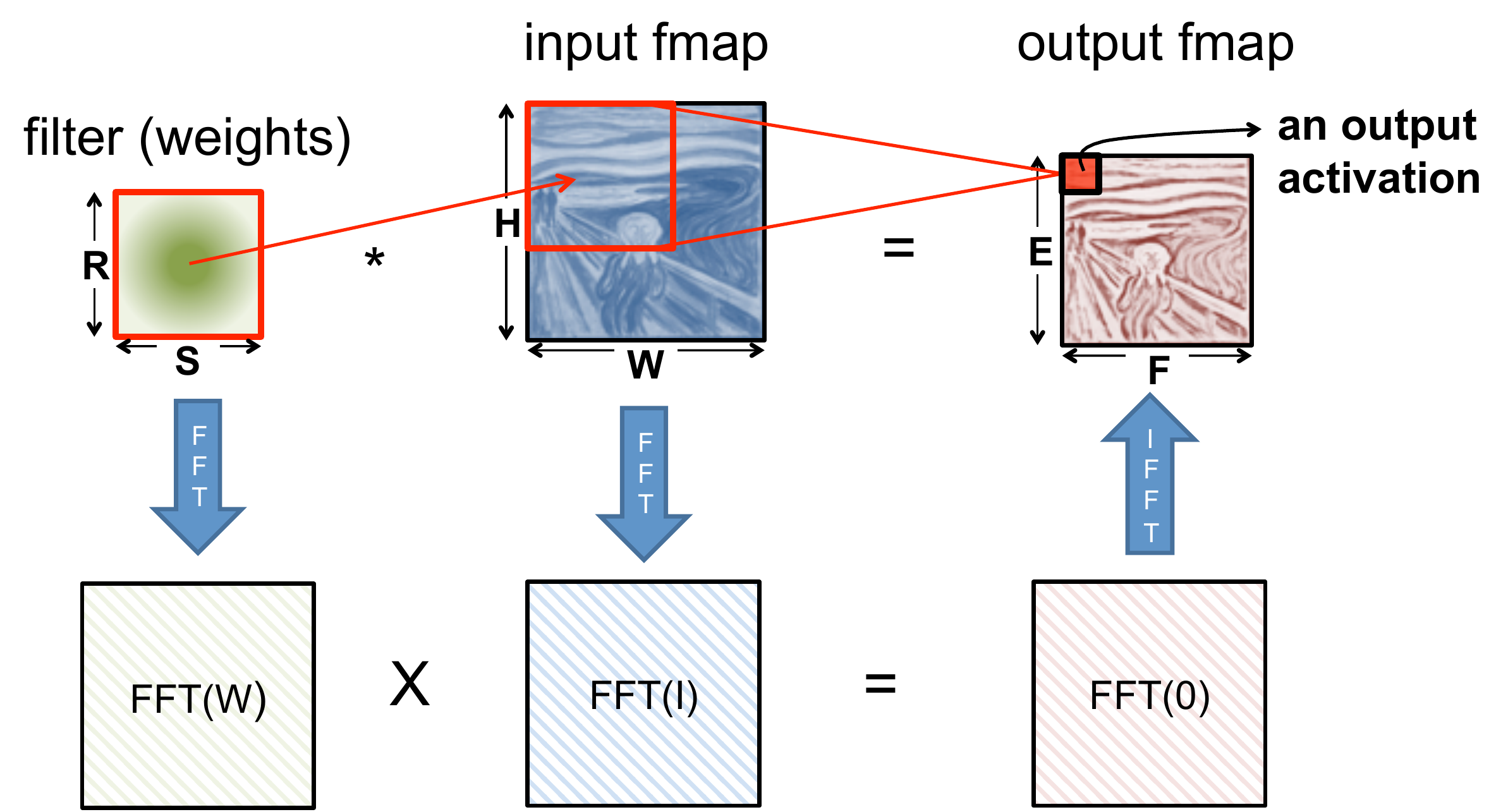}
        \caption{   FFT to accelerate DNN.
                }      
        \label{fig:FFT}
    \end{center}
    \vspace{-10pt}
\end{figure}

Fast Fourier Transform (FFT)~\cite{mathieu2013fast, dubout2012exact} is a well known approach, shown in Fig.~\ref{fig:FFT} that reduces the number of multiplications from O($N_{o}^{2}N_{f}^{2}$) to O($N_{o}^{2}log_{2}N_{o})$, where the output size is $N_{o}\times N_{o}$ and the filter size is $N_{f}\times N_{f}$.  To perform the convolution, we take the FFT of the filter and input feature map, and then perform the multiplication in the frequency domain; we then apply an inverse FFT to the resulting product to recover the output feature map in the spatial domain.   However, there are several drawbacks to using FFT: (1) the benefits of FFTs decrease with filter size; (2) the size of the FFT is dictated by the output feature map size which is often much larger than the filter; (3) the coefficients in the frequency domain are complex.  As a result, while FFT reduces computation, it requires larger storage capacity and bandwidth.  Finally, a popular approach for reducing complexity is to make the weights sparse, which will be discussed in Section~\ref{ssec:pruning}; using FFTs makes it difficult for this sparsity to be exploited.

Several optimizations can be performed on FFT to make it more effective for DNNs.  To reduce the number of operations, the FFT of the filter can be precomputed and stored. In addition, the FFT of the input feature map can be computed once and used to generate multiple channels in the output feature map.  Finally, since an image contains only real values, its Fourier Transform is symmetric and this can be exploited to reduce storage and computation cost.

Other approaches include Strassen~\cite{cong2014minimizing} and Winograd~\cite{lavin2015fast}, which rearrange the computation such that the number of multiplications reduce from O($N^{3}$) to O($N^{2.807}$) and by 2.25$\times$ for a 3$\times$3 filter, respectively, at the cost of reduced numerical stability, increased storage requirements, and specialized processing depending on the size of the filter.  

In practice, different algorithms might be used for different layer shapes and sizes (e.g., FFT for filters greater than 5$\times$5, and Winograd for filters 3$\times$3 and below). Existing platform libraries, such as MKL and cuDNN, dynamically chose the appropriate algorithm for a given shape and size~\cite{mkl2007intel,chetlur2014cudnn}.

%expand with gauss factorization example

\subsection{Energy-Efficient Dataflow for Accelerators}
\begin{figure}
    \begin{center}
        \includegraphics[width=0.9\linewidth]{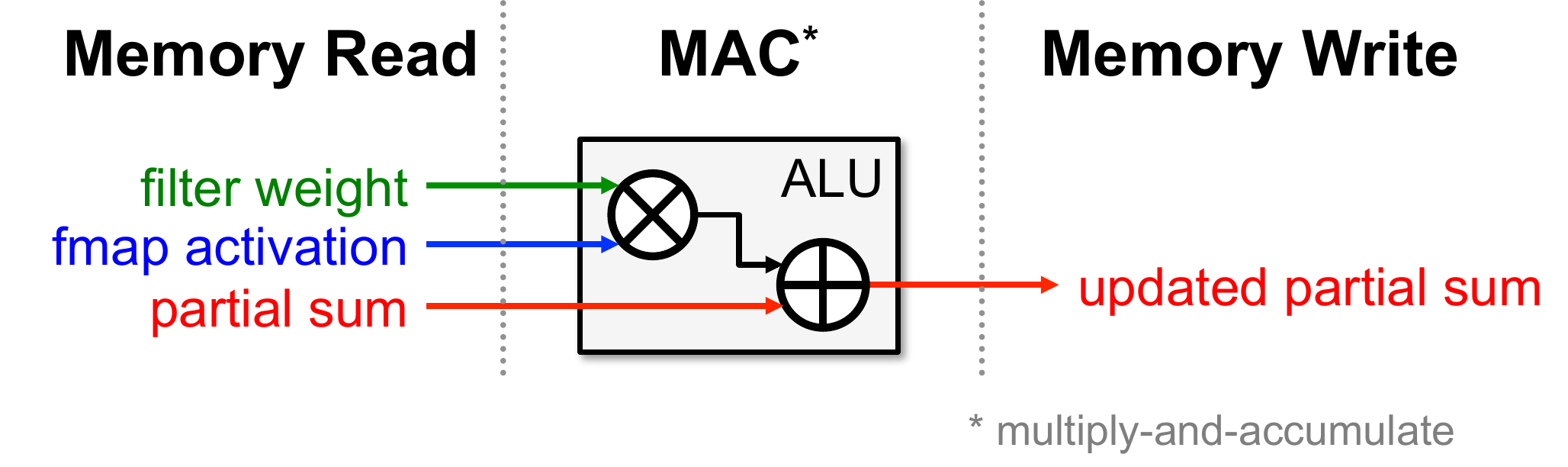}
        \caption{Read and write access per MAC.
                }      
        \label{fig:MAC_reads}
    \end{center}
\end{figure}

For DNNs, the bottleneck for processing is in the memory access. Each MAC requires three memory reads (for filter weight, fmap activation, and partial sum) and one memory write (for the updated partial sum) as shown in Fig.~\ref{fig:MAC_reads}. In the worst case, all of the memory accesses have to go through the off-chip DRAM, which will severely impact both throughput and energy efficiency. For example, in AlexNet, to support its 724M MACs, nearly 3000M DRAM accesses will be required.  Furthermore, DRAM accesses require up to several orders of magnitude higher energy than computation~\cite{isscc2014-horowitz}. 

\begin{figure}
    \begin{center}
        \includegraphics[width=0.9\linewidth]{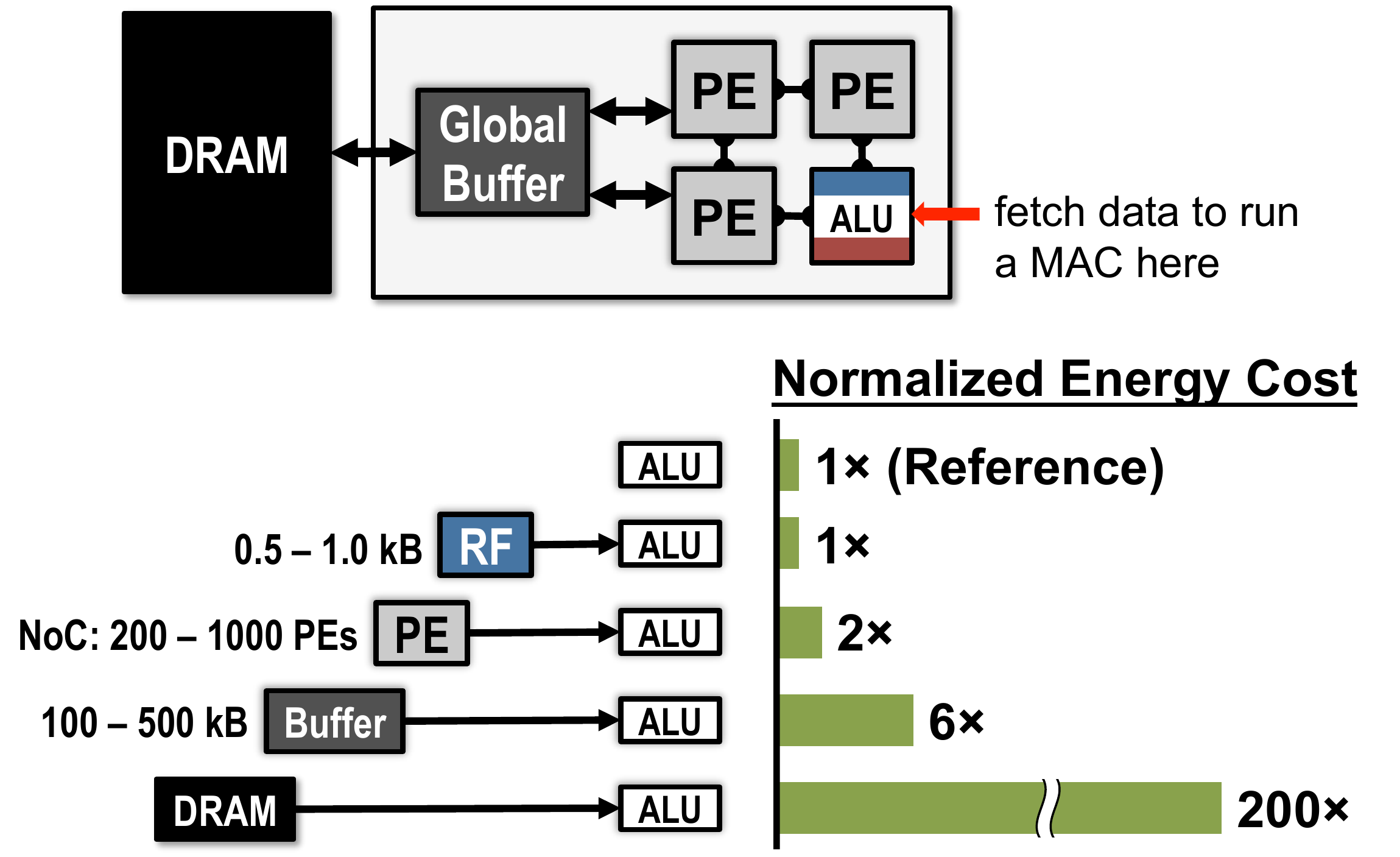}
        \caption{    Memory hierarchy and data movement energy~\cite{isca2016-chen}.
                }      
        \label{fig:memory_hierarchy}
    \end{center}
\end{figure}

Accelerators, such as spatial architectures as shown in Fig.~\ref{fig:parallel_compute}, provide an opportunity to reduce the energy cost of data movement by introducing several levels of local memory hierarchy with different energy cost as shown in Fig.~\ref{fig:memory_hierarchy}.  This includes a large global buffer with a size of several hundred kilobytes that connects to DRAM, an inter-PE network that can pass data directly between the ALUs, and a register file (RF) within each processing element (PE) with a size of a few kilobytes or less.  The multiple levels of memory hierarchy help to improve energy efficiency by providing low-cost data accesses. For example, fetching the data from the RF or neighbor PEs is going to cost 1 or 2 orders of magnitude lower energy than from DRAM. %However, because there’s only a limited amount of capacity in the local memory hierarchy, the hardware needs to exploit data reuse and local accumulation with limited amount of local storage resources and maximize energy efficiency. 

%We are able to come up with a systematic way to analyze this problem by proposing a taxonomy for the existing dataflows. 

Accelerators can be designed to support specialized processing dataflows that leverage this memory hierarchy. The dataflow decides what data gets read into which level of the memory hierarchy and when are they getting processed. Since there is no randomness in the processing of DNNs, it is possible to design a fixed dataflow that can adapt to the DNN shapes and sizes and optimize for the best energy efficiency. The optimized dataflow minimizes access from the more energy consuming levels of the memory hierarchy. Large memories that can store a significant amount of data consume more energy than smaller memories. For instance, DRAM can store gigabytes of data, but consumes two orders of magnitude higher energy per access than a small on-chip memory of a few kilobytes. Thus, every time a piece of data is moved from an expensive level to a lower cost level in terms of energy, we want to reuse that piece of data as much as possible to minimize subsequent accesses to the expensive levels. The challenge, however, is that the storage capacity of these low cost memories are limited. Thus we need to explore different dataflows that maximize reuse under these constraints.

\begin{figure}
    \begin{center}
        \includegraphics[width=0.9\linewidth]{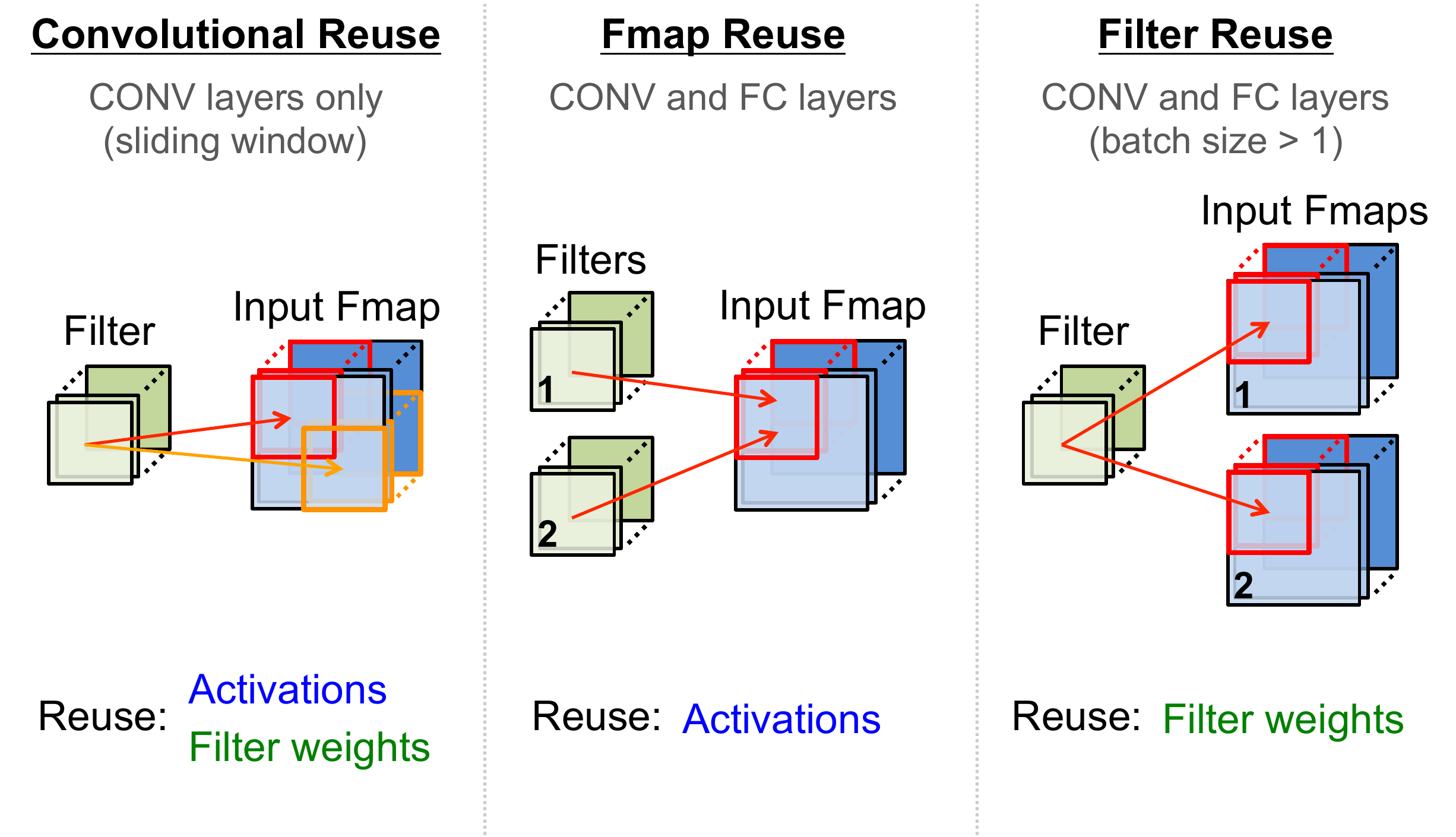}
        \caption{    Data reuse opportunities in DNNs~\cite{isca2016-chen}.
                }      
        \label{fig:reuse_opportunities}
    \end{center}
\end{figure}

For DNNs, we investigate dataflows that exploit three forms of input data reuse (convolutional, feature map and filter) as shown in Fig.~\ref{fig:reuse_opportunities}.  For convolutional reuse, the same input feature map activations and filter weights are used within a given channel, just in different combinations for different weighted sums. For feature map reuse, multiple filters are applied to the same feature map, so the input feature map activations are used multiple times across filters. Finally, for filter reuse, when multiple input feature maps are processed at once (referred to as a batch), the same filter weights are used multiple times across input features maps. 

If we can harness the three types of data reuse by storing the data in the local memory hierarchy and accessing them multiple times without going back to the DRAM, it can save a significant amount of DRAM accesses. For example, in AlexNet, the number of DRAM reads can be reduced by up to 500$\times$ in the CONV layers.  The local memory can also be used for partial sum accumulation, so they do not have to reach DRAM. In the best case, if all data reuse and accumulation can be achieved by the local memory hierarchy, the 3000M DRAM accesses in AlexNet can be reduced to only 61M.
   
%To evaluate and compare different dataflows, we use a spatial architecture with local memory (register file) at each ALU processing element (PE) on the order of 0.5 -- 1.0kB and a shared memory (global buffer) on the order of 100 -- 500kB. The sizes are these memories are selected to be comparable to a typical accelerator for multimedia processing, such as video coding~\cite{sze2014high}.  The global buffer communicates with the off-chip memory (e.g., DRAM).  Data movement is allowed between the PEs using an on-chip network (NoC) to reduce accesses to the global buffer and the off-chip memory.  Three types of data movement include input pixels, filter weights and partial sums (i.e., the product of pixels and weights) that are accumulated for the output.

\begin{figure}
    \begin{center}
        \includegraphics[width=0.97\linewidth]{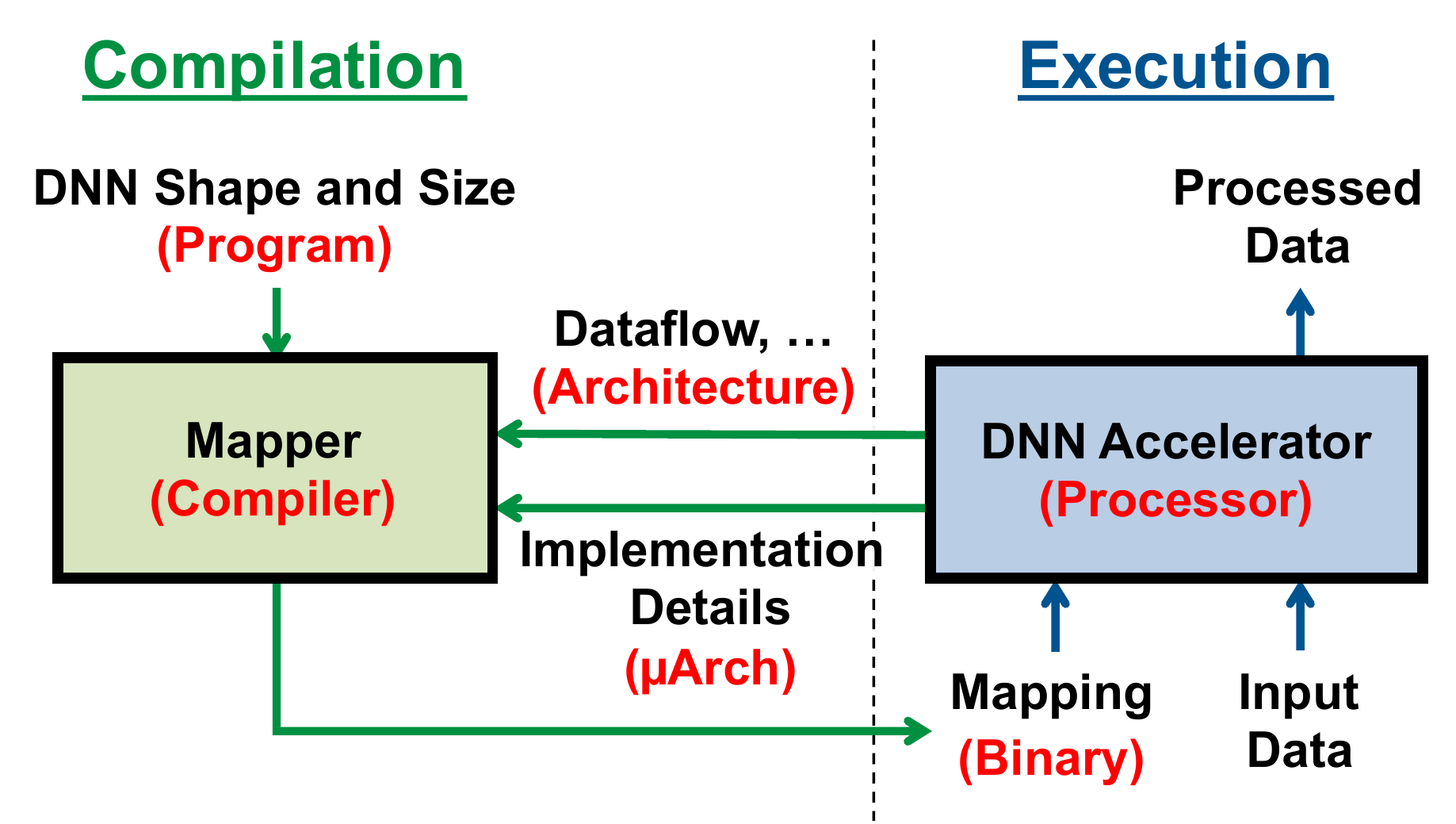}
        % \vspace{-5pt}
        \caption{   An analogy between the operation of DNN accelerators (texts in black) and that of general-purpose processors (texts in red). Figure adopted from~\cite{microTP2017-chen}.
                }
        % \vspace{-20pt}
        \label{fig:analogy}
    \end{center}
\end{figure}

The operation of DNN accelerators is analogous to that of general-purpose processors as illustrated in Fig.~\ref{fig:analogy}~\cite{microTP2017-chen}. In conventional computer systems, the compiler translates the program into machine-readable binary codes for execution given the hardware architecture (e.g., x86 or ARM); in the processing of DNNs, the mapper translates the DNN shape and size into a hardware-compatible computation mapping for execution given the dataflow. While the compiler usually optimizes for performance, the mapper optimizes for energy efficiency.

The following taxonomy (Fig.~\ref{fig:dataflow}) can be used to classify the DNN dataflows in recent works~\cite{asap2009-sankaradas, fpt2010-sriram, isca2010-chakradhar, cvprw2014-gokhale, isscc2015-park, glsvlsi2015-cavigelli, arxiv2015-gupta, isca2015-du, iccd2013-peemen, fpga2015-zhang, asplos2014-chen, micro2014-chen} based on their data handling characteristics~\cite{isca2016-chen}:

\begin{figure}
\centering{
    \subfigure[Weight Stationary]{
		\includegraphics[width=0.92\linewidth]{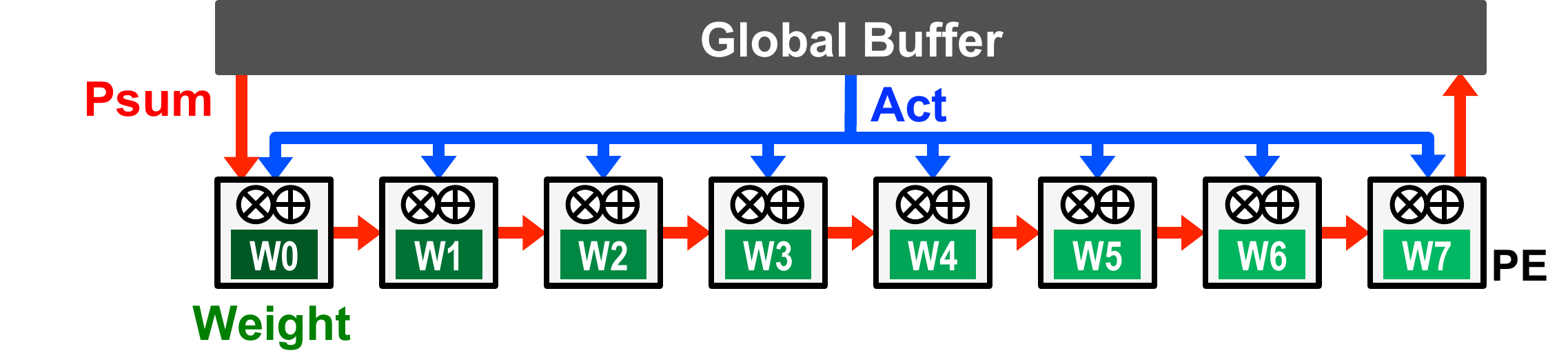}    
		\label{fig:weight_stationary}
	}	
    \subfigure[Output Stationary]{
		\includegraphics[width=0.9\linewidth]{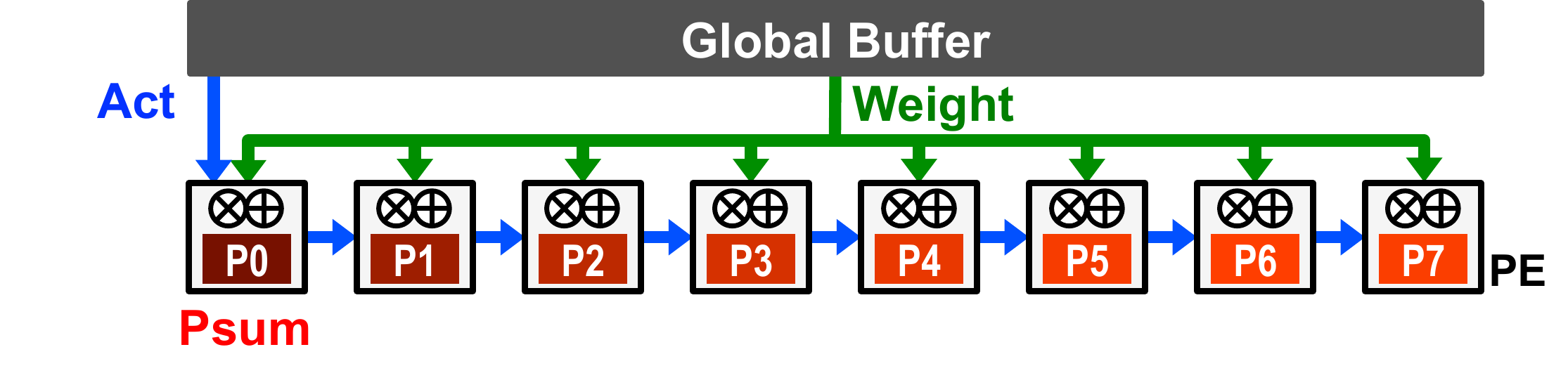}
				\label{fig:output_stationary}
	}
    \subfigure[No Local Reuse]{
		\includegraphics[width=0.9\linewidth]{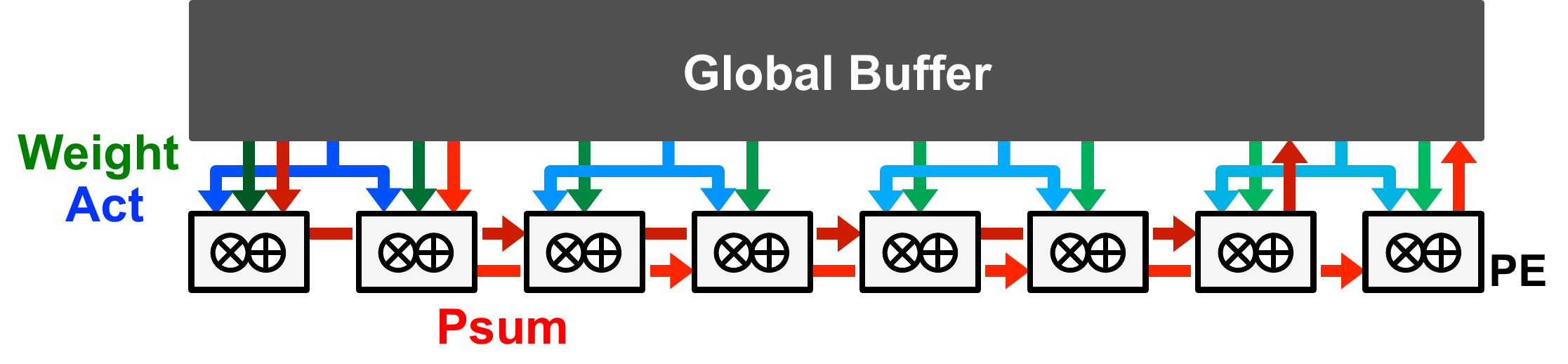}
				\label{fig:no_local_reuse}
	}	
}
\caption{Dataflows for DNNs~\cite{isca2016-chen}.} 
\label{fig:dataflow}
\end{figure}

\subsubsection{Weight stationary (WS)}
The weight stationary dataflow is designed to minimize the energy consumption of reading weights by maximizing the accesses of weights from the register file (RF) at the PE (Fig.~\ref{fig:weight_stationary}). Each weight is read from DRAM into the RF of each PE and stays stationary for further accesses.  The processing runs as many MACs that use the same weight as possible while the weight is present in the RF; it maximizes convolutional and filter reuse of weights. The inputs and partial sums must move through the spatial array and global buffer. The input fmap activations are broadcast to all PEs and then the partial sums are spatially accumulated across the PE array. 

One example of previous work that implement weight stationary dataflow is nn-X, or neuFlow~\cite{cvprw2014-gokhale}, which uses eight 2-D convolution engines for processing a 10$\times$10 filter. There are total 100 MAC units, i.e. PEs, per engine with each PE having a weight that stays stationary for processing. The input fmap activations are broadcast to all MAC units and the partial sums are accumulated across the MAC units. In order to accumulate the partial sums correctly, additional delay storage elements are required, which are counted into the required size of local storage. Other weight stationary examples are found in~\cite{asap2009-sankaradas, fpt2010-sriram, isca2010-chakradhar,  isscc2015-park, glsvlsi2015-cavigelli}.

%At a higher system level, multiple 2-D convolution engines are implemented in order to handle multiple filters and channels. Data routing between the engines are required in order to pass the same fmap activations to multiple engines for processing with multiple filters, or to pass the partial sums from one engine to the other so they can be further accumulated.

\subsubsection{Output stationary (OS)}
The output stationary dataflow is designed to minimize the energy consumption of reading and writing the partial sums (Fig.~\ref{fig:output_stationary}). It keeps the accumulation of partial sums for the same output activation value local in the RF. In order to keep the accumulation of partial sums stationary in the RF, one common implementation is to stream the input activations across the PE array and broadcast the weight to all PEs in the array.

One example that implements the output stationary dataflow is ShiDianNao~\cite{isca2015-du}, where each PE handles the processing for each output activation value by fetching the corresponding input activations from neighboring PEs. The PE array implements dedicated networks to pass data horizontally and vertically. Each PE also has data delay registers to keep data around for the required amount of cycles. At the system level, the global buffer streams the input activations and broadcasts the weights into the PE array. The partial sums are accumulated inside each PE and then get streamed out back to the global buffer.  Other examples of output stationary are found in~\cite{arxiv2015-gupta, iccd2013-peemen}.  

\begin{figure}
    \begin{center}
        \includegraphics[width=0.9\linewidth]{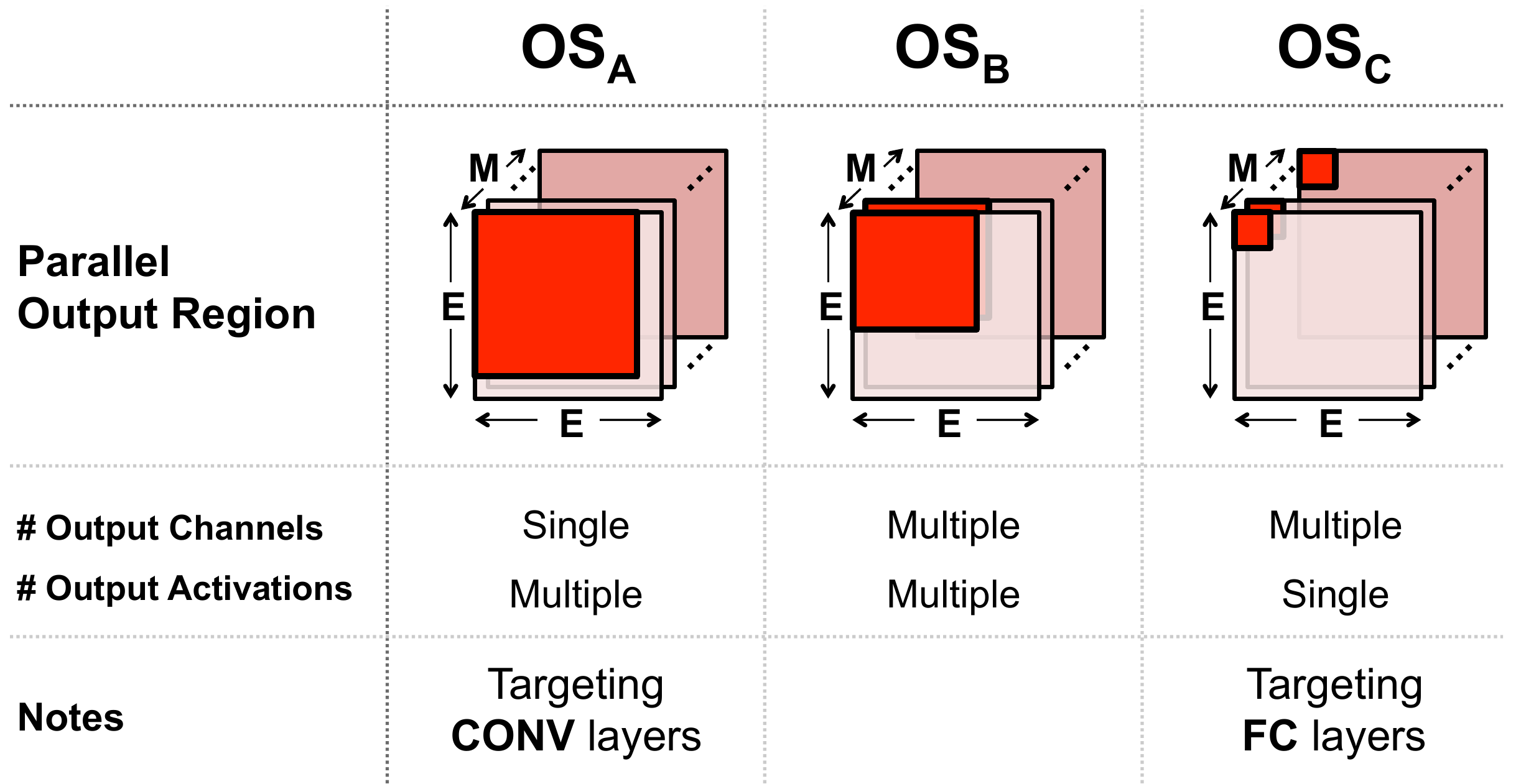}
        \caption{     Variations of output stationary~\cite{isca2016-chen}. 
                }
        \label{fig:os_options}
    \end{center}
\end{figure}

There are multiple possible variants of output stationary as shown in Fig.~\ref{fig:os_options} since the output activations that get processed at the same time can come from different dimensions. For example, the variant $OS_{A}$ targets the processing of CONV layers, and therefore focuses on the processing of output activations from the same channel at a time in order to maximize data reuse opportunities. The variant $OS_{C}$ targets the processing of FC layers, and focuses on generating output activations from all different channels, since each channel only has one output activation. The variant $OS_{B}$ is something in between  $OS_{A}$ and $OS_{C}$. Example of variants $OS_{A}$, $OS_{B}$, and $OS_{C}$ are~\cite{isca2015-du},~\cite{arxiv2015-gupta}, and~\cite{iccd2013-peemen}, respectively.

\subsubsection{No local reuse (NLR)}
While small register files are efficient in terms of energy (pJ/bit), they are inefficient in terms of area ($\mu m^{2}$/bit).  In order to maximize the storage capacity, and minimize the off-chip memory bandwidth, no local storage is allocated to the PE and instead all that area is allocated to the global buffer to increase its capacity (Fig.~\ref{fig:no_local_reuse}).  The no local reuse dataflow differs from the previous dataflows in that nothing stays stationary inside the PE array.  As a result, there will be increased traffic on the spatial array and to the global buffer for all data types. Specifically, it has to multicast the activations, single-cast the filter weights, and then spatially accumulate the partial sums across the PE array. 

In an example of the no local reuse dataflow from UCLA~\cite{fpga2015-zhang}, the filter weights and input activations are read from the global buffer, processed by the MAC units with custom adder trees that can complete the accumulation in a single cycle, and the resulting partial sums or output activations are then put back to the global buffer. Another example is DianNao~\cite{asplos2014-chen}, which also reads input activations and filter weights from the buffer, and processes them through the MAC units with custom adder trees. However, DianNao implements specialized registers to keep the partial sums in the PE array, which helps to further reduce the energy consumption of accessing partial sums. Another example of no local reuse dataflow is found in~\cite{micro2014-chen}.

\subsubsection{Row stationary (RS)}
A row stationary dataflow is proposed in~\cite{isca2016-chen}, which aims to maximize the reuse and accumulation at the RF level for \emph{all} types of data (weights, pixels, partial sums) for the overall energy efficiency. This differs from WS or OS dataflows, which optimize for only weights and partial sums, respectively. 

%A row of the filter weights remains stationary within a PE to exploit  Multiple 1-D rows are combined in the spatial array to exhaustively exploit all convolutional reuse (Fig.~\ref{fig:row_stationary_array}), which reduces accesses to the global buffer.  Multiple 1-D rows from different channels and filters are mapped to each PE to reduce partial sum data movement and exploit filter reuse, respectively.  Finally, multiple passes across the spatial array allow for additional image and filter reuse using the global buffer. This dataflow is demonstrated in~\cite{isscc2016-chen}.

\begin{figure}
\centering{
    \subfigure[Step 1]{
		\includegraphics[width=0.28\linewidth]{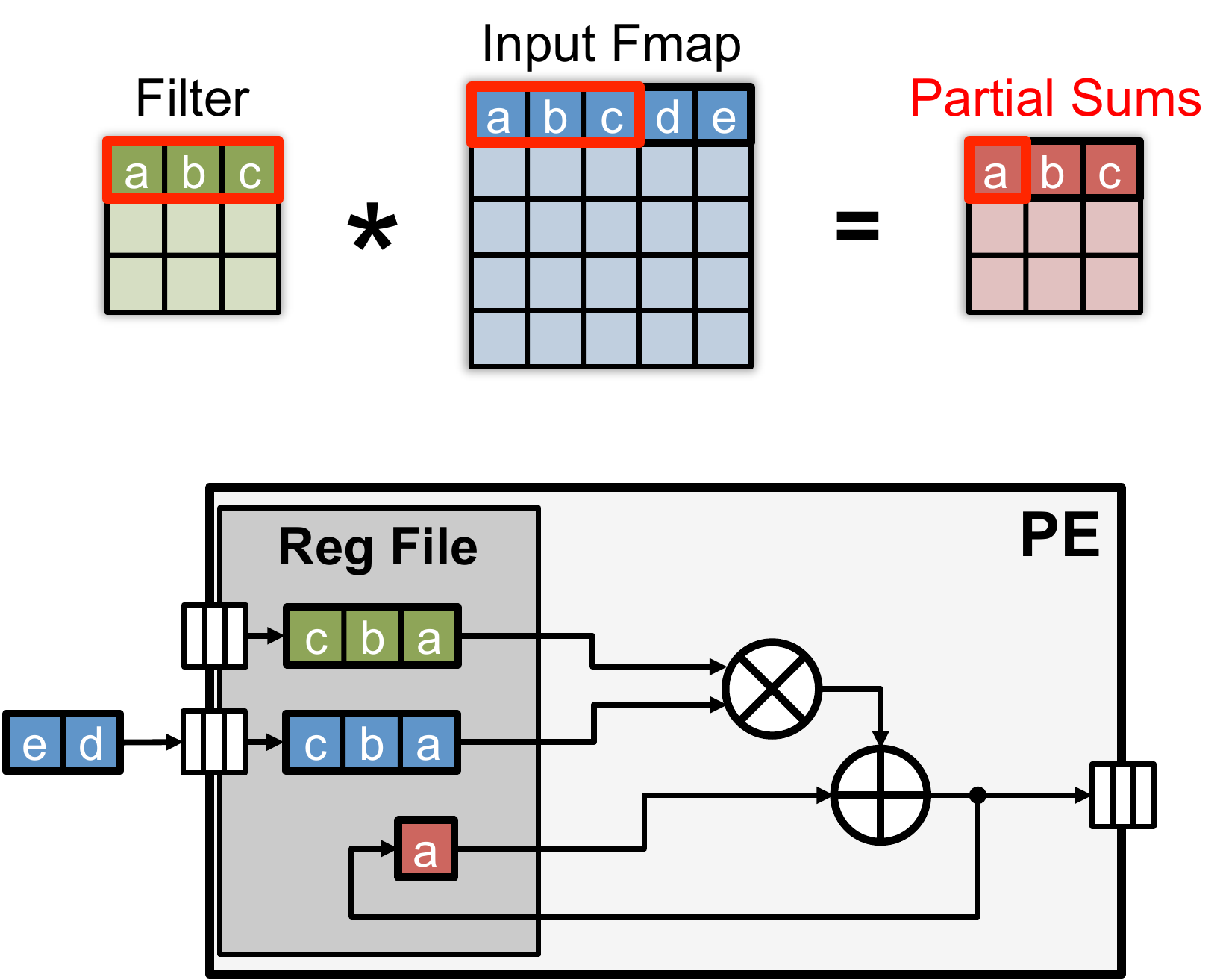}
		\label{fig:reuse_pe1}
	}	
    \subfigure[Step 2]{
		\includegraphics[width=0.28\linewidth]{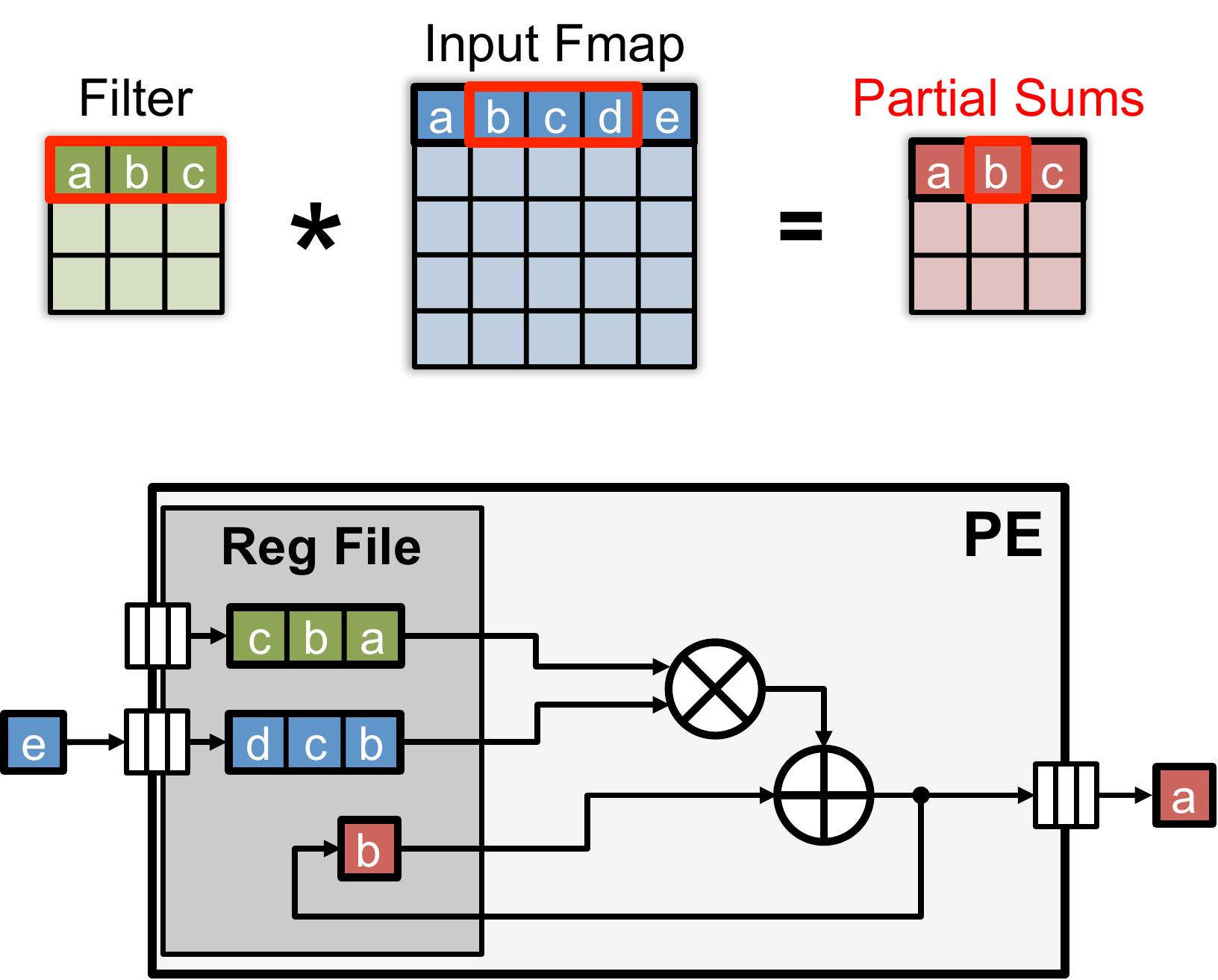}
		\label{fig:reuse_pe2}
	}
    \subfigure[Step 3]{
		\includegraphics[width=0.28\linewidth]{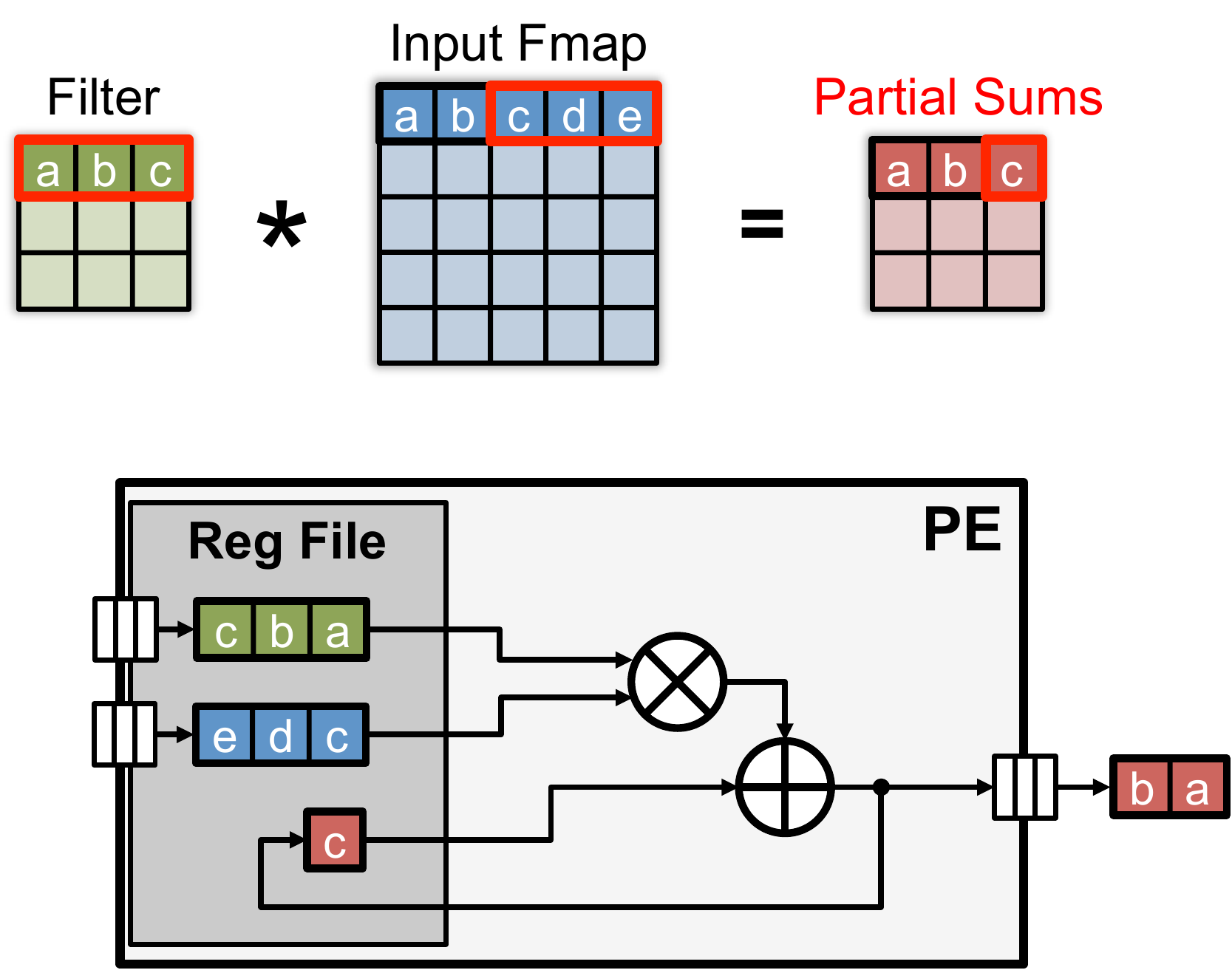}
		\label{fig:reuse_pe3}
	}	
}
\caption{1-D Convolutional reuse within PE for Row Stationary Dataflow~\cite{isca2016-chen}.}
\label{fig:row_stationary_PE}
\end{figure}

The row stationary dataflow assigns the processing of a 1-D row convolution into each PE for processing as shown in Fig.~\ref{fig:row_stationary_PE}. It keeps the row of filter weights stationary inside the RF of the PE and then streams the input activations into the PE. The PE does the MACs for each sliding window at a time, which uses just one memory space for the accumulation of partial sums. Since there are overlaps of input activations between different sliding windows, the input activations can then be kept in the RF and get reused. By going through all the sliding windows in the row, it completes the 1-D convolution and maximize the data reuse and local accumulation of data in this row.

\begin{figure}
    \begin{center}
        \includegraphics[width=0.9\linewidth]{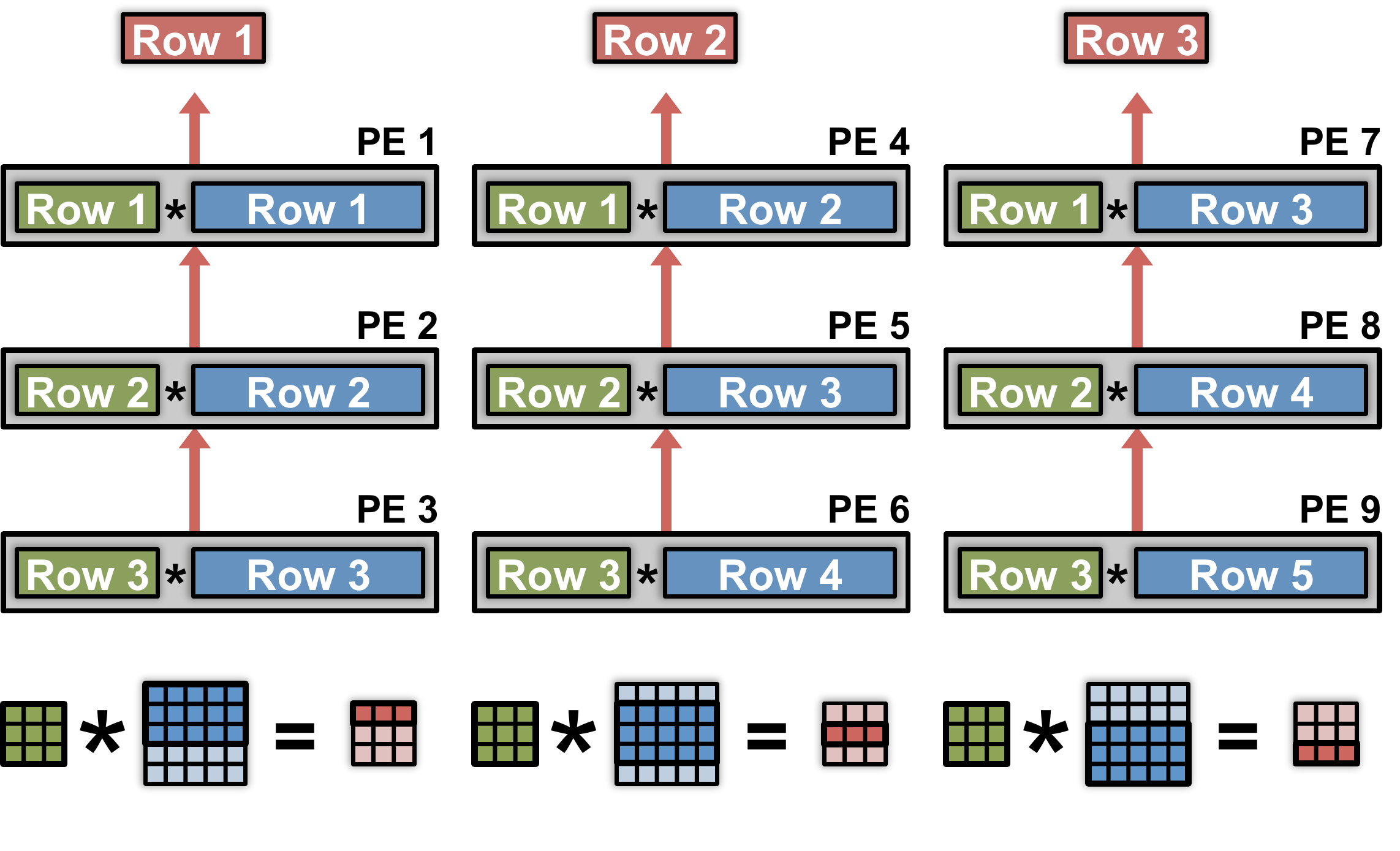}
\caption{2-D convolutional reuse within spatial array for Row Stationary Dataflow~\cite{isca2016-chen}.}
\label{fig:row_stationary_array}
    \end{center}
\end{figure}

With each PE processing a 1-D convolution, multiple PEs can be aggregated to complete the 2-D convolution as shown in Fig.~\ref{fig:row_stationary_array}. For example, to generate the first row of output activations with a filter having three rows, three 1-D convolutions are required. Therefore, we can use three PEs in a column, each running one of the three 1-D convolutions. The partial sums are further accumulated vertically across the three PEs to generate the first output row. To generate the second row of output, we use another column of PEs, where three rows of input activations are shifted down by one row, and use the same rows of filters to perform the three 1-D convolutions. Additional columns of PEs are added until all rows of the output are completed (i.e., the number of PE columns equals the number of output rows).  

This 2-D array of PEs enables other forms of reuse to reduce accesses to the more expensive global buffer. For example, each filter row is reused across multiple PEs horizontally. Each row of input activations is reused across multiple PEs diagonally. And each row of partial sums are further accumulated across the PEs vertically. Therefore, 2-D convolutional data reuse and accumulation are maximized inside the 2-D PE array.

\begin{figure}
    \begin{center}
        \includegraphics[width=0.9\linewidth]{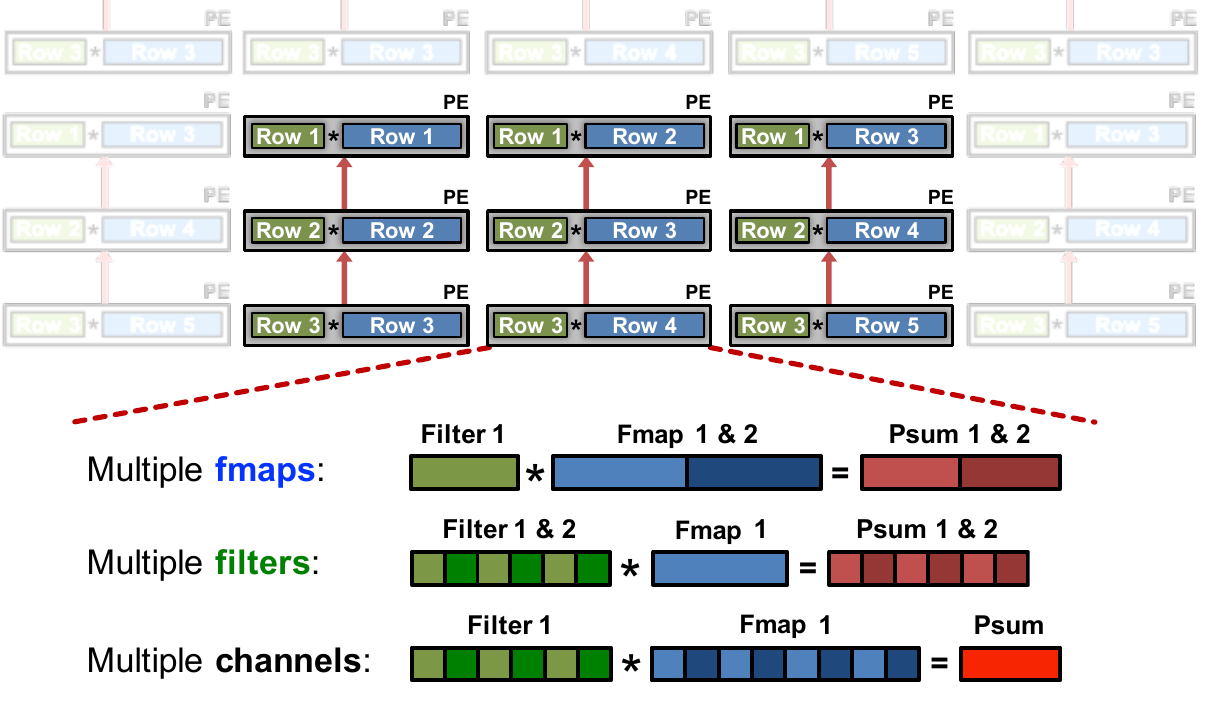}
\caption{Multiple rows of different input feature maps, filters and channels are mapped to same PE within array for additional reuse in the Row Stationary Dataflow~\cite{isca2016-chen}.}
\label{fig:row_stationary_high}
    \end{center}
\end{figure}

To address the high-dimensional convolution of the CONV layer (i.e., multiple fmaps, filters, and channels), multiple rows can be mapped onto the same PE as shown in Fig.~\ref{fig:row_stationary_high}. The 2-D convolution is mapped to a set of PEs, and the additional dimensions are handled by interleaving or concatenating the additional data.  For filter reuse within the PE, different rows of fmaps are concatenated and run through the same PE as a 1-D convolution. For input fmap reuse within the PE, different filter rows are interleaved and run through the same PE as a 1-D convolution. Finally, to increase local partial sum accumulation within the PE, filter rows and fmap rows from different channels are interleaved, and run through the same PE as a 1-D convolution. The partial sums from different channels then naturally get accumulated inside the PE. 

%In the processing of a CONV layer, there are 3 additional dimensions left beyond the 2-D convolution, including multiple fmaps, filters, and channels. The challenge is to handle these dimensions while still achieving data reuse and local accumulation.

%In the case of having multiple fmaps, different fmap rows will run through the same row of filter weights. In order to achieve filter reuse, the RS dataflow simple concatenate different rows of fmaps and run through them as a 1-D convolution. 

%In the case of having multiple filters, the same fmap row is shared across the filters. The fmap activations can be reused in the PE by interleaving the filter rows and run it with the fmap row as a 1-D convolution. 

%In the case when there are multiple channels, the partial sums from different 1-D convolutions can be further accumulated together. In order to achieve partial sum accumulation inside the PE, the RS dataflow interleaves the filter rows and fmap rows from different channels, and run it through the same PE as a 1-D convolution. The partial sums from different channels then naturally get accumulated inside the PE.

\begin{figure}
    \begin{center}
        \includegraphics[width=0.9\linewidth]{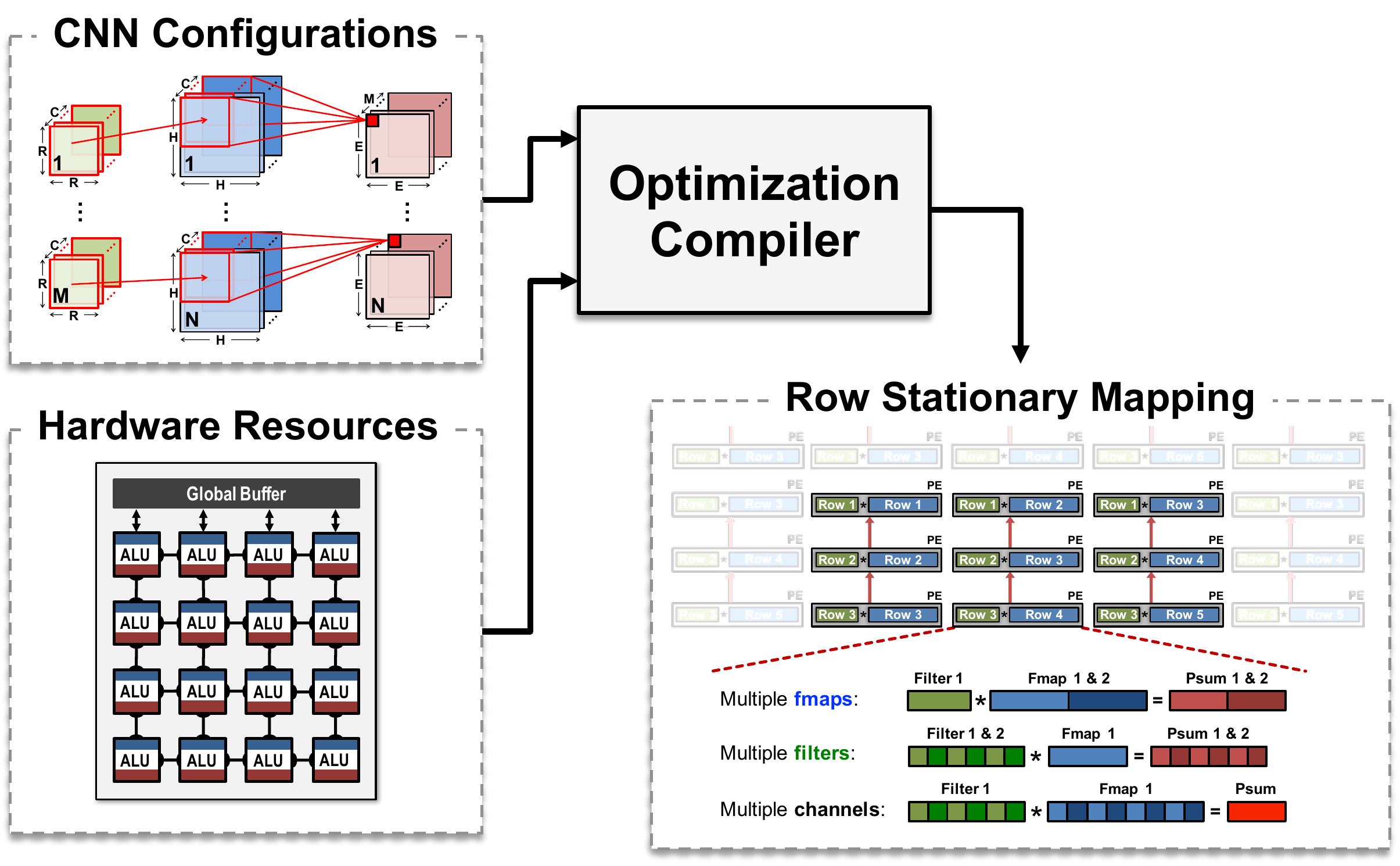}
\caption{Mapping optimization takes in hardware and DNNs shape constraints to determine optimal energy dataflow~\cite{isca2016-chen}.}
\label{fig:mapping_optimization}
    \end{center}
\end{figure}

The number of filters, channels, and fmaps that can be processed at the same time is programmable, and there exists an optimal mapping for the best energy efficiency, which depends on the shape configuration of the DNN as well as the hardware resources provided, e.g., the number of PEs and the size of the memory in the hierarchy. Since all of the variables are known before runtime, it is possible to build a compiler (i.e., mapper) to perform this optimization off-line to configure the hardware for different mappings of the RS dataflow for different DNNs as shown in Fig.~\ref{fig:mapping_optimization}.

\begin{figure}
    \begin{center}
        \includegraphics[width=0.9\linewidth]{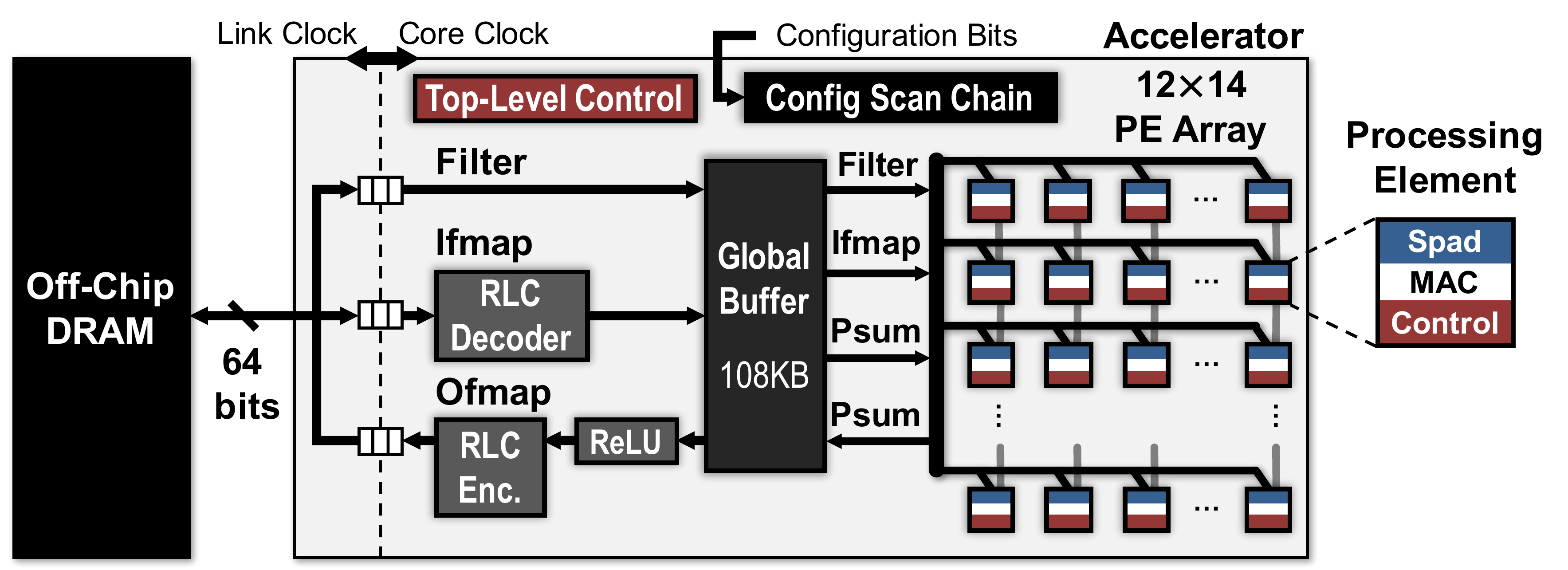}
        \caption{    Eyeriss DNN accelerator~\cite{isscc2016-chen}.
                }      
        \label{fig:eyeriss}
    \end{center}
\end{figure}

One example that implements the row stationary dataflow is Eyeriss~\cite{isscc2016-chen}. It consists of a 14$\times$12 PE array, a 108KB global buffer, ReLU and fmap compression units as shown in Fig.~\ref{fig:eyeriss}. The chip communicates with the off-chip DRAM using a 64-bit bidirectional data bus to fetch data into the global buffer. The global buffer then streams the data into the PE array for processing.

In order to support the RS dataflow, two problems need to be solved in the hardware design. First, how can the fixed-size PE array accommodate different layer shapes? Second, although the data will be passed in a very specific pattern, it still changes with different shape configurations. How can the fixed design pass data in different patterns?

\begin{figure}
    \begin{center}
        \includegraphics[width=0.9\linewidth]{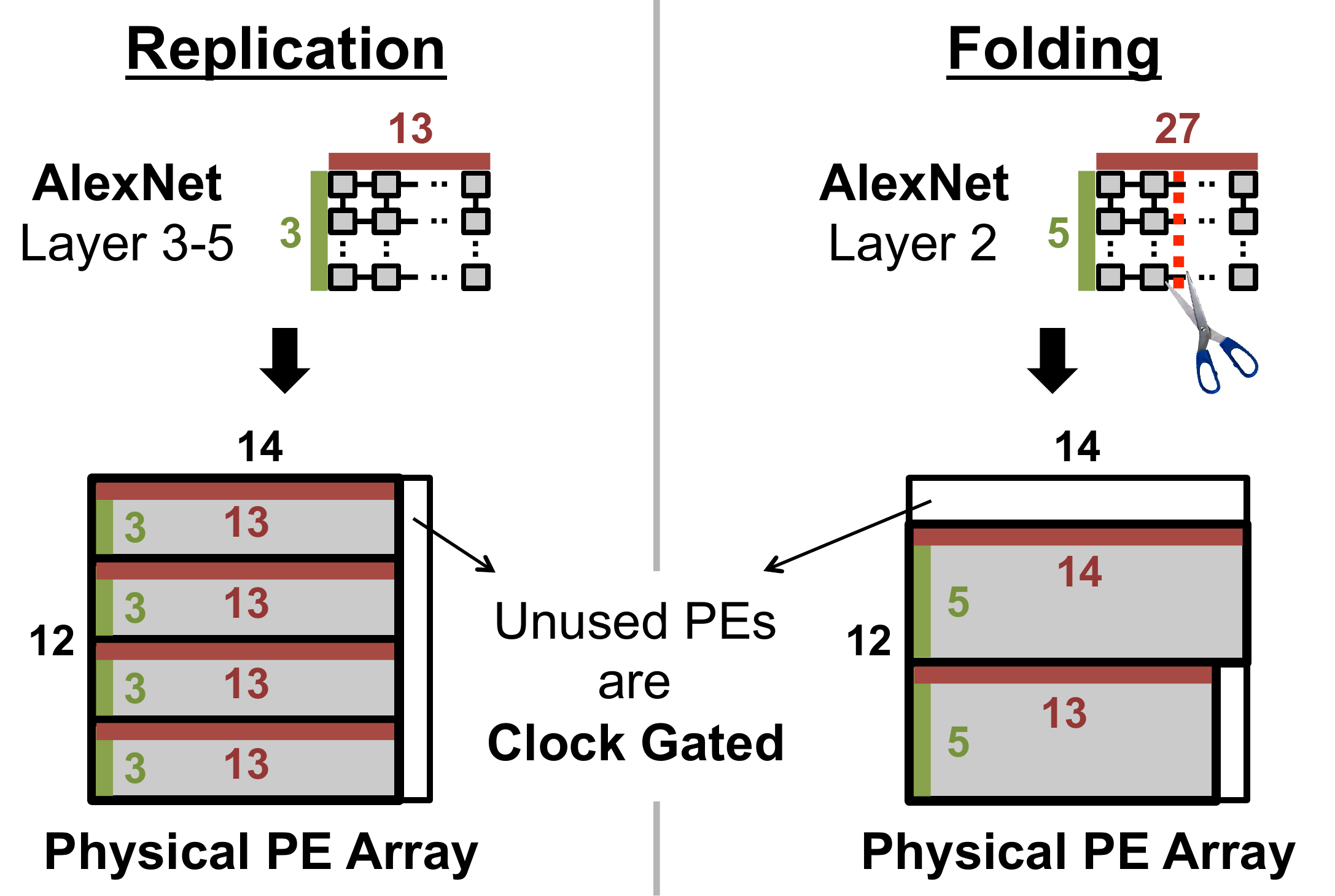}
        \caption{    Mapping uses replication and folding to maximized utilization of PE array~\cite{isscc2016-chen}.
                }      
        \label{fig:mapping}
    \end{center}
\end{figure}

Two mapping strategies can be used to solve the first problem as shown in Fig.~\ref{fig:mapping}. First, replication can be used to map shapes that do not use up the entire PE array. For example, in the third to fifth layers of AlexNet, each 2-D convolution only uses a 13$\times$3 PE array. This structure is then replicated four times, and runs different channels and filters in each replication. The second strategy is called folding. For example, in the second layer of AlexNet, it requires a 27$\times$5 PE array to complete the 2-D convolution. In order to fit it into the 14$\times$12 physical PE array, it is folded into two parts, 14$\times$5 and 13$\times$5, and each are  vertically mapped into the physical PE array. Since not all PEs are used by the mapping, the unused PEs can be clock gated to save energy consumption.

A custom multicast network is used to solve the second problem about flexible data delivery. The simplest way to pass data to multiple destinations is to broadcast the data to all PEs and let each PE decide if it has to process the data or not. However, it is not very energy efficient especially when the size of PE array is large. Instead, a multicast network is used to send data to only the places where it is needed. %This is achieved by using the multicast controllers on the network paths that only pass data when there are destinations that require the data downstream. To determine which data to pass through each controller, each data is sent from the global buffer to the network with a tag value. Each multicast controller is also configured with an ID off-line. The controller then check if the tag matches its local ID to determine the passing of data.

\subsubsection{Energy comparison of different dataflows}

To evaluate and compare different dataflows, the same total hardware area and number of PEs (256) are used in the simulation of a spatial architecture for all dataflows. The local memory (register file) at each processing element (PE) is on the order of 0.5 -- 1.0kB and a shared memory (global buffer) is on the order of 100 -- 500kB. The sizes of these memories are selected to be comparable to a typical accelerator for multimedia processing, such as video coding~\cite{sze2014high}. The memory sizes are further adjusted for the needs of each dataflow under the same area constraint. For example, since the no local reuse dataflow does not require any RF in PE, it is allocated with a much larger global buffer. The simulation uses the layer configurations from AlexNet with a batch size of 16. The simulation also takes into account the fact that accessing different levels of the memory hierarchy requires different energy cost.

%The global buffer communicates with the off-chip memory (e.g., DRAM).  Data movement is allowed between the PEs using an on-chip network (NoC) to reduce accesses to the global buffer and the off-chip memory.  Three types of data movement include input pixels, filter weights and partial sums (i.e., the product of pixels and weights) that are accumulated for the output.

\begin{figure}
\centering{
    \subfigure[Energy breakdown across memory hierarchy]{
		\includegraphics[width=0.98\linewidth]{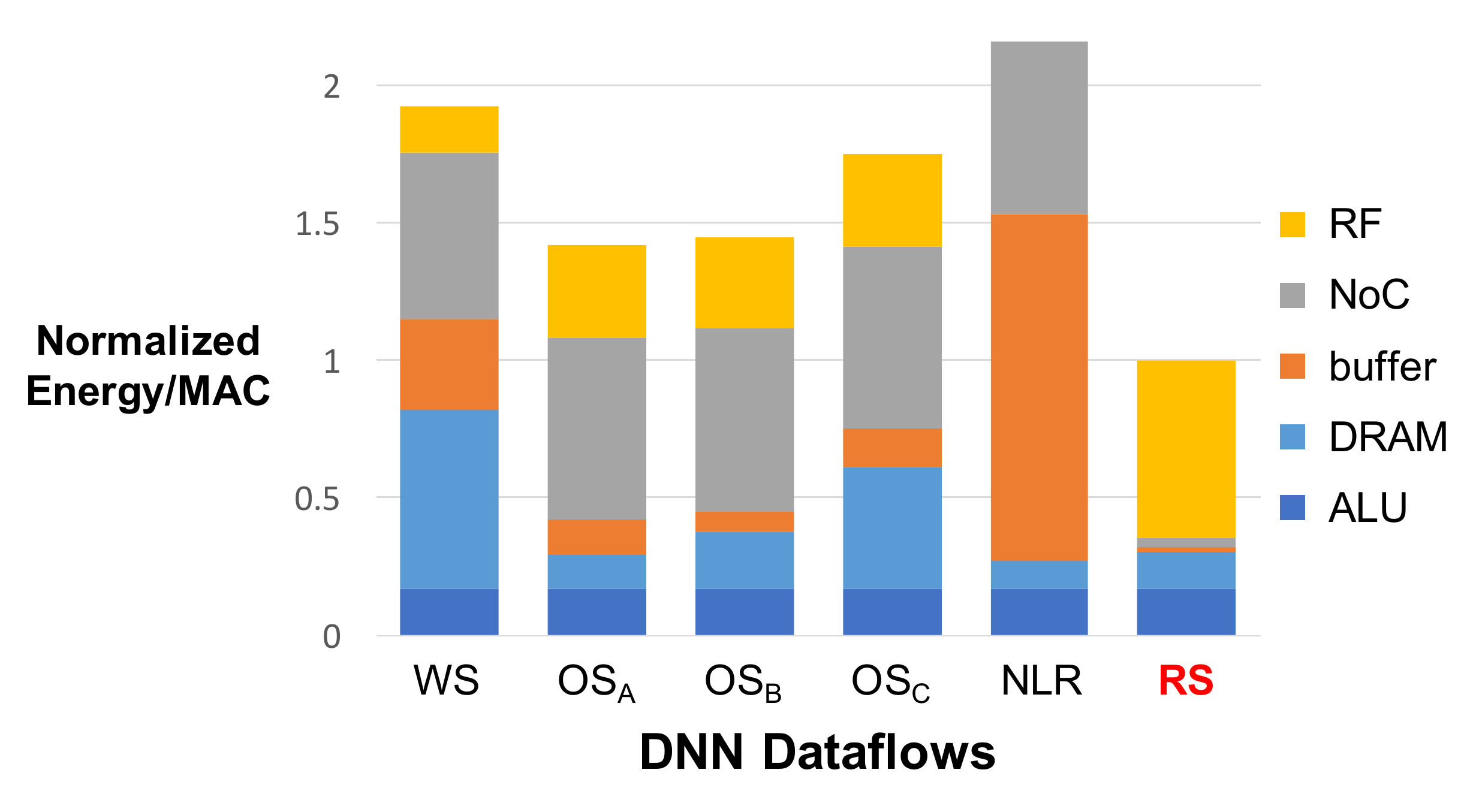}
		\label{fig:sim_results_conv_storage_hierarchy}
	}	
    \subfigure[Energy breakdown across data type]{
		\includegraphics[width=0.98\linewidth]{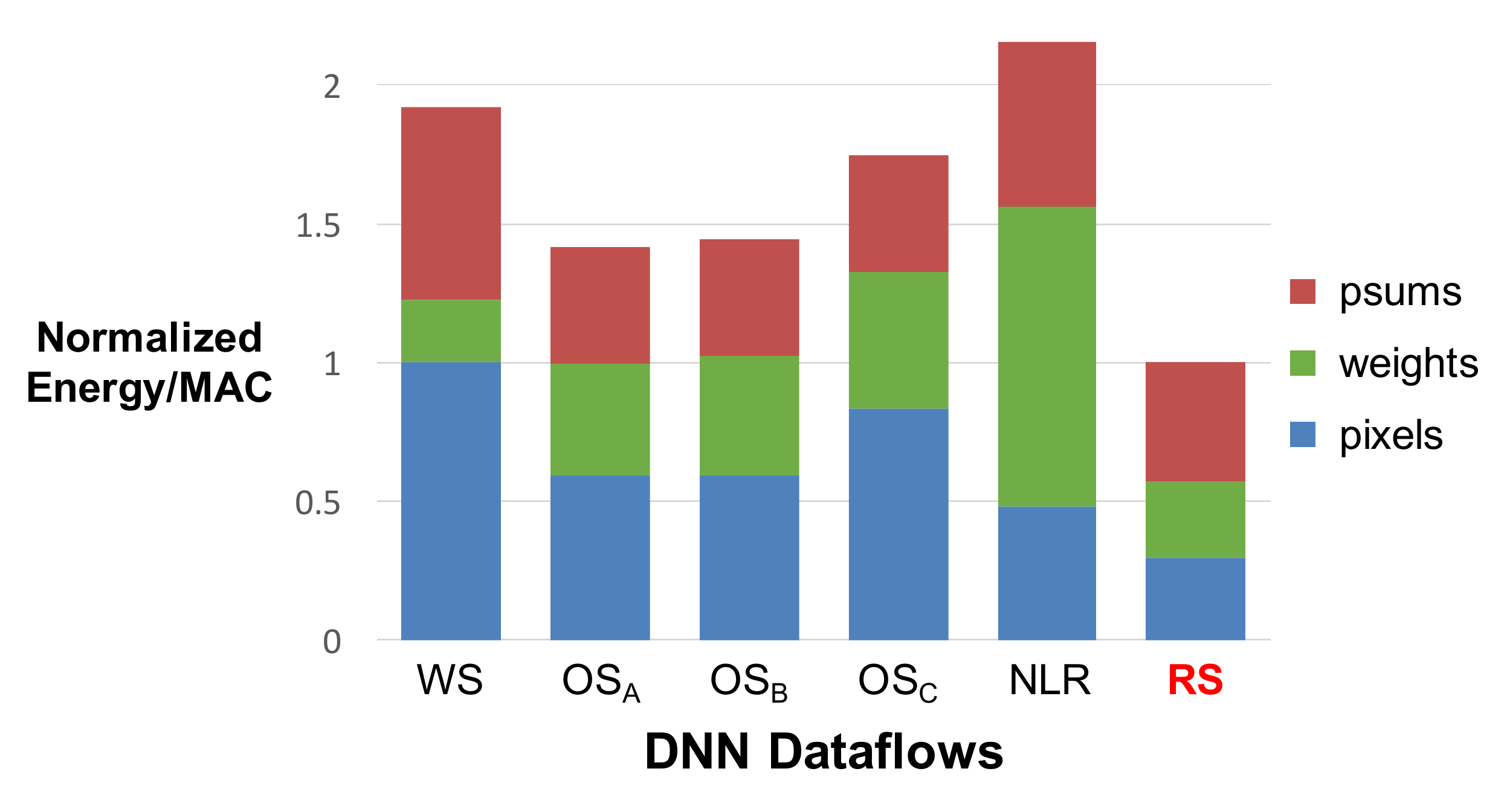}
		\label{fig:sim_results_conv_data_type}
	}
}
    \caption{   Comparison of energy efficiency between different dataflows in the CONV layers of AlexNet with a batch size of 16~\cite{nips2012-krizhevsky}: (a) breakdown in terms of storage levels and ALU, (b) breakdown in terms of data types. OS$_A$, OS$_B$ and OS$_C$ are three variants of the OS dataflow that are commonly seen in different implementations~\cite{isca2016-chen}.
            }
\label{fig:sim_results_conv}
\end{figure}

Fig.~\ref{fig:sim_results_conv} compares the chip and DRAM energy consumption of each dataflow for the CONV layers of AlexNet with a batch size of 16. The WS and OS dataflows have the lowest energy consumption for accessing weights and partial sums, respectively. However, the RS dataflow has the lowest total energy consumption since it optimizes for the overall energy efficiency instead of only for a certain data type. 

Fig.~\ref{fig:sim_results_conv_storage_hierarchy} shows the same results with breakdown in terms of memory hierarchy. The RS dataflow consumes the most energy in the RF, since by design most of the accesses have been moved to the lowest level of the memory hierarchy. This helps to achieve the lowest total energy consumption since RF has the lowest energy per access. The NLR dataflow has the lowest energy consumption at the DRAM level, since it has a much larger global buffer and thus higher on-chip storage capacity compared to others. However, most of the data accesses in the NLR dataflow is from the global buffer, which still has a relatively large energy consumption per access compared to accessing data from RF or inside the PE array. As a result, the overall energy consumption of the NLR dataflow is still fairly high. Overall, RS dataflow uses 1.4$\times$ to 2.5$\times$ lower energy than other dataflows. 

%The WS and OS dataflows have the lowest energy consumption for accessing weights and partial sums, respectively. This is what those dataflows are designed for. In particular, $OS_{A}$ consumes lower energy than $OS_{C}$, since it is designed for the CONV layers while $OS_{C}$ is for the FC layers. Overall, however, the RS dataflow has the lowest energy consumption since it optimizes for the overall energy efficiency instead of only for a certain data type. 

\begin{figure}
    \begin{center}
        \includegraphics[width=0.97\linewidth]{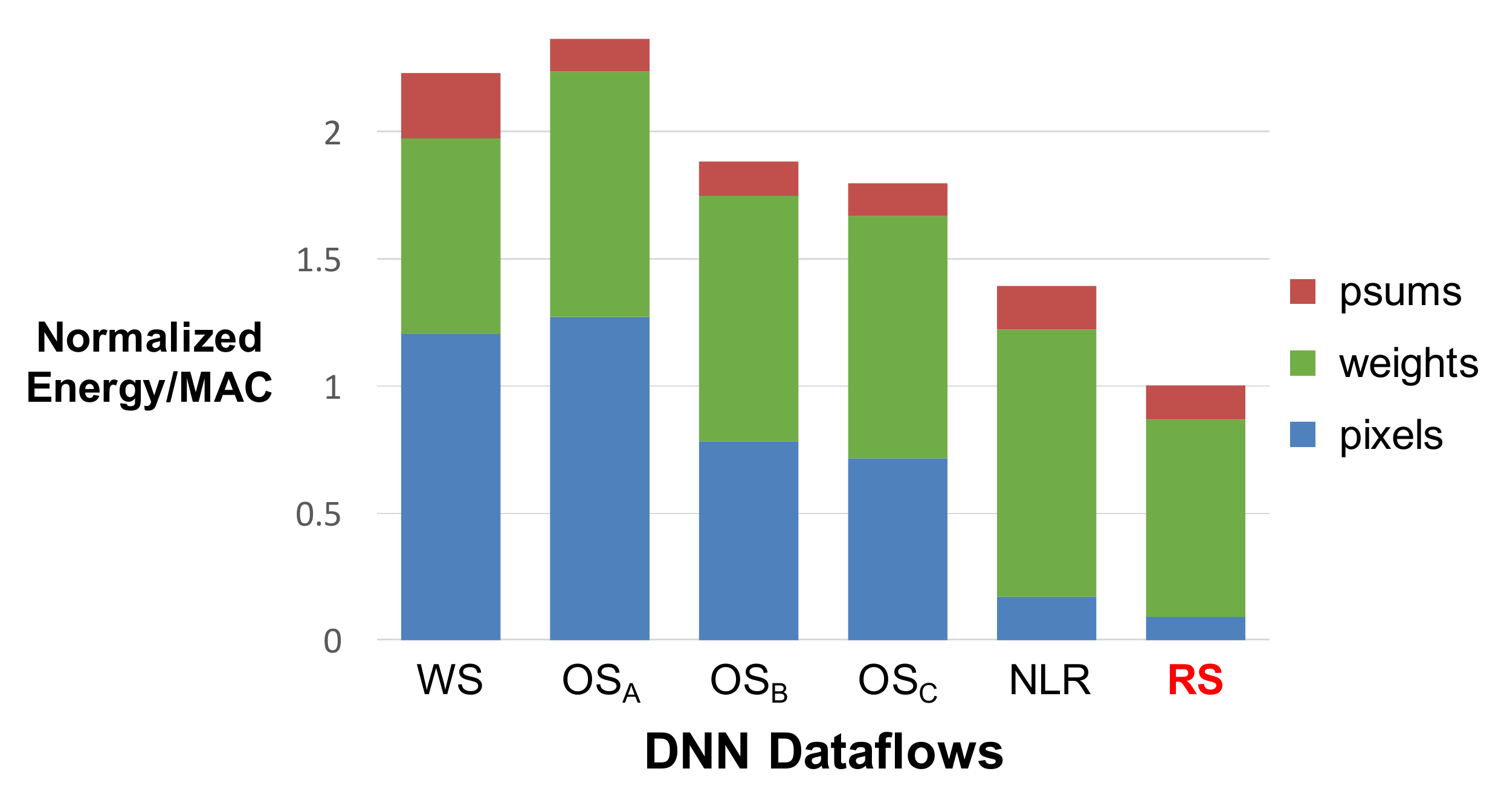}
        % \vspace{-5pt}
        \caption{    Comparison of energy efficiency between different dataflows in the FC layers of AlexNet with a batch size of 16~\cite{isca2016-chen}.
                }
        % \vspace{-10pt}
        \label{fig:fc_plot}
    \end{center}
\end{figure}

Fig.~\ref{fig:fc_plot} shows the energy efficiency between different dataflows in the FC layers of AlexNet with a batch size of 16. Since there is not as much data reuse in the FC layers as in the CONV layers, all dataflows spend a significant amount of energy on reading weights. However, RS dataflow still has the lowest energy consumption because it optimizes for the energy of accessing input activations and partial sums. For the OS dataflows, $OS_{C}$ now consumes lower energy than $OS_{A}$ since it is designed for the FC layers. Overall, RS still consumes 1.3$\times$ lower energy compared to other dataflows at the batch size of 16.

\begin{figure}
    \begin{center}
        \includegraphics[width=0.9\linewidth]{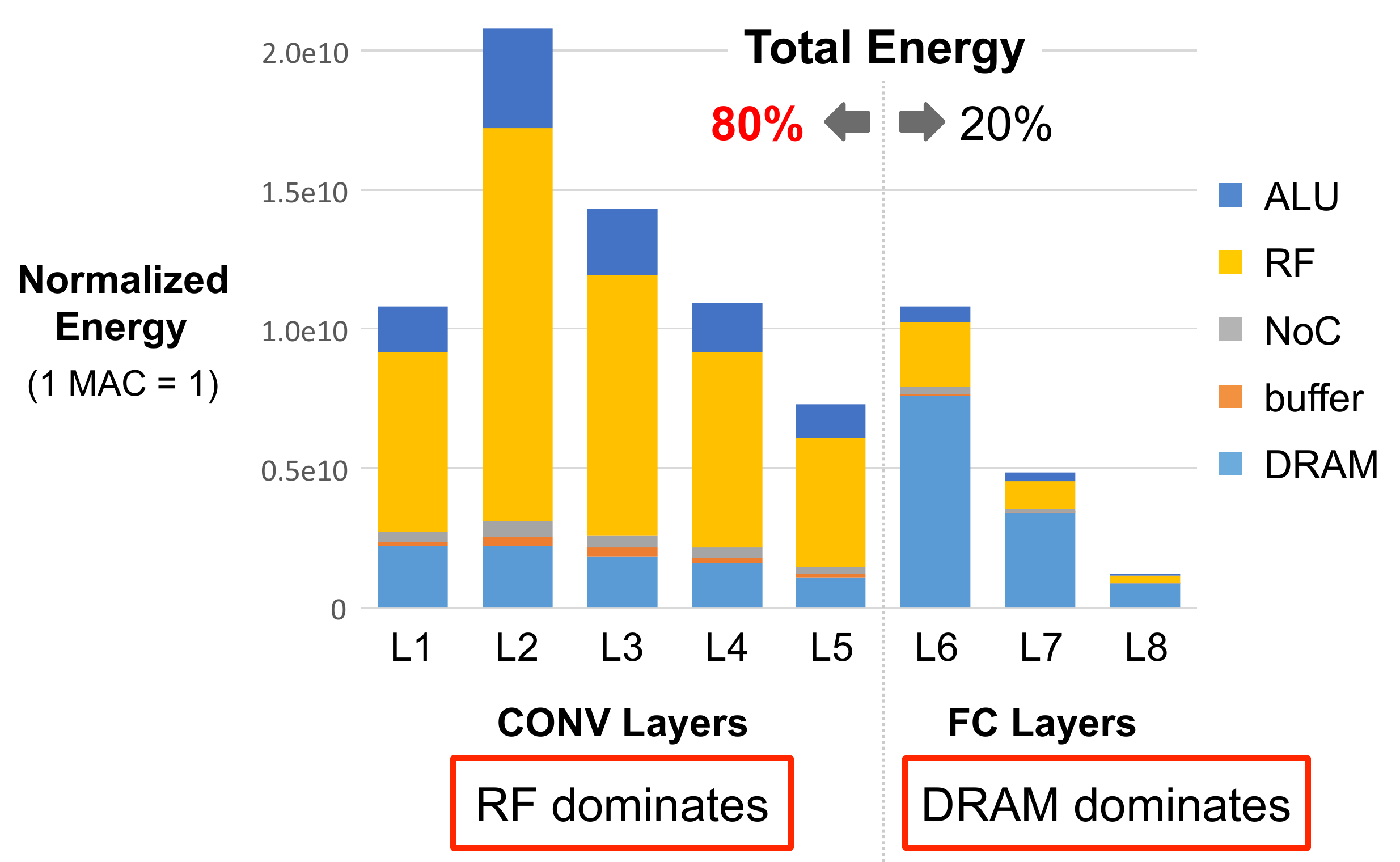}
        \caption{    Energy breakdown across layers of the AlexNet~\cite{isca2016-chen}.  RF energy dominates in convolutional layers. DRAM energy dominates in the fully connected layer.  Convolutional layer dominate energy consumption.
                }       
        \label{fig:breakdown_across_layers}
    \end{center}
\end{figure}

Fig.~\ref{fig:breakdown_across_layers} shows the RS dataflow design with energy breakdown in terms of different layers of AlexNet. In the CONV layers, the energy is mostly consumed by the RF, while in the FC layers, the energy is mostly consumed by DRAM. However, most of the energy is consumed by the CONV layers, which takes around 80\% of the energy.  As recent DNN models go deeper with more CONV layers, the ratio between number of CONV and FC layers only gets larger. Therefore, moving forward, significant effort should be placed on energy optimizations for CONV layers.

%From our simulation, using the multicast design saves more than 80\% of NoC energy consumption compared to a broadcast design.

%This shows the specifications of the Eyeriss implementation. It natively supports a wide range of DNN shape configuration and further uses the RS dataflow to optimize for energy consumption.

%This table shows the actual chip measurement results running the 5 CONV layers of AlexNet. The total amount of operations are 2.66 GMACS, which requires reading 16GB inputs and writing 5.4GB outputs. However, after the optimization of the RS dataflow, overall it only accesses 15.4MB of data from DRAM and 208.5MB of data from the global buffer. This is achieved by having most data reuse and accumulation in the PE array.

%In other words, for each input pixel that go through AlexNet, more than 50k operand accesses are required. But after the optimization of the RS dataflow, it only access 37 operand/pixel from DRAM and 506 operand/pixel from the global buffer.

%Here shows a comparison of Eyeriss with a mobile GPU, Nvidia TK1. Even though Eyeriss uses a much older technology, less multipliers and less on-chip storage, it still achieves more than 10 times higher energy efficiency than the TK1.

Finally, up until now, we have been looking at architectures with relatively limited storage on the order of a few hundred kilobytes. With much larger storage on the order of a few megabytes, additional dataflows can be considered. For example, Fused-Layer looks at dataflow optimizations across layers~\cite{alwani2016fused}.

%% file: technology.tex
\section{Near-Data Processing} 
\label{sec:technology}
The previous section highlighted that data movement dominates energy consumption.  While spatial architectures distribute the on-chip memory such that it is closer to the computation (e.g., into the PE), there have also been efforts to bring the off-chip high density memory closer to the computation or to integrate the computation into the memory itself; the latter is often referred to as \emph{processing-in-memory} or \emph{logic-in-memory}. In embedded systems, there have also been efforts to bring the computation into the sensor where the data is first collected. In this section, we will discuss how moving compute and data closer to reduce data movement (i.e., near-data processing) can be achieved using mixed-signal circuit design and advanced memory technologies.  

Many of these works use analog processing which has the drawback of increased sensitivity to circuit and device non-idealities. Consequentially, the computation is often performed at reduced precision, which can be accounted for during the training of the DNNs using the techniques discussed in Section~\ref{sec:algorithms}. Another factor to take into consideration is that DNNs are often trained in the digital domain; thus for analog processing, there is an additional overhead cost for analog-to-digital conversion (ADC) and digital-to-analog conversion (DAC).

\subsection{DRAM}
Advanced memory technology can reduce the access energy for high density memories such as DRAMs.  For instance, \emph{embedded DRAM (eDRAM)} brings high density memory on-chip to avoid the high energy cost of switching off-chip capacitance~\cite{keitel2001embedded}; eDRAM is 2.85$\times$ higher density than SRAM and 321$\times$ more energy efficient than DRAM (DDR3)~\cite{micro2014-chen}. eDRAM also offers higher bandwidth and lower latency compared to DRAM. In DNN processing, eDRAM can be used to store tens of megabytes of weights and activations on-chip to avoid off-chip access, as demonstrated in DaDianNao~\cite{micro2014-chen}.  The downside of eDRAM is that it has lower density than off-chip DRAM and can increase the cost of the chip. 

Rather than integrating DRAM into the chip itself, the DRAM can also be stacked on top of the chip using through silicon vias (TSV). This technology is often referred to as \emph{3-D memory}, and has been commercialized in the form of Hybrid Memory Cube (HMC)~\cite{jeddeloh2012hybrid} and High Bandwidth Memory (HBM)~\cite{standard2013high}. 3-D memory delivers an order of magnitude higher bandwidth and reduces access energy by up to 5$\times$ relative to existing 2-D DRAMs, as TSV have lower capacitance than typical off-chip interconnects.  Recent works have explored the use of HMC for efficient DNN processing in a variety of ways.  For instance, Neurocube~\cite{kim2016neurocube} integrates SIMD processors into the logic die of the HMC to bring the memory and computation closer together. Tetris~\cite{asplos2017-gao} explores the use of HMC with the Eyeriss spatial architecture and row stationary dataflow. It proposes allocating more area to computation than on-chip memory (i.e., larger PE array and smaller global buffer) in order to exploit the low energy and high throughput properties of the HMC. It also adapts the dataflow to account for the HMC memory and smaller on-chip memory. Tetris achieves a 1.5$\times$ reduction in energy consumption and 4.1$\times$ increase in throughput over a baseline system with conventional 2-D DRAM.

\subsection{SRAM}
\begin{figure}
\centering{
    \subfigure[Multiplication performed by bit-cell (Figure from~\cite{zhang2016machine}) ]
    {
		\includegraphics[width=0.5\linewidth]{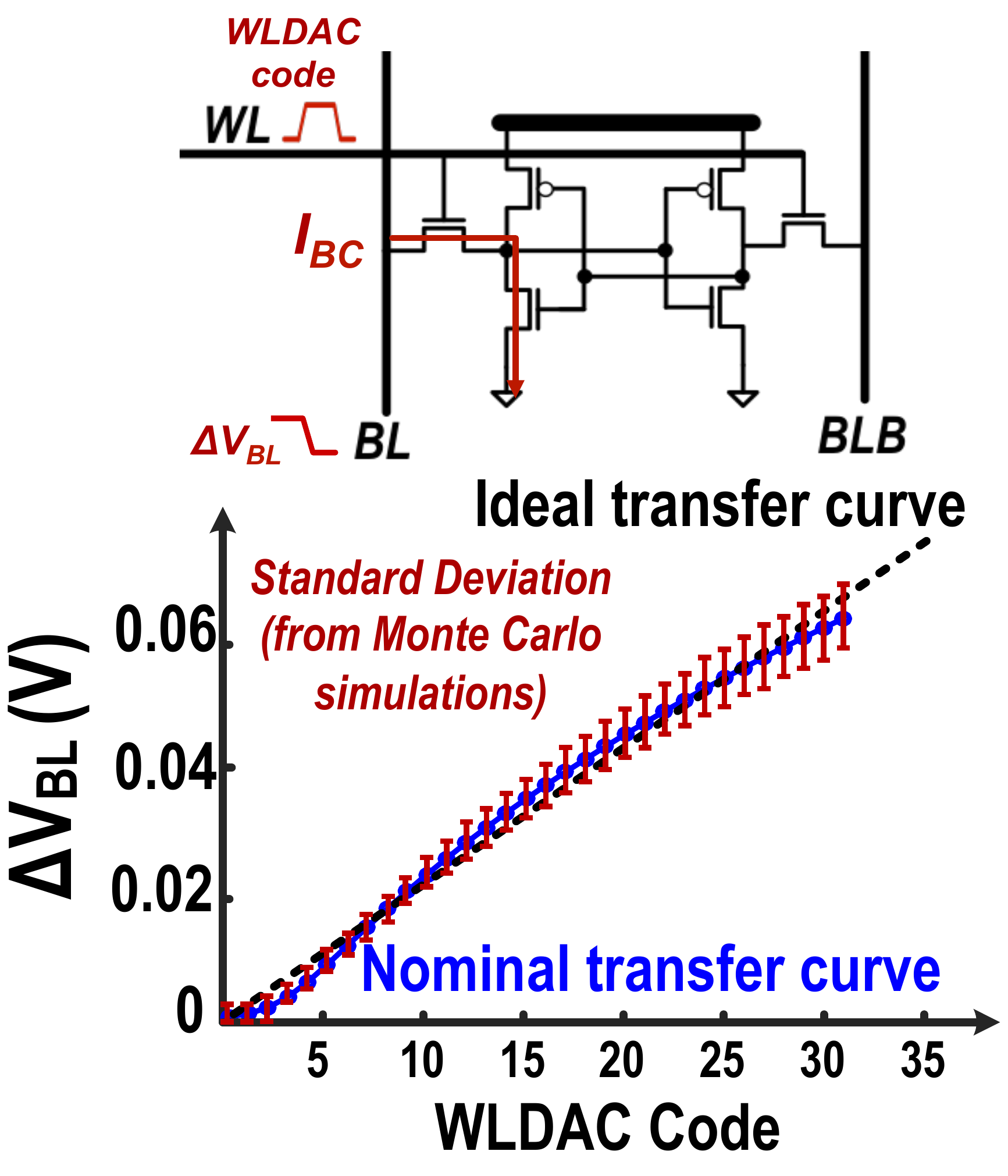}
		\label{fig:sram}
	}\hfill
    \subfigure[$G_{i}$ is conductance of resistive memory (Figure from~\cite{shafiee2016isaac}) ]
    {
		\includegraphics[width=0.4\linewidth]{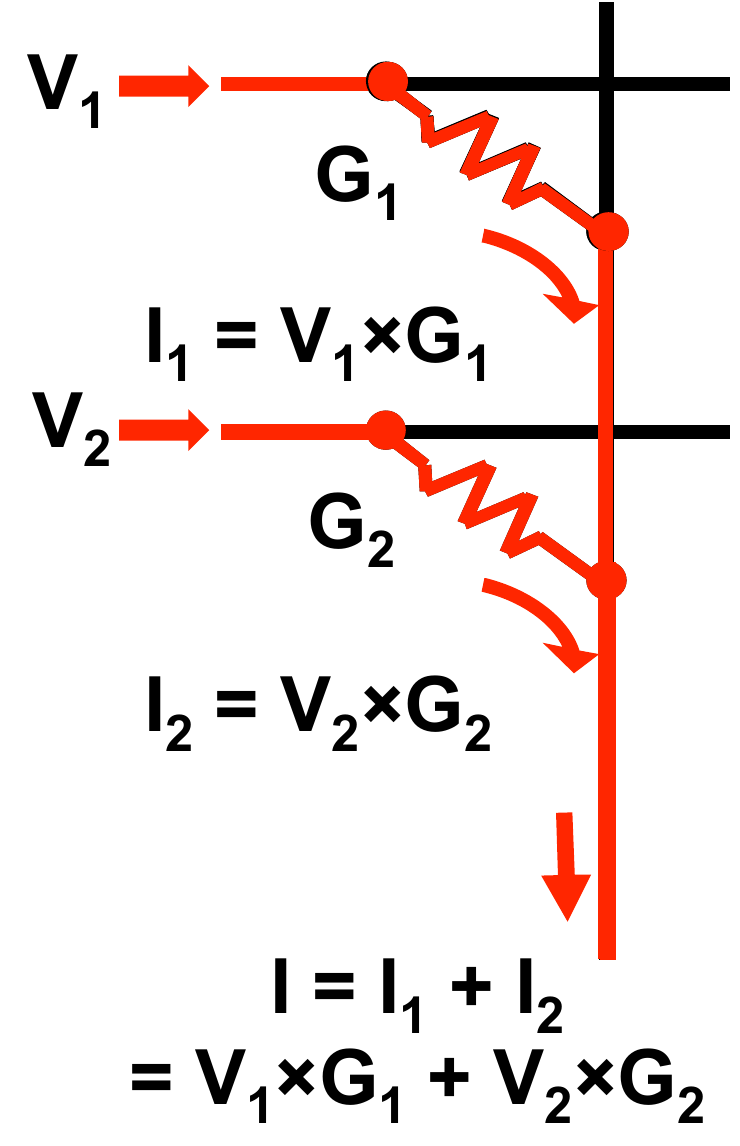}
				\label{fig:memristor}
	}
}
\caption{Analog computation by (a) SRAM bit-cell and (b) non-volatile resistive memory.}
\label{fig:analog}
\end{figure}

Rather than bringing the memory near the compute, recent work has also investigated bringing the compute into the memory. For instance, the multiply and accumulate operation can be directly integrated into the bit-cells of an SRAM array~\cite{zhang2016machine}, as shown in Fig.~\ref{fig:sram}. In this work, a 5-bit DAC is used to drive the word line (WL) to an analog voltage that represents the feature vector, while the bit-cells store the binary weights $\pm 1$.  The bit-cell current ($I_{BC}$) is effectively a product of the value of the feature vector and the value of the weight stored in the bit-cell; the currents from the bit-cells within a column add together to discharge the bitline ($V_{BL}$).  This approach gives 12$\times$ energy savings compared to reading the 1-bit weights from the SRAM and performing the computation separately. To counter circuit non-idealities, the DAC accounts for the non-linear bit-line discharge with respect to the WL voltage, and boosting is used to combine the weak classifiers that are susceptible to device variations to form a strong classifier~\cite{wang2014error}. 

%specifically, HMC gives 6.25$\times$ higher bandwidth than DDR3, where HMC has 16 channels at 10 GB/s versus 2 channel at 12.8 GB/s for DDR3.  NeuroCube also proposes to bring the computation closer to the memory (i.e., Processing in Memory) by integrating the computation into the logic die of the HMC of the DRAM.  The energy per bit was reported to be 3.7 pJ/bit for the DRAM layers and 6.78 pJ/bit for the logic layer, resulting in a total of 10.48 pJ/bit total for the HMC prototype. This is 6$\times$ more energy efficient than the reported 65 pJ/bit for existing DDR3 modules.

\subsection{Non-volatile Resistive Memories}
The multiply and accumulate operation can also be directly integrated into advanced \emph{non-volatile} high density memories by using them as programmable resistive elements, commonly referred to as \emph{memristors}~\cite{chua1971memristor}.  Specifically, a multiplication is performed with the resistor's conductance as the weight, the voltage as the input, and the current as the output as shown in Fig.~\ref{fig:memristor}. The addition is done by summing the currents of different memristors with Kirchhoff's current law.  This is the ultimate form of a weight stationary dataflow, as the weights are always held in place. The advantages of this approach include reduced energy consumption since the computation is embedded within memory which reduces data movement, and increased density since memory and computation can be densely packed with a similar density to DRAM~\cite{wilson2013international}.\footnote{The resistive devices can be inserted between the cross-point of two wires and in certain cases can avoid the need for an access transistor.} 

There are several popular candidates for non-volatile resistive memory devices including phase change memory (PCM), resistive RAM (RRAM or ReRAM), conductive bridge RAM (CBRAM), and spin transfer torque magnetic RAM (STT-MRAM)~\cite{micro_2016_devices_tutorial}. These devices have different trade-offs in terms of endurance (i.e., how many times it can be written), retention time, write current, density (i.e., cell size), variations and speed.

%\begin{itemize}
%    \item Phase change memory (PCM) changes between crystalline and amorphous phase with heating and cooling (induced by passing current through material to heat it up); a variant of PCM is used in DVDs. 
%    \item Metal oxide resistive RAM (RRAM) operated by applying an electric field to create motion of oxygen vacancies resulting on formation of conductive filament. 
%    \item Conductive bridge RAM (CBRAM) is similar except the metal oxide replaced by a solid electrolyte.  
%    \item Spin transfer torque magnetic RAM (STT-MRAM) stores state with ferro magnet spin up and spin down.
%\end{itemize}.

%PCM has better endurance ($10^{9}$) than FLASH and can support multi-bit storage (100$\times$ higher resistance ratio than STT-MRAM); however, it requires large programming current and its state drifts over time. STT-MRAM has high endurance cycling and speed and has a one to one ratio of write current and retention time; however, it’s low resistance ratio requires memory architecture that limits density. Finally, RRAM and CBRAM uses material common to semiconductor manufacturing and their endurance on par or better than PCM and higher speed; however, their resistance variation worst than PCM and STT-MRAM.

Processing with non-volatile resistive memories has several drawbacks as described in~\cite{eryilmaz2016neuromorphic}. First, it suffers from the reduced precision and ADC/DAC overhead of analog processing described earlier.  Second, the array size is limited by the wires that connect the resistive devices; specifically, wire energy dominates for large arrays (e.g., 1k$\times$1k), and the IR drop along wire can degrade the read accuracy. Third, the write energy to program the resistive devices can be costly, in some cases requiring multiple pulses. Finally, the resistive devices can also suffer from device-to-device and cycle-to-cycle variations with non-linear conductance across the conductance range.

%Similar to the mixed-signal circuits, the precision is limited, and the ADC and DAC conversion overhead must be considered in the overall cost, especially when the weights are trained in the digital domain.  

There have been several recent works that explore the use of memristors for DNNs. ISAAC~\cite{shafiee2016isaac} replaces the eDRAM in DaDianNao with memristors.  To address the limited precision support, ISAAC computes a 16-bit dot product operation with 8 memristors each storing 2-bits; a 1-bit$\times$2-bit multiplication is performed at each memristor, where a 16-bit input requires 16 cycles to complete. In other words, the ISAAC architecture trades off area and time for increased precision. Finally, ISAAC arranges its 25.1M memristors in a hierarchical structure to avoid issues with large arrays. PRIME~\cite{chi2016prime} also replaces the DRAM main memory with memristors; specifically, it uses 256$\times$256 memristor arrays that can be configured for 4-bit multi-level cell computation or 1-bit single level cell storage. It should be noted that results from ISAAC and PRIME are obtained from simulations.  The task of actually fabricating large memristors arrays is still very much a research challenge; for instance,~\cite{prezioso2015training} uses a fabricated 12$\times$12 memristor array to demonstrate a linear classifier.

%memristors to compute the product of a 3-bit input and 4-bit weight with dynamic fixed point to support a 6-bit output.  Two memristors can be combined to support 6-bit input and 8-bit weight. The

%In addition, in order to avoid issues with large arrays, ISAAC uses a hierarchical structure with each memristor crossbar (XB) being 128$\times$128, grouping 8 XBs to form a In-situ Multiply accumulate (IMA) and 12 IMAs per tile, and 16 tiles on their chip for a total of 25.1 million memrsistors - or total storage of 6.3MB.

%The ADC overhead can be avoided by training the weights directly in the analog domain. This is demonstrated in~\cite{prezioso2015training} using a fabricated 12$\times$12 memristor array that performs a linear classification on a 3$\times$3 binary image to generate three outputs.  The weights are stored in a differential form.  Including the bias, a total of 60 memristors are used.

\subsection{Sensors}
In certain applications, such as image processing, the data movement from the sensor itself can account for a significant portion of the system energy consumption. Thus there has also been research on performing the computation as close as possible to the sensor.  In particular, much of the work focuses on moving the computation into the analog domain to avoid using the ADC within the sensor, which accounts for a significant portion of the sensor power.  However, as mentioned earlier, lower precision is required for analog computation due to circuit non-idealities. 

In~\cite{zhang201518}, the matrix multiplication is integrated into the ADC, where the most significant bits of the multiplications are performed using switched capacitors in an 8-bit successive approximation format. This is extended in~\cite{lee201624} to not only perform the multiplications, but also the accumulations in the analog domain. In this work, it is assumed that 3-bits and 6-bits are sufficient to represent the weights and activations, respectively. This reduces the number of ADC conversions in the sensor by 21$\times$. RedEye~\cite{likamwa2016redeye} takes this approach even further by performing the entire convolution layer (including convolution, max pooling and quantization) in the analog domain at the sensor. It should be noted that~\cite{zhang201518} and~\cite{lee201624} report measured results from fabricated test chips, while results in~\cite{likamwa2016redeye} are from simulations.

It is also feasible to embed the computation not just before the ADC, but into the sensor itself.  For instance, in~\cite{wang2012180nm} an Angle Sensitive Pixels sensor is used to compute the gradient of the input, which along with compression, reduces the data movement from the sensor by 10$\times$. In addition, since the first layer of the DNN often outputs a gradient-like feature map, it maybe possible to skip the computations in the first layer, which further reduces energy consumption as discussed in~\cite{chen2016asp, suleiman2014energy}.

%% file: algorithms.tex
\section{Co-design of DNN models and Hardware} 
\label{sec:algorithms}
In earlier work, the DNN models were designed to maximize accuracy without much consideration of the implementation complexity. However, this can lead to designs that are challenging to implement and deploy.  To address this, recent work has shown that DNN models and hardware can be co-designed to jointly maximize accuracy and throughput, while minimizing energy and cost, which increases the likelihood of adoption. 
In this section, we will highlight various efforts that have been made towards the co-design of DNN models and hardware. Note that unlike Section~\ref{sec:architecture}, the techniques discussed in this section can affect the accuracy; thus, the goal is to not only substantially reduce energy consumption and increase throughput, but also to minimize any degradation in accuracy.

%A similar co-design approach was successfully applied widely used multimedia applications such as video compression.  Previously, video coding standards were mostly focused on algorithm development. Going from MPEG-2 to H.264/AVC gave a 2$\times$ improvement in coding efficiency at the cost of a 4$\times$ increase in decoder complexity~\cite{ostermann2004video}.  More recently, the development of the latest video coding standard H.265/HEVC~\cite{h265} employed algorithm and hardware co-design; as a result, it achieved an additional 2$\times$ improvement in coding efficiency over H.264/AVC~\cite{h264} at the cost of only 2$\times$ increase in decoder complexity. 

The co-design approaches can be loosely grouped into the following categories:
\begin{itemize}
    \item \emph{Reduce precision of operations and operands.}  This includes going from floating point to fixed point, reducing the bitwidth, non-linear quantization and weight sharing.
    \item \emph{Reduce number of operations and model size.} This includes techniques such as compression, pruning and compact network architectures.
\end{itemize}

\subsection{Reduce Precision}
\label{ssec:precision}
Quantization involves mapping data to a smaller set of quantization levels. The ultimate goal is to minimize the error between the reconstructed data from the quantization levels and the original data.  The number of quantization levels reflects the \emph{precision} and ultimately the number of bits required to represent the data (usually $log_{2}$ of the number of levels); thus, \emph{reduced precision} refers to reducing the number of levels, and thus the number of bits. The benefits of reduced precision include reduced storage cost and/or reduced computation requirements. 

\begin{figure}
\centering{	
    \subfigure[Linear Quantization]{
		\includegraphics[width=0.45\linewidth]{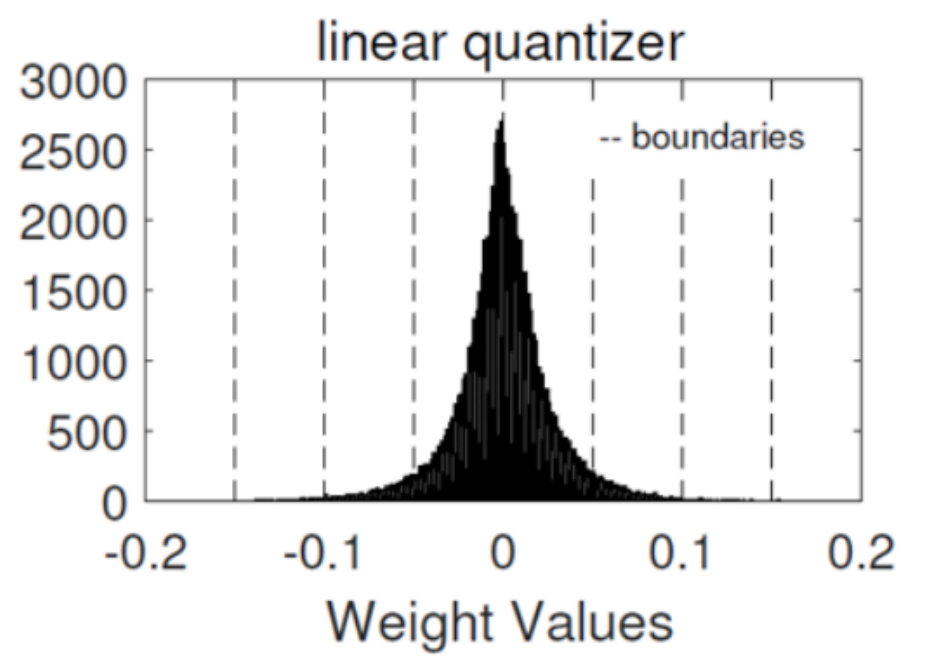}
		\label{fig:linear_quantization}
	}
    \subfigure[Log Quantization]{
		\includegraphics[width=0.45\linewidth]{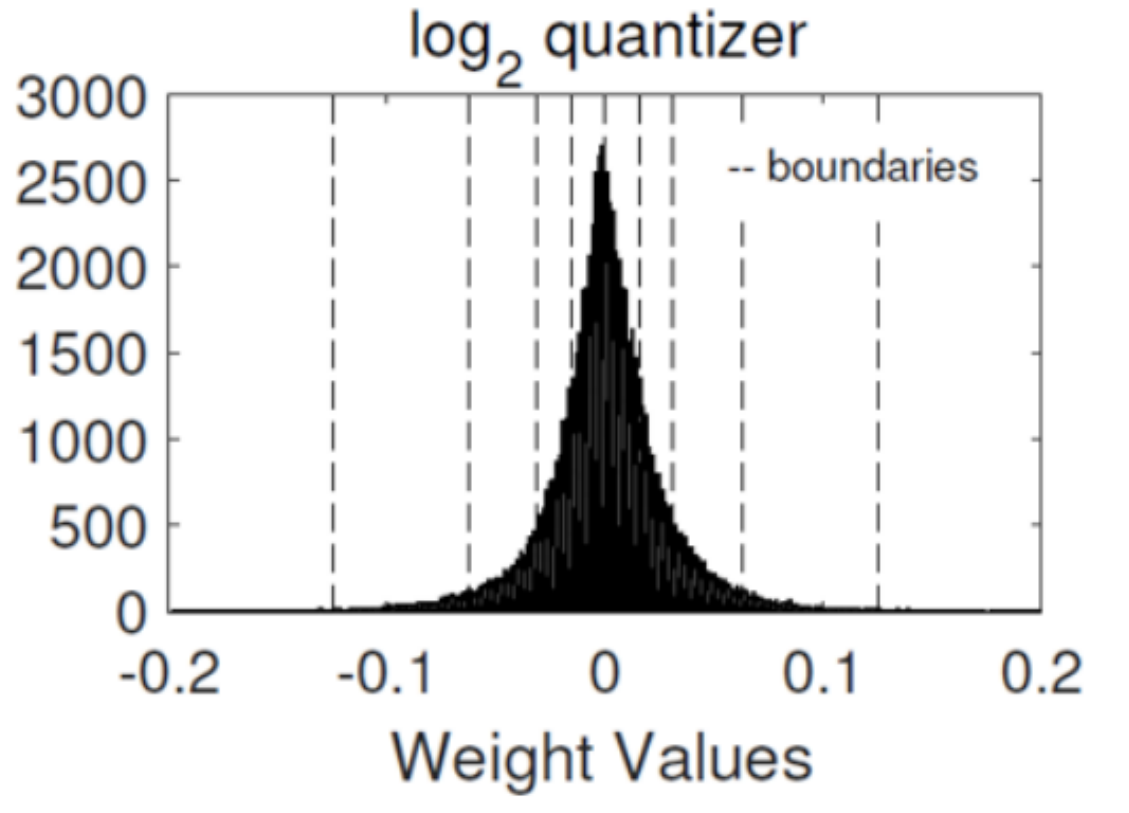}
		\label{fig:log_quantization}
	}
	    \subfigure[Non-Linear Quantization]{
		\includegraphics[width=0.4\linewidth]{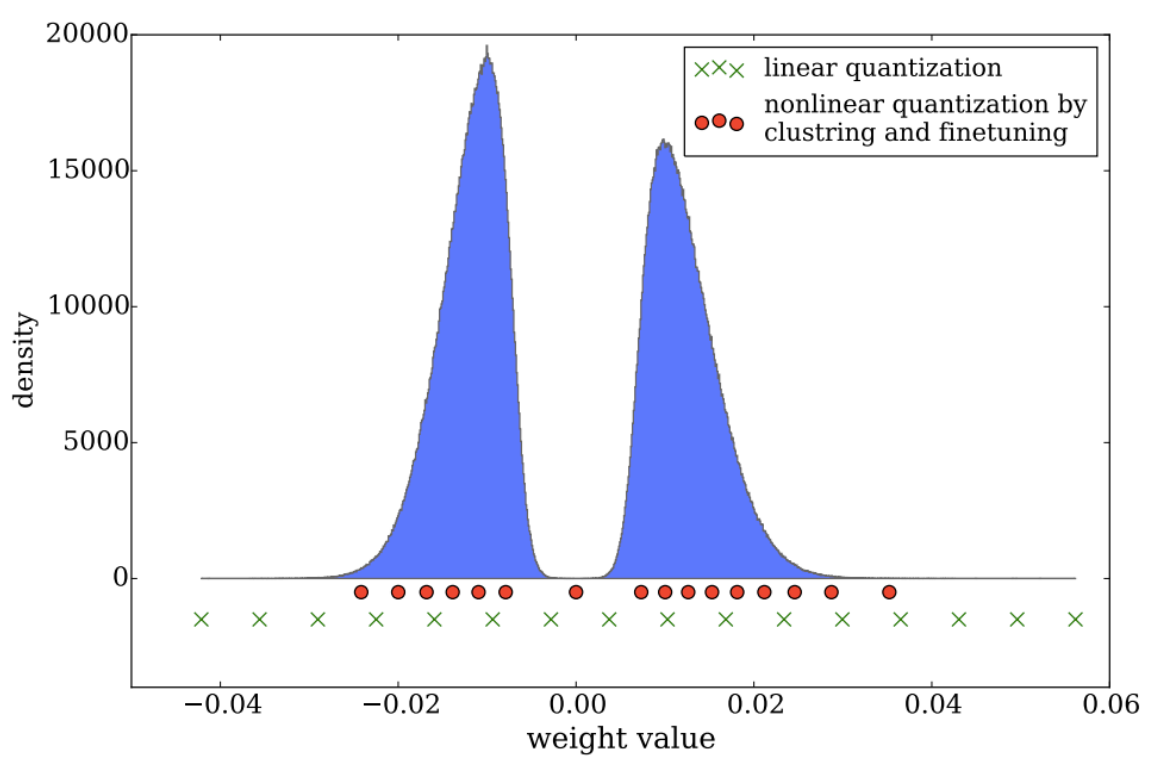}
		\label{fig:nonlinear_quantization}
	}
}
\caption{Various methods of quantization (Figures from~\cite{lee2017lognet,iclr2016-han-deep_comp}).}
\label{fig:quantization}
\end{figure}

There are several ways to map the data to quantization levels.  The simplest method is a linear mapping with uniform distance between each quantization level (Fig.~\ref{fig:linear_quantization}). Another approach is to use a simple mapping function such as a \emph{log function} (Fig.~\ref{fig:log_quantization}) where the distance between the levels  varies; this mapping can often be implemented with simple logic such as a shift. Alternatively, a more complex mapping function can be used where the quantization levels are determined or learned from the data (Fig.~\ref{fig:nonlinear_quantization}), e.g., using k-means clustering; for this approach, the mapping is usually implemented with a look up table.

Finally, the quantization can be fixed (i.e., the same method of quantization is used for all data types and layers, filters, and channels in the network); or it can be variable (i.e., different methods of quantization can be used for weights and activations, and different layers, filters, and channels in the network).

%We will now walk through the various forms of quantization that have been explored in recent works for improving the efficiency of DNNs. We will highlight their impact on storage and computation cost as well as their impact on accuracy.

%\subsubsection{Reducing the precision of a MAC}
Reduced precision research initially focused on reducing the precision of the weights rather than the activations, since weights directly increase the storage capacity requirement, while the impact of activations on storage capacity depends on the network architecture and dataflow.  However, more recent works have also started to look at the impact of quantization on activations.  Most reduced precision research also focuses on reducing the precision for inference rather than training (with some exceptions~\cite{arxiv2015-gupta, hubara2016quantized, zhou2016dorefa}) due to the sensitivity of the gradients to quantization. 

The key techniques used in recent work to reduce precision are summarized in Table~\ref{tab:precision}; both linear and non-linear quantization applied to weights and activations are explored. The impact on accuracy is reported relative to a baseline precision of 32-bit floating point, which is the default precision used on platforms such as GPUs and CPUs.

\subsubsection{Linear quantization}
\begin{figure}
\centering{	
    \subfigure[32-bit floating point example]{
		\includegraphics[width=0.96\linewidth]{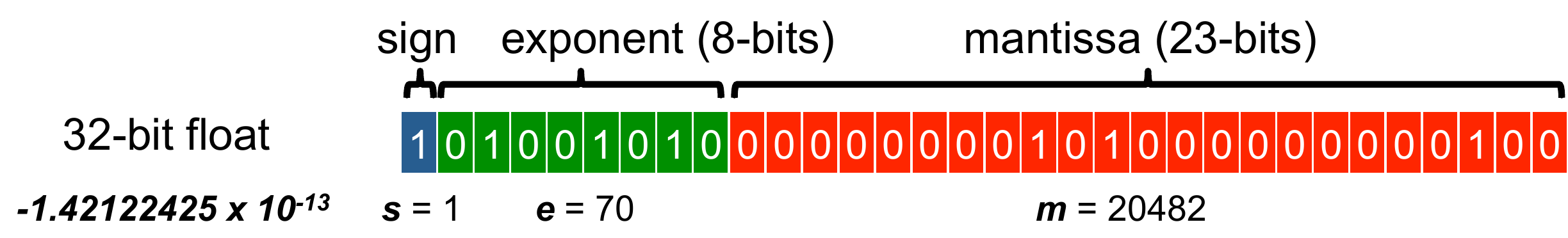}
		\label{fig:32bfloat}
	}
	    \subfigure[8-bit dynamic fixed point examples]{
		\includegraphics[width=0.96\linewidth]{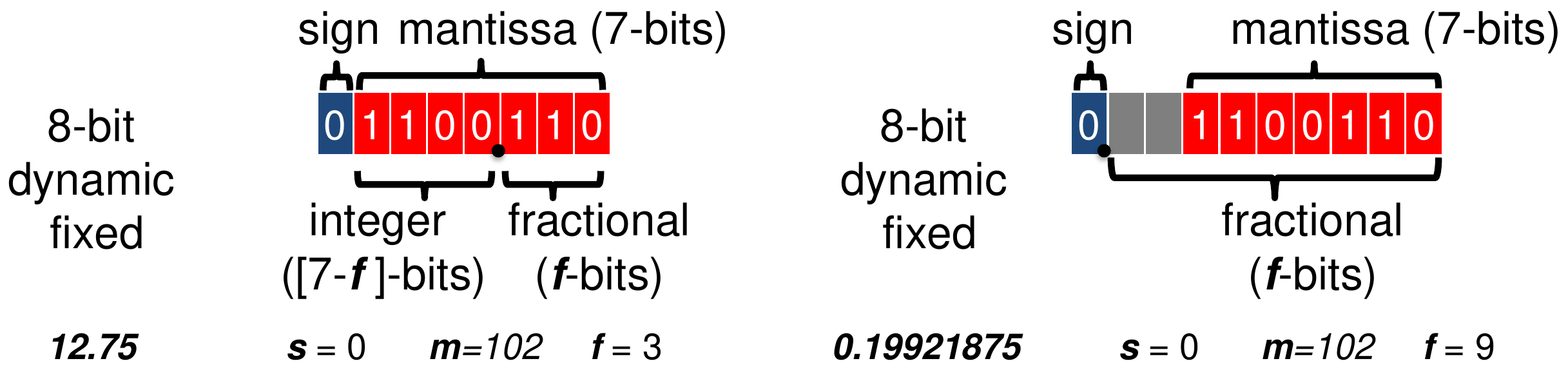}
		\label{fig:8bfixed}
	}
}
\caption{Various methods of number representations.}
\label{fig:precision}
\end{figure}

The first step of reducing precision is usually to convert values and operations from floating point to fixed point. A 32-bit floating point number, as shown in Fig.~\ref{fig:32bfloat}, is represented by $(-1)^{s}\times m \times2^{(e-127)}$, where $s$ is the sign bit, $e$ is the 8-bit exponent, and $m$ is the 23-bit mantissa, and covers the range of $10^{-38}$ to $10^{38}$.
    
An N-bit fixed point number is represented by $(-1)^{s}\times m \times 2^{-f}$, where $s$ is the sign bit, $m$ is the (N-1)-bit mantissa, and $f$ determines the location of the decimal point and acts as a scale factor.  For instance, for an 8-bit integer, when $f=0$, the dynamic range is -128 to 127, whereas when $f=10$, the dynamic range is -0.125 to 0.124023438. \emph{Dynamic} fixed point representation allows $f$ to vary based on the desired dynamic range as shown in Fig.~\ref{fig:8bfixed}. This is useful for DNNs, since the dynamic range of the weights and activations can be quite different. In addition, the dynamic range can also vary across layers and layer types (e.g., convolutional vs. fully connected). Using dynamic fixed point, the bitwidth can be reduced to 8 bits for the weights and 10 bits for the activations without any fine-tuning of the weights~\cite{ma2016scalable}; with fine-tuning, both weights and activations can reach 8-bits~\cite{gysel2016hardware}.

Using 8-bit fixed point has the following impact on energy and area~\cite{isscc2014-horowitz}:

\begin{itemize}
    \item An 8-bit fixed point add consumes 3.3$\times$ less energy (3.8$\times$ less area) than a 32-bit fixed point add, and 30$\times$ less energy (116$\times$ less area) than a 32-bit floating point add. The energy and area of a fixed-point add scales approximately linearly with the number of bits.  
    \item  An 8-bit fixed point multiply consumes 15.5$\times$ less energy (12.4$\times$ less area) than a 32-bit fixed point multiply, and 18.5$\times$ less energy (27.5$\times$ less area) than a 32-bit floating point multiply. The energy and area of a fixed-point multiply scales approximately quadratically with the number of bits.
\end{itemize}

Reducing the precision also reduces the energy and area cost for storage, which is important since memory access and data movement dominate energy consumption as described earlier.  The energy and area of the memory scale approximately linearly with number of bits. It should be noted, however, that changing from floating point to fixed point, without reducing bit-width, does not reduce the energy or area cost of the memory.

%It should be noted that multiplications are significantly more expensive than an addition; even at 8-bit fixed point, a multiply requires 6.7$\times$ more energy and 7.8$\times$ more energy than an add.  Thus there is more effort in reducing the cost of multiplication than the accumulation.   

%Reducing the precision also reduces the energy and area cost for storage.  As mentioned in previous sections, the energy cost of reading from and writing to memory is often greater than the computation itself.  For instance, reading a 32-bit value from an 8kB SRAM requires 1.35$\times$ and 5.55$\times$ more energy than a 32-bit floating-point multiplication and addition, respectively~\cite{isscc2014-horowitz}. The memory access energy also increases for larger capacity memories due to the switching capacitance of the bit lines and word lines. Finally, very large capacity memories, in the gigabytes range, are often placed off-chip in DRAM, which increases the energy by over two orders of magnitude. 

\begin{figure}
    \begin{center}
        \includegraphics[width=0.9\linewidth]{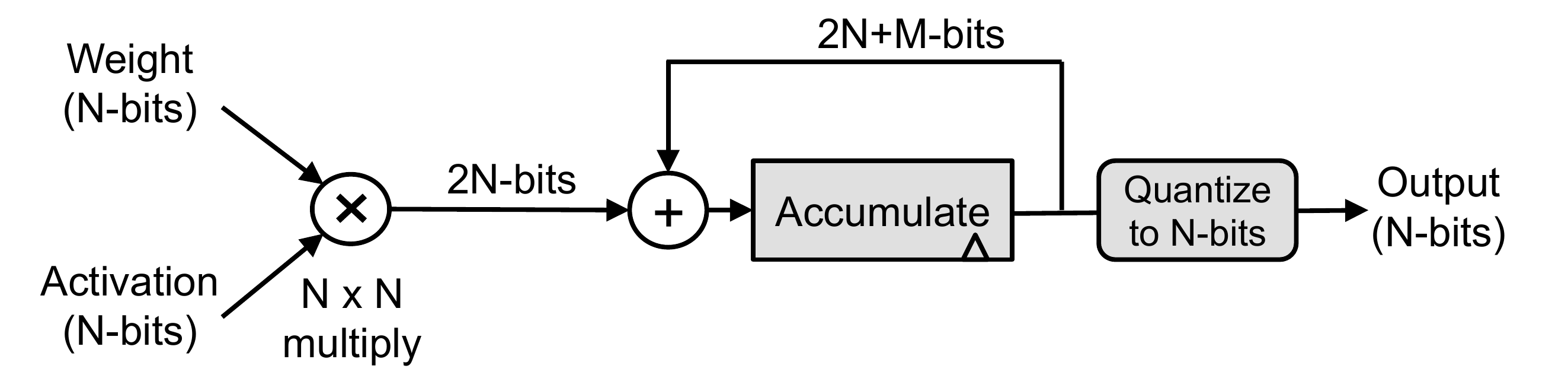}
        \caption{    Reducing the precision of multiply and accumulate (MAC).
                }              
        \label{fig:MAC_quantization}
    \end{center}
\end{figure}

For completeness, it should be noted that the precision of the internal values of a fixed-point multiply and accumulate (MAC) operation are typically higher than the weights and activations.  To guarantee no precision loss, weights and input activations with N-bit fixed-point precision would require an N-bit$\times$N-bit multiplication which generates a 2N-bit output product; that output would need to be accumulated with 2N+M-bit precision, where M is determined based on the largest filter size $log_{2}$($C\times R \times S$ from Fig.~\ref{fig:DNN_conv}), which is in the range of 10 to 16 bits for the popular DNNs described in Section~\ref{sec:popular}.  After accumulation, the precision of the final output activation is typically reduced to N-bits~\cite{arxiv2015-gupta, ma2016scalable}, as shown in Fig.~\ref{fig:MAC_quantization}. The reduced output precision does not have a significant impact on accuracy if the distribution of the weights and activations are centered near zero such that the accumulation would not move only in one direction; this is particularly true when batch normalization is used.  

The reduced precision is not only explored in research, but has been used in recent commercial platforms for DNN processing. For instance, Google's Tensor Processing Unit (TPU) which was announced in May 2016, was designed for 8-bit integer arithmetic~\cite{tpu}.  Similarly, Nvidia's PASCAL GPU, which was announced in April 2016, also has 8-bit integer instructions for deep learning inference~\cite{pascal}. In general purpose platforms such as CPUs and GPUs, the main benefit of using 8-bit computation is an increase in throughput, as four 8-bit operations rather than one 32-bit operation can be performed for a given clock cycle.

While general purpose platforms usually support 8-bit, 16-bit and/or 32-bit operations, it has been shown that the minimum bit precision for DNNs can actually vary in a more fine grained manner. For instance, the weight and activation precision can vary between 4 and 9 bits for AlexNet across different layers without significant impact on accuracy (i.e., a change of less than 1\%)~\cite{judd2016stripes, moons20160}. This fine-grained variation can be exploited for increased throughput or reduced energy consumption with specialized hardware. For instance, if bit-serial processing is used, where the number of clock cycles to complete an operation is proportional to the bitwidth, adapting to fine-grain variations in bit precision can result in a 2.24$\times$ speed up versus 16-bits~\cite{judd2016stripes}. Alternatively, a multiplier can be designed such that its critical path reduces based on the bit precision as fewer adders are needed to resolve the product; this can be combined with voltage scaling for a 2.56$\times$ energy savings versus 16-bits~\cite{moons20160}. While these bit scaling results are reported relative to 16-bit, it would be interesting to see their impact relative to the maximum precision required across layers (i.e., 9-bits for~\cite{judd2016stripes, moons20160}).

The precision can be reduced even more aggressively to a single bit; this area of research is often referred to as \emph{binary nets}.  BinaryConnect (BC)~\cite{nips2015-courbariaux-binaryconnect} introduced the concept of binary weights (i.e., -1 and 1), where using a binary weight reduced the multiplication in the MAC to addition and subtraction only. This was later extended in Binarized Neural Networks (BNN)~\cite{courbariaux2016binarynet} that uses binary weights \emph{and} activations, which reduces the MAC to an XNOR. However, BC and BNN have an accuracy loss of 19\% and 29.8\%, respectively~\cite{eccv2016-rastegari-xnor_net}.   

In order to reduce this accuracy loss, Binary Weight Nets (BWN) and XNOR-Nets introduced several significant modifications to the DNN processing~\cite{eccv2016-rastegari-xnor_net}. This includes multiplying the outputs with a scale factor to recover the dynamic range (i.e., the weights effectively become -$w$ and $w$, where $w$ is the average of the absolute values of the
weights in the filter)\footnote{This can also be thought of as a form of weights sharing, where only two weights are used per filter.}, keeping the first and last layers at 32-bit floating point precision, and performing normalization before convolution to reduce the dynamic range of the activations.  With these changes, BWN reduced the accuracy loss to 0.8\%, while XNOR-Nets reduced the loss to 11\%.  The loss of XNOR-Net can be further reduced by increasing the precision of the activations to be slightly larger than one bit. For instance, Quantized Neural Networks (QNN)~\cite{hubara2016quantized}, DoReFa-Net~\cite{zhou2016dorefa}, and HWGQ-Net~\cite{cai17hwgq} allow the activations to have 2-bits, while the weights remain at 1-bit; in HWGQ-Net, this reduces the accuracy loss to 5.2\%.

%note that this might affect the dataflow resulting in higher energy cost

All the previously described binary nets limit the weights to two values (-$w$ and $w$); however, there may be benefits for allowing weights to be zero (i.e., -$w$, 0, $w$). Although this requires an additional bit per weight compared to binary weights, the sparsity of the weights can be exploited to reduce computation and storage cost, which can potentially cancel out the cost of the additional bit.  This is explored in Ternary Weight Nets (TWN)~\cite{li2016ternary} and then extended in Trained Ternary Quantization (TTQ) where a different scale is trained for each weight (i.e., -$w_{1}$, 0, $w_{2}$) for an accuracy loss of 0.6\%~\cite{zhu2016trained}, assuming 32-bit floating point for the activations.

Hardware implementations for binary/ternary nets have been explored in recent publications. YodaNN~\cite{andri2016yodann} uses binary weights, while BRein~\cite{BRein2017} uses binary weights and activations. Binary weights are also used in the compute in SRAM work~\cite{zhang2016machine} described in Section~\ref{sec:technology}. Finally, the nominally spike-inspired TrueNorth chip can implement a reduced precision neural network with binary activations and ternary weights using TrueNorth's quantized weight table~\cite{esser2016convolutional}. These works tend not to support state-of-the-art DNN models (with the exception of YodaNN).

%\begin{figure}
%    \begin{center}
%        \includegraphics[width=0.9\linewidth]{figures/binary.pdf}
%        \caption{Binarization of weights and activations. Figures adopted from ~\cite{eccv2016-rastegari-xnor_net}.
%                }      
%        \label{fig:binary}
%    \end{center}
%    \vspace{-10pt}
%\end{figure}

\begin{table*}
\centering
\begin{tabular}{|l|c|c|c|c|c|c|}
\hline
\multicolumn{2}{|c|}{\multirow{2}{*}{\textbf{Reduce Precision Method}}}  &\multicolumn{2}{|c|}{\textbf{bitwidth}}& \textbf{Accuracy loss vs.}\\\cline{3-4}
\multicolumn{2}{|c|}{} & \textbf{Weights}& \textbf{Activations}& \textbf{32-bit float (\%)}\\\hline
\multirow{2}{*}{\textbf{Dynamic Fixed Point}} &\textbf{w/o fine-tuning~\cite{ma2016scalable}}& {8}& {10} &{0.4}\\\cline{2-5}
 & \textbf{w/ fine-tuning~\cite{gysel2016hardware}}&{8}& {8} &{0.6}\\\hline
\multirow{4}{*}{\textbf{Reduce Weight}}&\textbf{BinaryConnect~\cite{nips2015-courbariaux-binaryconnect}} & {1}& {32 (float)} &{19.2}\\\cline{2-5}
&\textbf{Binary Weight Network (BWN)~\cite{eccv2016-rastegari-xnor_net}} & {1*}& {32 (float)} &{0.8}\\\cline{2-5}
&\textbf{Ternary Weight Networks (TWN)~\cite{li2016ternary}} & {2*}& {32 (float)} &{3.7}\\\cline{2-5}
&\textbf{Trained Ternary Quantization (TTQ)~\cite{zhu2016trained}} & {2*}& {32 (float)} &{0.6}\\\hline
\multirow{5}{*}{\textbf{Reduce Weight and Activation}}&\textbf{XNOR-Net~\cite{eccv2016-rastegari-xnor_net}} & {1*}& {1*} &{11} \\\cline{2-5}
&\textbf{Binarized Neural Networks (BNN)~\cite{courbariaux2016binarynet}} & {1}& {1} &{29.8}\\\cline{2-5}
&\textbf{DoReFa-Net~\cite{zhou2016dorefa}} & {1*}& {2*} &{7.63}\\\cline{2-5}
&\textbf{Quantized Neural Networks (QNN)~\cite{hubara2016quantized}} & {1}& {2*} &{6.5}\\\cline{2-5}
&\textbf{HWGQ-Net~\cite{cai17hwgq}} & {1*}& {2*} &{5.2}
\\\hline
\multirow{4}{*}{\textbf{Non-linear Quantization}}
&\textbf{LogNet~\cite{miyashita2016convolutional}} & {5 (conv), 4 (fc)}& {4} &{3.2}\\\cline{2-5}
&\textbf{Incremental Network Quantization (INQ)~\cite{zhou2016inq}} & {5}& {32 (float)} &{-0.2}\\\cline{2-5}
&\multirow{2}{*}{\textbf{Deep Compression~\cite{iclr2016-han-deep_comp}}} & {8 (conv), 4 (fc)}& {16} &{0}\\\cline{3-5}
& & {4 (conv), 2 (fc)}& {16} &{2.6}\\\hline
\end{tabular}
\caption{Methods to reduce numerical precision for AlexNet.  Accuracy measured for Top-5 error on ImageNet. *Not applied to first and/or last layers}
\label{tab:precision}
\end{table*}

\subsubsection{Non-linear quantization}
\label{sssec:nonlinear}
The previous works described involve linear quantization where the levels are uniformly spaced out.  It has been shown that the distributions of the weights and activations are not uniform~\cite{iclr2016-han-deep_comp, miyashita2016convolutional}, and thus a non-linear quantization can potentially improve accuracy.  Specifically, there have been two popular approaches taken in recent works: (1) log domain quantization; (2) learned quantization or weight sharing.

\emph{Log domain quantization}
If the quantization levels are assigned based on a logarithmic distribution as shown in Fig~\ref{fig:log_quantization}, the weights and activations are more equally distributed across the different levels and each level is used more efficiently resulting in less quantization error.  For instance, using 4 bits in linear quantization results in a 27.8\% loss in accuracy versus a 5\% loss for log base-2 quantization for VGG-16~\cite{lee2017lognet}. Furthermore, when weights are quantized to powers of two, the multiplication can be replaced with a bit-shift~\cite{gysel2016hardware, miyashita2016convolutional}.\footnote{Note however that multiplications do not account for a significant portion of the total energy.}  Incremental Network Quantization (INQ) can be used to further reduce the loss in accuracy by dividing the large and small weights into different groups, and then iteratively quantizing and re-training the weights~\cite{zhou2016inq}.

\emph{Weight Sharing} forces several weights to share a single value. This reduces the number of unique weights in a filter or a layer. One example is to group the weights by using a hashing function and use one value for each group~\cite{icml2015-chen}. Alternatively, the weights can be grouped by the k-means algorithm~\cite{iclr2016-han-deep_comp}. Both the shared weights and the indexes indicating which weight to use at each position of the filter are stored. This leads to a two step process to fetch the weight: (1) read the weight index; (2) using the weight index, read the shared weights.  This approach can reduce the cost of reading and storing the weights if the weight index ($log_{2}$ of the number of unique weights) is less than the bitwidth of the weight itself. 

\begin{figure}
    \begin{center}
        \includegraphics[width=0.9\linewidth]{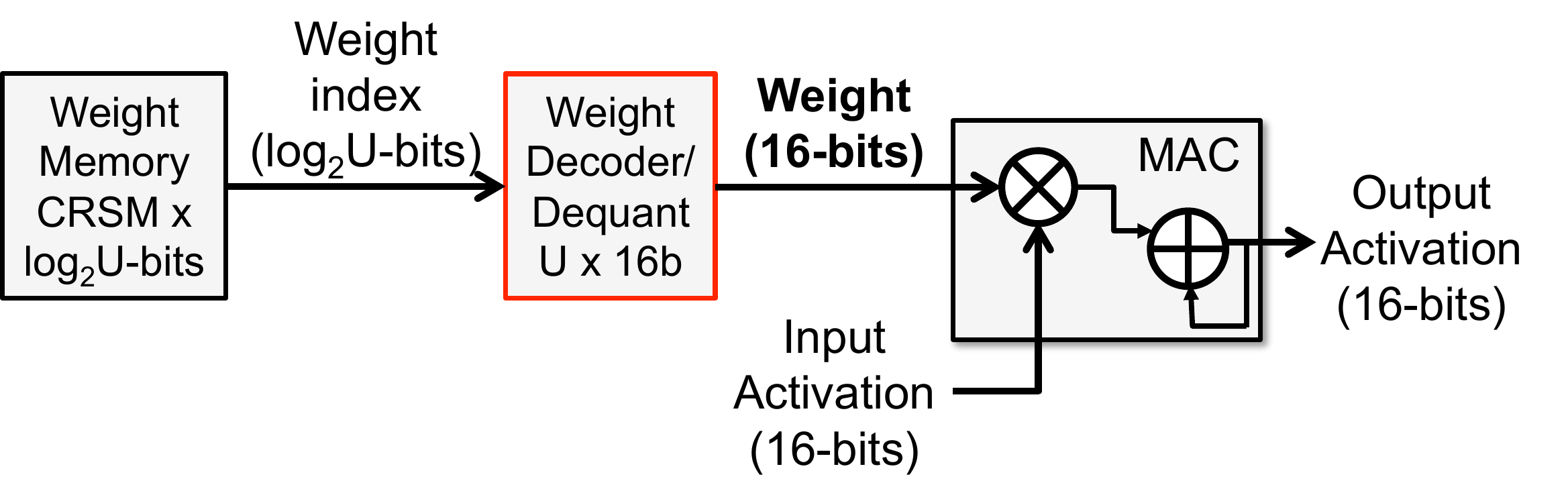}
        \caption{   Weight sharing hardware.
                }      
        \label{fig:weight_sharing}
    \end{center}
    \vspace{-10pt}
\end{figure}

For instance, in Deep Compression~\cite{iclr2016-han-deep_comp}, the number of unique weights per layer is reduced to 256 for convolutional layers and 16 for fully-connected layers in AlexNet, requiring 8-bit and 4-bit weight indexes, respectively.  Assuming there are $U$ unique weights and the size of the filters in the layer is C$\times$R$\times$S$\times$M from Fig.~\ref{fig:DNN_conv}, there will be energy savings if reading from a $CRSM\times log_{2}U$-bit memory plus a $U\times$16-bit memory (as shown in Fig.~\ref{fig:weight_sharing}) cost less than reading from a $CRSM\times$16-bit memory. Note that unlike the previous quantization methods, the weight sharing approach does not reduce the precision of the MAC computation itself and only reduces the weight storage requirement.

\subsection{Reduce Number of Operations and Model Size}
In addition to reducing the size of each operation or operand (weight/activation), there is also a significant amount of research on methods to reduce the number of operations and model size.  These techniques can be loosely classified as exploiting activation statistics, network pruning, network architecture design and knowledge distillation.

\subsubsection{Exploiting Activation Statistics}
\label{ssec:statistics}
\begin{figure}
\centering{
    \subfigure[ReLU non-linearity]{
		\includegraphics[width=0.9\linewidth]{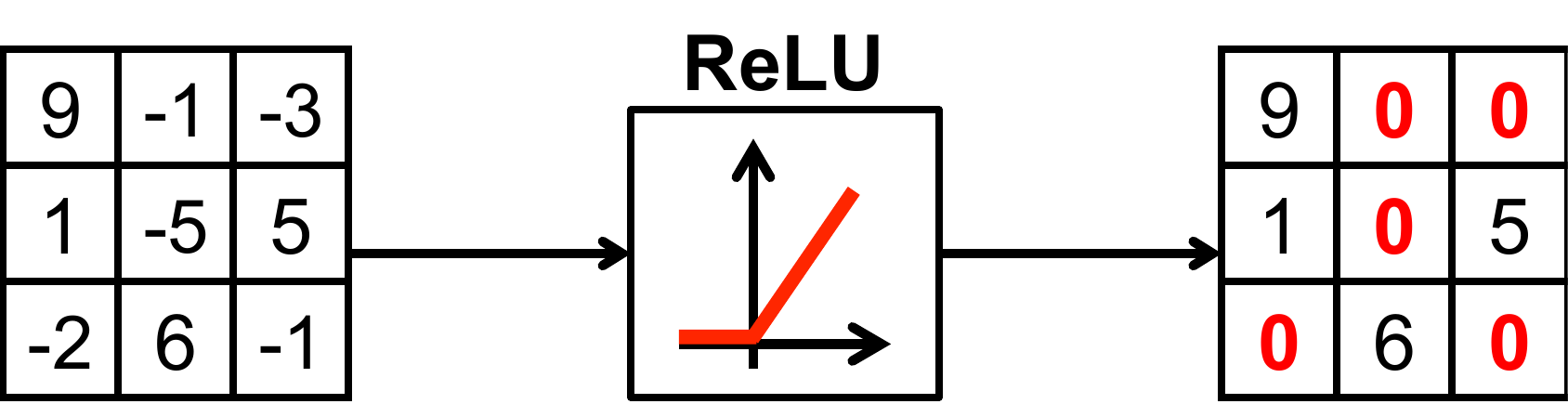}
		\label{fig:relu}
	}	
    \subfigure[Distribution of activation after ReLU of AlexNet]{
		\includegraphics[width=0.9\linewidth]{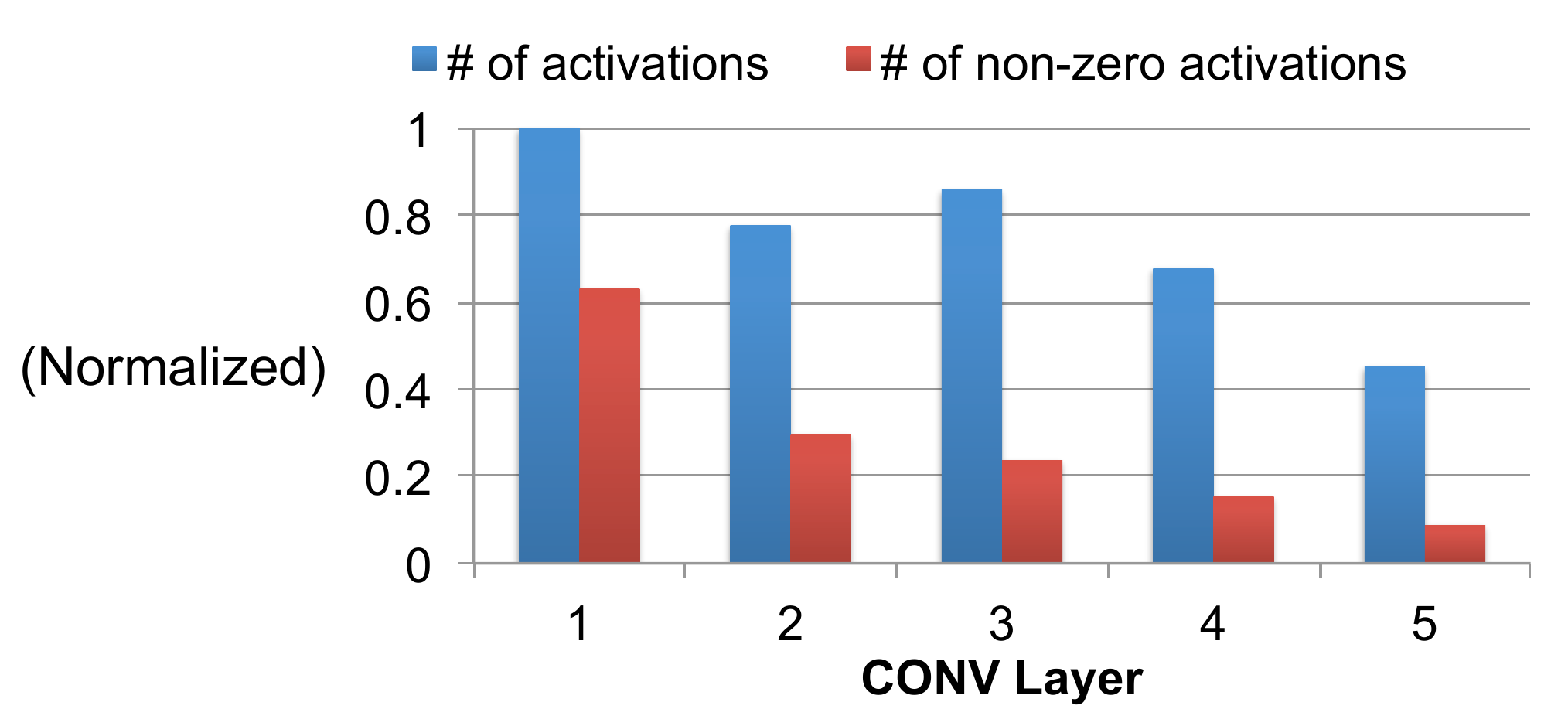}
				\label{fig:sparsity}
	}
}
\caption{Sparsity in activations due to ReLU.}
\label{fig:sparsity_relu}
\end{figure}

As discussed in Section~\ref{sssec:non-linearity}, ReLU is a popular form of non-linearity used in DNNs that sets all negative values to zero as shown in Fig.~\ref{fig:relu}.  As a result, the output activations of the feature maps after the ReLU are sparse; for instance, the feature maps in AlexNet have sparsity between 19\% to 63\% as shown in Fig.~\ref{fig:sparsity}.  This sparsity gives ReLU an implementation advantage over other non-linearities such as sigmoid, etc.  

The sparsity can be exploited for energy and area savings using compression, particularly for off-chip DRAM access which is expensive.  For instance, a simple run length coding that involves signaling non-zero values of 16-bits and then runs of zeros up to 31 can reduce the external memory bandwidth of the activations by 2.1$\times$ and the overall external bandwidth (including weights) by 1.5$\times$ ~\cite{jssc2017-chen}.\footnote{This simple run length compression is within 5-10\% of the theoretical entropy limit.} 
In addition to compression, the hardware can also be modified such that it skips reading the weights and performing the MAC for zero-valued activations to reduce energy cost by 45\%~\cite{isscc2016-chen}.  Rather than just gating the read and MAC computation, the hardware could also skip the cycle to increase the throughput by 1.37$\times$~\cite{albericio2016cnvlutin}.

The activations can be made to be even more sparse by pruning the low-valued activations.  For instance, if all activations with small values are pruned, this can be translated into an additional 11\% speed up~\cite{albericio2016cnvlutin} or 2$\times$ power reduction~\cite{reagen2016minerva} with little impact on accuracy. Aggressively pruning more activations can provide additional throughput improvement at a cost of reduced accuracy.

\subsubsection{Network Pruning}
\label{ssec:pruning}
To make network training easier, the networks are usually over-parameterized. Therefore, a large amount of the weights in a network are redundant and can be removed (i.e., set to zero). This process is called network pruning. Aggressive network pruning often requires some fine-tuning of the weights to maintain the original accuracy. This was first proposed in 1989 through a technique called Optimal Brain Damage~\cite{nips1990-lecun-opt_brain_damage}. The idea was to compute the impact of each weight on the training loss (discussed in Section~\ref{ssec:training}), referred to as the weight saliency. The low-saliency weights were removed and the remaining weights were fine-tuned; this process was repeated until the desired weight reduction and accuracy were reached. 

In 2015, a similar idea was applied to modern DNNs in~\cite{nips2015-han-learn_conn}. Rather than using the saliency as a metric, which is too difficult to compute for the large-scaled DNNs, the pruning was simply based on the magnitude of the weights. Small weights were pruned and the model was fine-tuned to restore the accuracy. Without fine-tuning the weights, about 50\% of the weights could be pruned. With fine-tuning, over 80\% of the weights were pruned. Overall this approach can reduce the number of weights in AlexNet by 9$\times$ and the number of MACs by 3$\times$. Most of the weight reduction comes from the fully-connected layers (9.9$\times$ for fully-connected layers versus 2.7$\times$ for convolutional layers).

%For AlexNet, 16\%-65\% of the weights are removed for the convolutional layer, while 75\%-91\% of the weights are removed for the fully connected layers. The impact of this is slightly higher in larger models such as VGG-16, where 42-78\% of the weights are pruned from the convolutional layer and 77-96\% of the weights are pruned from the fully connected layers. This shows that both AlexNet and VGG-16 are over-parametrized. Overall, the number of weights are reduced by 9$\times$ and the \vscomment{number of operations by 3.3$\times$}{check if they count only non-zero operations including activations}. However, most of the savings is in the fully connected layers, which as previously mentioned, has less impact on energy or throughput.  Still, this approach reduces the weights and computations of the convolutional layers by 2.7$\times$ and 3$\times$, respectively.  The reported savings on the fully connected layer is an average of 3.2$\times$ speed up on GPU, 3$\times$ on GPU and 5$\times$ on mobile GPU assuming a batch size of 1~\cite{han2016eie}. \vscomment{The impact of this speed up will likely reduce for larger batch sizes.}{discuss with Yu-Hsin, doesn't increasing batch size increase throughput of FC? if so, savings should be diminished}

\begin{figure}
    \begin{center}
        \includegraphics[width=0.9\linewidth]{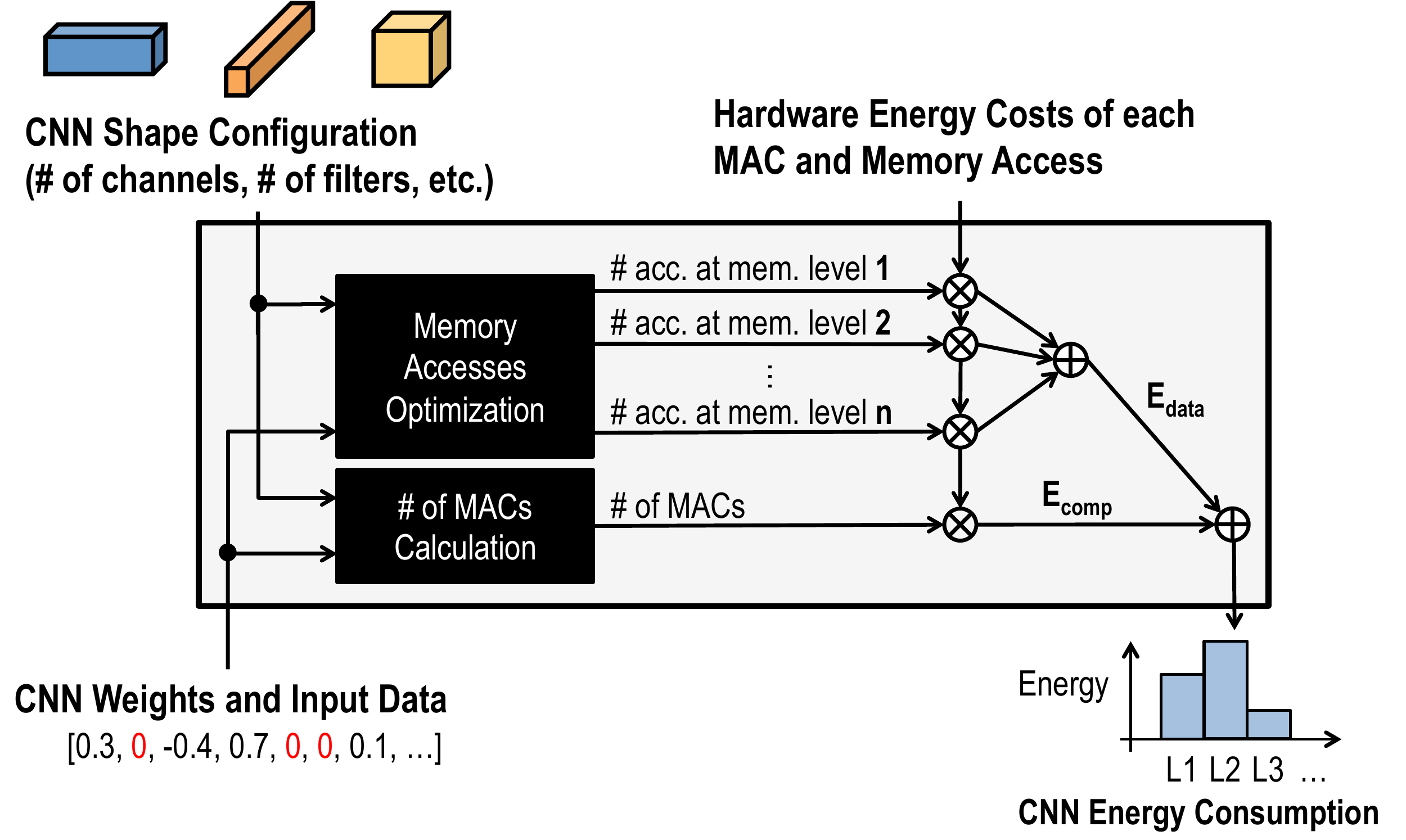}
        \caption{   Energy estimation methodology from~\cite{yang2016}, which estimates the energy based on data movement from different levels of the memory hierarchy, number of MACs, and data sparsity.
                }      
        \label{fig:energy_estimation}
    \end{center}
\end{figure}

\begin{figure}
\centering{	
    \subfigure[Energy versus accuracy trade-off of popular DNN models.]{
		\includegraphics[width=0.9\linewidth]{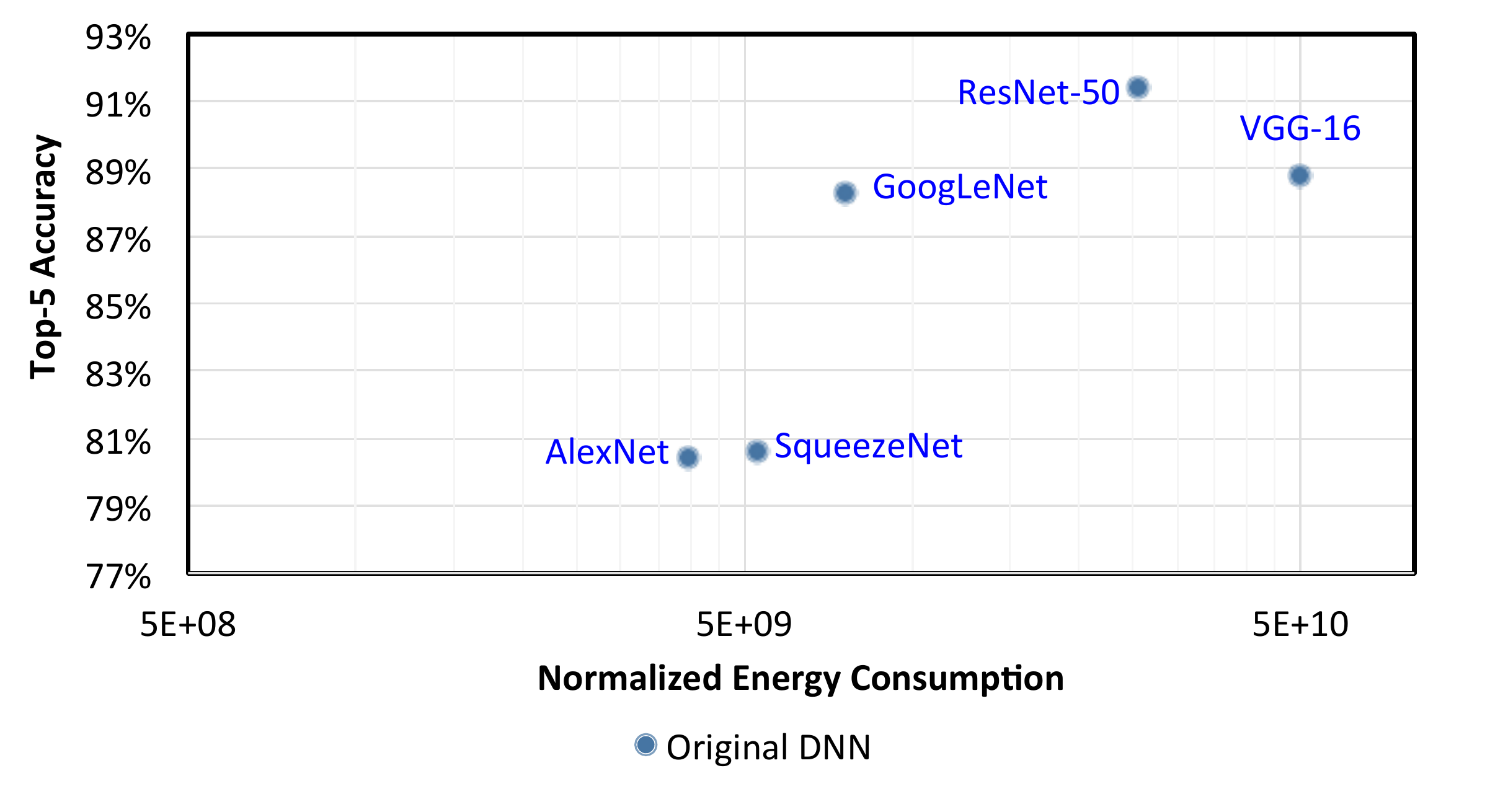}
		\label{fig:energy_survey_dense}
	}
	    \subfigure[Impact of energy-aware pruning.]{
		\includegraphics[width=0.9\linewidth]{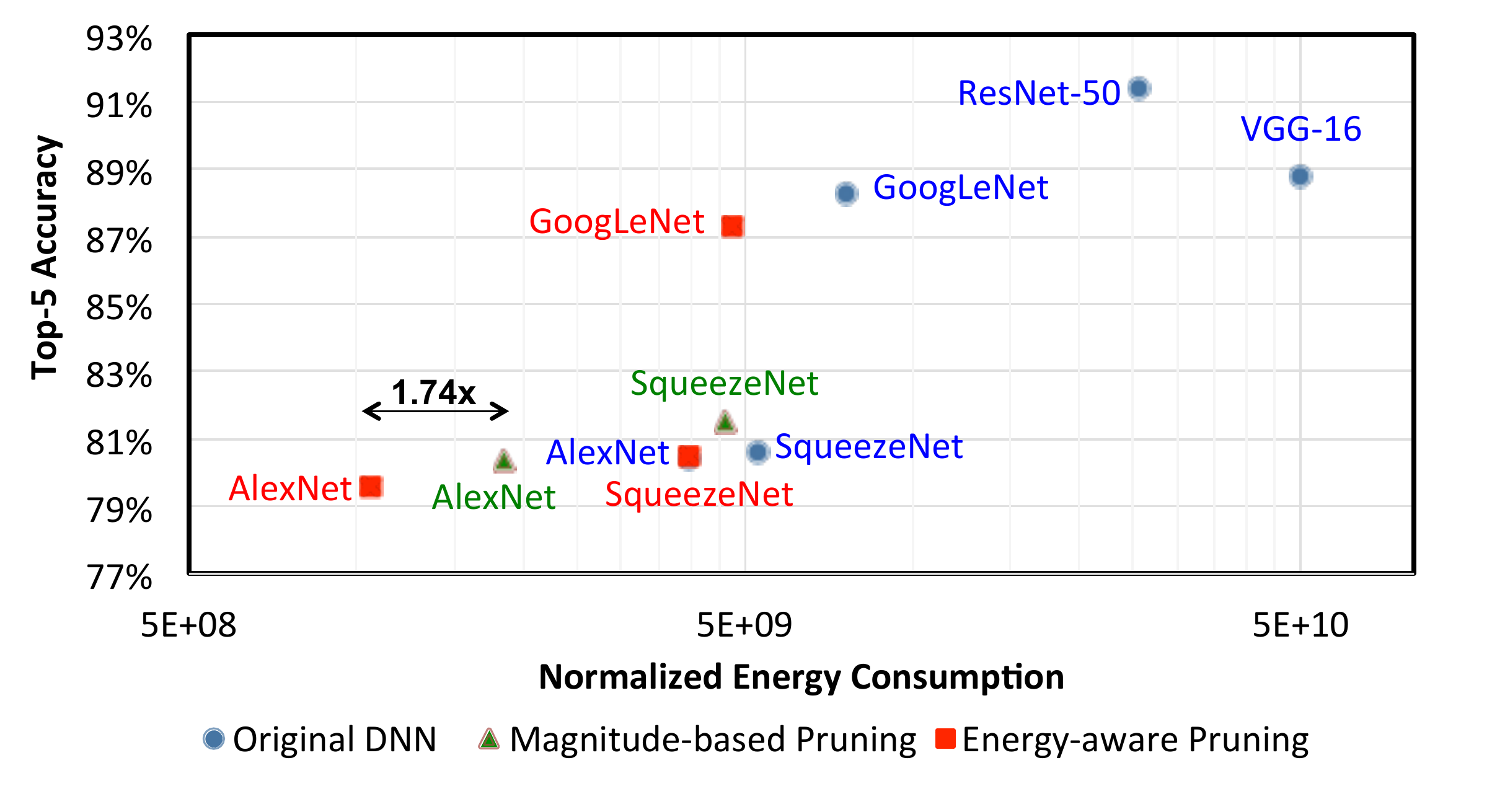}
		\label{fig:energy_survey}
	}
}
\caption{Energy values estimated with methodology in~\cite{yang2016}.}
\label{fig:estimated_energy}
\end{figure}

However, the number of weights alone is not a good metric for energy. For instance, in AlexNet, the number of weights in the fully-connected layers is much larger than in the convolutional layers; however, the energy of the convolutional layers is much higher than the fully-connected layers as shown in Fig.~\ref{fig:breakdown_across_layers}~\cite{isca2016-chen}. Rather than using the number of weights and MAC operations as proxies for energy, the pruning of the weights can be directly driven by energy itself~\cite{yang2016}. An energy evaluation method can be used to estimate the DNN energy that accounts for the data movement from different levels of the memory hierarchy, the number of MACs, and the data sparsity as shown in Fig.~\ref{fig:energy_estimation}; this energy estimation tool is available at~\cite{energy_estimation}. The resulting energy values for popular DNN models are shown in Fig.~\ref{fig:energy_survey_dense}. Energy-aware pruning can then be used to prune weights based on energy to reduce the overall energy across all layers by 3.7$\times$ for AlexNet, which is 1.74$\times$ more efficient than magnitude-based approaches~\cite{nips2015-han-learn_conn} as shown in Fig.~\ref{fig:energy_survey}.  As mentioned previously, it is well known that AlexNet is over-parameterized. The energy-aware pruning can also be applied to GoogleNet, which is already a small DNN model, for a 1.6$\times$ energy reduction. 

Recent works have examine how to efficiently support processing of sparse weights in hardware.  One area of interest is how to best store the sparse weights after pruning.  Similar to compressing the sparse activations discussed in Section~\ref{ssec:statistics}, the sparse weights can be compressed to reduce memory access bandwidth by 20 to 30\%~\cite{iclr2016-han-deep_comp}. 

\begin{figure}
\centering{
    \subfigure[Compressed sparse row (CSR) ]{
		\includegraphics[width=0.8\linewidth]{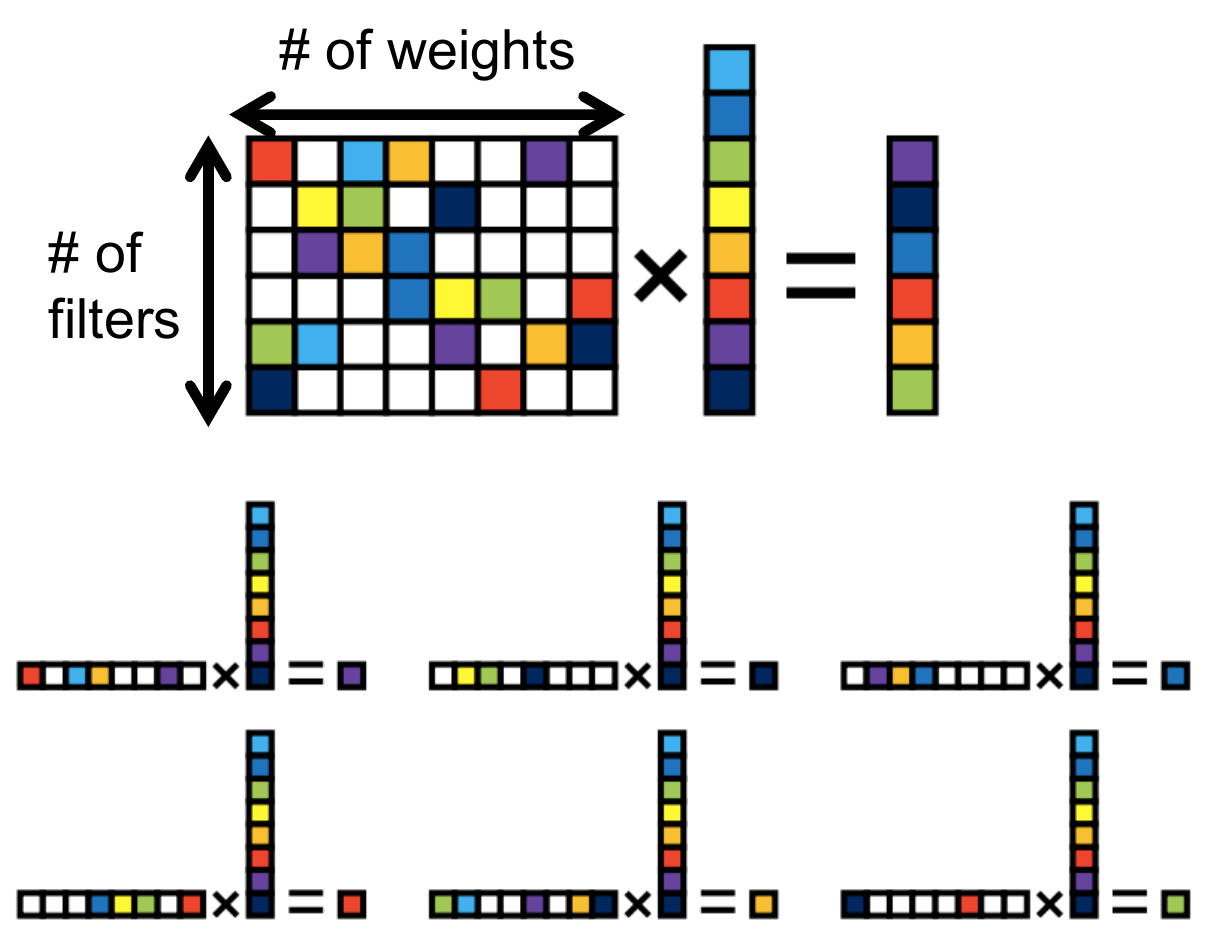}
		\label{fig:CSR}
	}	
    \subfigure[Compressed sparse column (CSC)]{
		\includegraphics[width=0.8\linewidth]{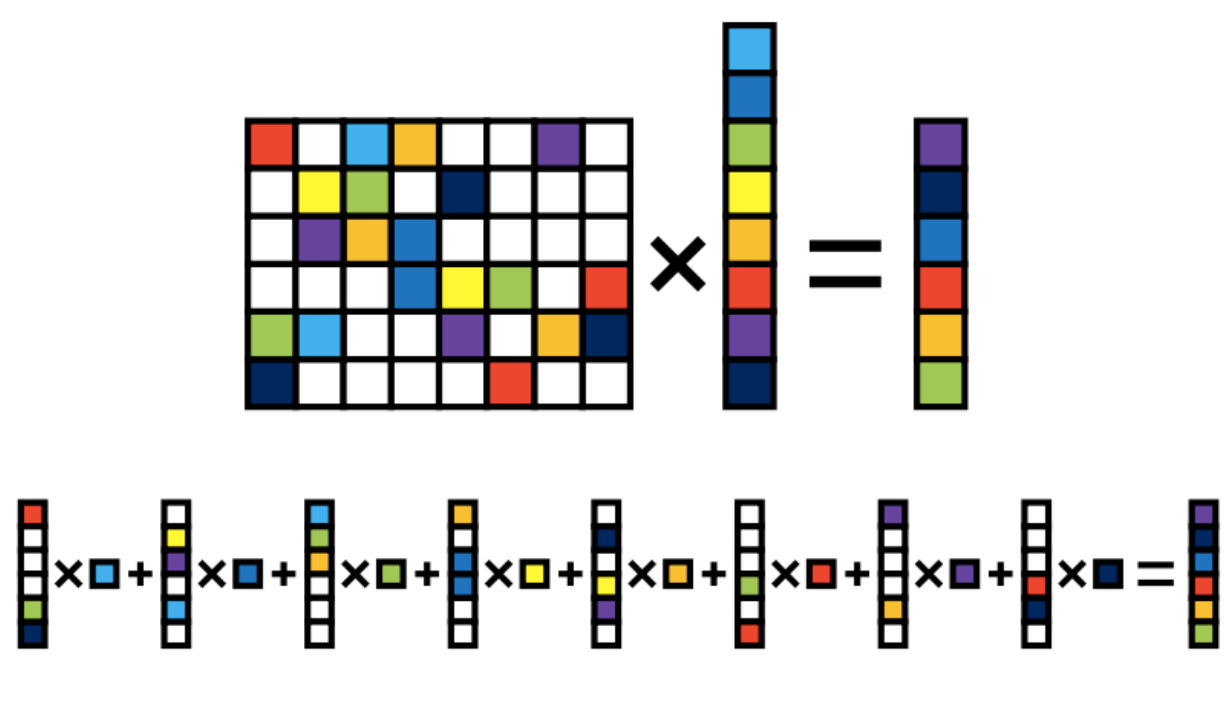}
				\label{fig:CSC}
	}
}
\caption{Sparse matrix-vector multiplications using different storage formats (Figure from~\cite{dorrance2014scalable}).}
\label{fig:sparse_matrix_vector}
\end{figure}

When DNN processing is performed as a matrix-vector multiplication, as shown in Fig.~\ref{fig:matrix_vec_FC}, one challenge is to determine how to store the sparse weight matrix in a compressed format. The compression can be applied either in row or column order. A compressed sparse row (CSR) format, as shown in Fig.~\ref{fig:CSR}, is often used to perform Sparse Matrix-Vector multiplication. However, the input vector needs to be read in multiple times even though only a subset of it is used since each row of the matrix is sparse.  Alternatively, a compressed sparse column (CSC) format, as shown in Fig.~\ref{fig:CSC}, can be used, where the output is updated several times, and only one element of the input vector is read at a time~\cite{dorrance2014scalable}.  The CSC format will provide an overall lower memory bandwidth than CSR if the output is smaller than the input, or in the case of DNN, if the number of filters is \emph{not} significantly larger than the number of weights in the filter ($C \times R \times S$ from Fig.~\ref{fig:DNN_conv}).  Since this is often true, CSC can be an effective format for sparse DNN processing.

Custom hardware has been explored to efficiently support pruned DNN models. Many works aim to perform the processing without decompressing the weights or activations.  EIE~\cite{han2016eie} performs the sparse matrix-vector multiplication specifically for the fully connected layers.  It stores the weights in a CSC format along with the start location of each column, which needs to be stored since the compressed weights have variable length.  When the input is not zero, the compressed weight column is read and the output is updated. To handle the sparsity, additional logic is used to keep track of the location of the output that should be updated. SCNN~\cite{parashar2017scnn} supports processing of convolutional layers in a compressed format. It uses an input stationary dataflow to deliver the compressed weights and activations to a multiplier array followed by a scatter network to add the scattered partial sums. 

Recent works have also explored the use of structured pruning to avoid the need for custom hardware~\cite{wen2016learning, anwar2017structured}.  Rather than pruning individual weights (also referred to as fine-grained pruning), structured pruning involves pruning groups of weights (also referred to as coarse-grained pruning). The benefits of structured pruning are (1) the resulting weights can better align with the data-parallel architecture (e.g., SIMD) found in existing general purpose hardware, which results in more efficient processing~\cite{yu2017scalpel}; (2) it amortizes the overhead cost required to signal the location of the non-zero weights across a group of weights, which improves compression and thus reduces storage cost. These groups of weights can include a pair of neighboring weights, an entire row or column of a filter, an entire channel of a filter or the entire filter itself; using larger groups tends to result in higher loss in accuracy~\cite{mao2017exploring}.

\subsubsection{Compact Network Architectures}
The number of weights and operations can also be reduced by improving the network architecture itself. The trend is to replace a large filter with a series of smaller filters, which have fewer weights in total; when the filters are applied sequentially, they achieve the same overall effective receptive field (i.e., the region the filter uses from input image to compute an output). This approach can be applied during the network architecture design (before training) or by decomposing the filters of a trained network (after training). The latter one avoids the hassle of training networks from scratch. However, it is less flexible than the former one. For example, existing methods can only decompose a filter in a trained network into a series of filters without non-linearity between them.

\paragraph{Before Training}
In recent DNN models, filters with a smaller width and height are used more frequently because concatenating several of them can emulate a larger filter as shown in Fig.~\ref{fig:decompose_filters}. For example, one 5$\times$5 convolution can be replaced with two 3$\times$3 convolutions. Alternatively, one N$\times$N convolution can be decomposed into two 1-D convolutions, one 1$\times$N and one N$\times$1 convolution~\cite{szegedy2016rethinking}; this basically imposes a restriction that the 2-D filter must be separable, which is a common constraint in image processing~\cite{lim1990two}.  Similarly, a 3-D convolution can be replaced by a set of 2-D convolutions (i.e., applied only on one of the input channels) followed by 1$\times$1 3-D convolutions as demonstrated in Xception~\cite{arXiv2016-chollet-xception} and MobileNets~\cite{howard2017mobilenets}. The order of the 2-D convolutions and 1$\times$1 3-D convolutions can be switched.

%However, in practice, this decomposition tends to only works well for layers whose feature maps have the width and height ranging from 12 to 20~\cite{szegedy2016rethinking}

1$\times$1 convolutional layers can also be used to reduce the number of channels in the output feature map for a given layer, which reduces the number of filter channels and thus computation cost for the filters in the next layer as demonstrated in~\cite{lin2013network, cvpr2015-szegedy, cvpr2016-he}; this is often referred to as a `bottleneck' as discussed in Section~\ref{sec:popular}.  For this purpose, the number of 1$\times$1 filters has to be less than the number of channels in the 1$\times$1 filter.  For example, 32 filters of 1$\times$1$\times$64 can transform an input with 64 channels to an output of 32 channels and reduce the number of filter channels in the next layer to 32. SqueezeNet uses many 1$\times$1 filters to aggressively reduce the number of weights~\cite{arxiv2016-iandola}.  It proposes a \emph{fire} module that first `squeezes' the network with 1$\times$1 convolution filters and then expands it with multiple 1$\times$1 and 3$\times$3 convolution filters.  It achieves an overall 50$\times$ reduction in number of weights compared to AlexNet, while maintaining the same accuracy.  It should be noted, however, that reducing the number of weights does not necessarily reduce energy; for instance, SqueezeNet consumes more energy than AlexNet, as shown in Fig.~\ref{fig:energy_survey_dense}.

\paragraph{After Training}
Tensor decomposition can be used to decompose filters in a trained network without impacting the accuracy. It treats weights in a layer as a 4-D tensor and breaks it into a combination of smaller tensors (i.e., several layers). Low-rank approximation can then be applied to further increase the compression rate at the cost of accuracy degradation, which can be restored by fine-tuning the weights. 

This approach is demonstrated using Canonical Polyadic (CP) decomposition, a high-order extension of singular value decomposition that can be solved by various methods, such as a greedy algorithm~\cite{nips2014-denton_exploiting_linear_structure} or a non-linear least-square method~\cite{iclr2015-lebedev-cp_decomposition}. Combining CP-decomposition with low-rank approximation achieves a 4.5$\times$ speed-up on CPUs~\cite{iclr2015-lebedev-cp_decomposition}. However, CP-decomposition cannot be computed in a numerically stable way when the dimension of the tensor, which represents the weights, is larger than two~\cite{iclr2015-lebedev-cp_decomposition}. To alleviate this problem, Tucker decomposition is adopted instead in~\cite{iclr2016-kim-low_power_mobile_app}.

\subsubsection{Knowledge Distillation}

Using a deep network or averaging the predictions of different models (i.e., ensemble) gives a better accuracy than using a single shallower network. However, the computational complexity is also higher. To get the best of both worlds, knowledge distillation transfers the knowledge learned by the complex model (teacher) to the simpler model (student). The student network can therefore achieve an accuracy that would be unachievable if it was directly trained with the same dataset~\cite{kdd2006-bucilua-model_compression, arXiv2013-ba-need_to_be_deep}. For example, ~\cite{nips2014-hinton-distill_knowledge} shows how using knowledge distillation can improve the speech recognition accuracy of a student net by 2\%, which is similar to the accuracy of a teacher net that is composed of an ensemble of 10 networks.

\begin{figure}
    \begin{center}
        \includegraphics[width=0.7\linewidth]{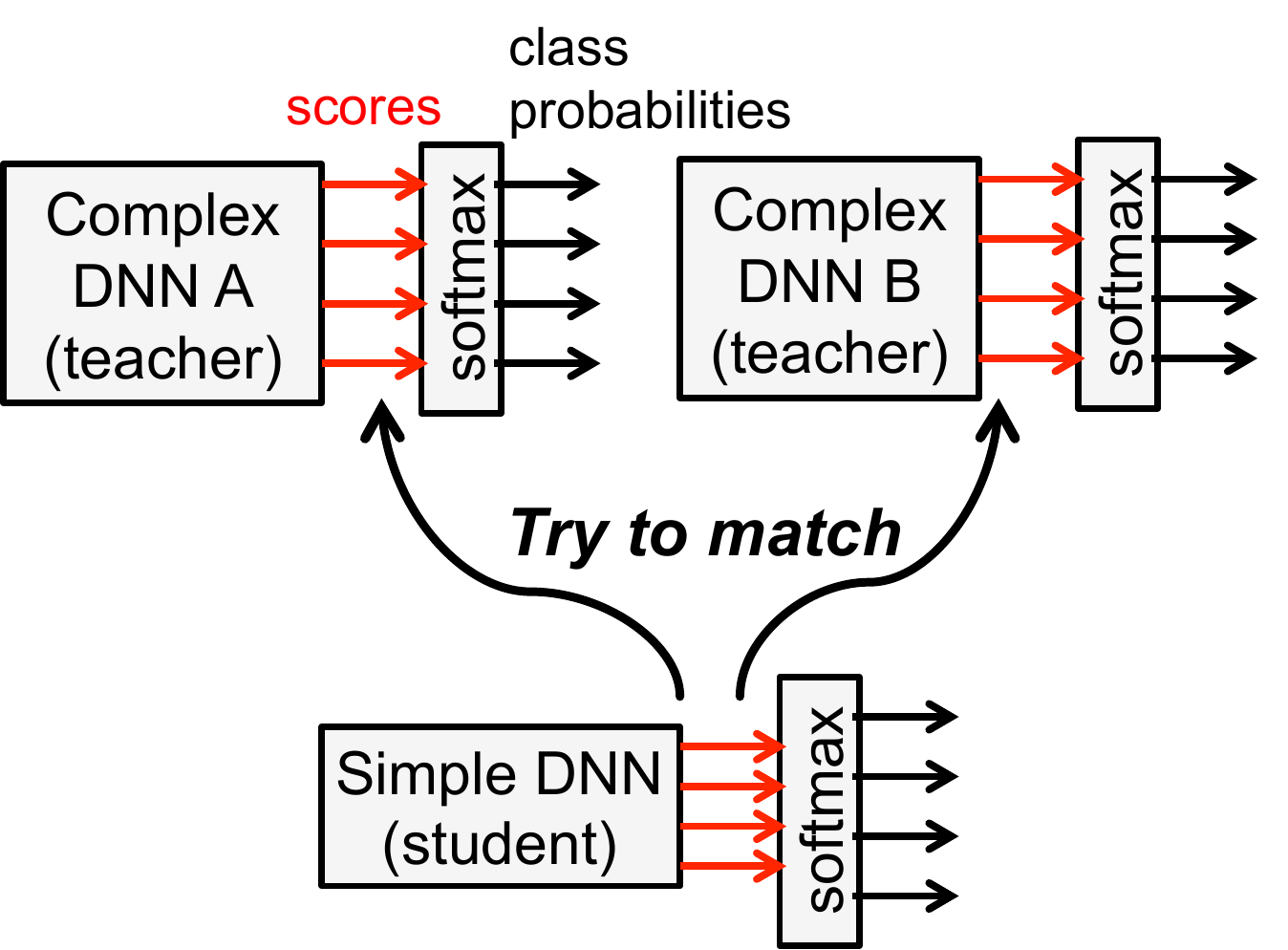}
        \caption{   Knowledge distillation matches the class scores of a small DNN to an ensemble of large DNNs.
                }      
        \label{fig:knowledge_distillation}
    \end{center}
\end{figure}

Fig.~\ref{fig:knowledge_distillation} shows the simplest knowledge distillation method~\cite{kdd2006-bucilua-model_compression}. The softmax layer is commonly used as the output layer in the image classification networks to generate the class probabilities from the class scores\footnote{Also commonly referred to as logits.}; it squashes the class scores into values between 0 and 1 that sum up to 1. For this knowledge distillation method, soft targets (values between 0 and 1) such as the class scores of the teacher DNN (or an ensemble of teacher DNNs) are used instead of the hard targets (values of either 0 or 1) such as the labels in the dataset.  The objective is to minimize the squared difference between the soft targets and the class scores of the student DNN. Class scores are used as the soft targets instead of the class probabilities because small values in the class scores contain important information that may be eliminated by the softmax. Alternatively, class probabilities after the softmax layer can be used as soft targets if the softmax is configured to generate softer class probabilities where the smaller values retain more information~\cite{nips2014-hinton-distill_knowledge}. Finally, the intermediate representations of the teacher DNN can also be incorporated as the extra hints to train the student DNN~\cite{iclr2015-romero-fitnets}.

%This method is a special case of using the outputs of a modified softmax layer (Eq.~\ref{eq:distill_knowledge_softmax}) as the soft targets~\cite{nips2014-hinton-distill_knowledge}. 
%\begin{equation}
%\label{eq:distill_knowledge_softmax}
%q_i = \frac{exp(z_i/T)}{\sum_{j}exp(z_j/T)},
%\end{equation}
%where $z_i$ is an input logit, $q_i$ is an output class probability of a softmax layer, and $T$ is the temperature. The higher the temperature is, the smoother the output class probabilities will be. For using the class probabilities to train the student net, the temperature of the final softmax is raised until the teacher net generates soft enough targets. Moreover, the intermediate representations of the teacher net can also be incorporated as the extra hints to train the student net~\cite{iclr2015-romero-fitnets}.

%Bucilua et al.~\cite{kdd2006-bucilua-model_compression} use the logits (i.e., the input to the final softmax layer) of the teacher net as the soft targets, and minimizes the squared difference between the soft targets and the logits of the student net. The reason of using the logits as the soft targets instead of the class probabilities (i.e., the output of the softmax layer) is that much of the information is in the small values in the soft targets~\cite{nips2014-hinton-distill_knowledge}, which becomes almost zero after applying softmax. Hinton et al.~\cite{nips2014-hinton-distill_knowledge} proposes directly using the class probabilities but raising the temperature of the final softmax until the teacher net generates soft enough targets.
%\begin{equation*}
%q_i = \frac{exp(z_i/T)}{\sum_{j}exp(z_j/T)},
%\end{equation*}
%where $z_i$ is an input logit, $q_i$ is an output class probability of a softmax layer, and $T$ is the temperature. The higher the temperature is, the smoother the output class probabilities will be. Extending this idea, Fitnets~\cite{iclr2015-romero-fitnets} takes the intermediate representations of the teacher net as the extra hints to train the student net.

%It is also feasible to train a large network (teacher) and then transfer the knowledge to a smaller network (student).

%% file: benchmarking.tex
\section{Benchmarking Metrics for DNN Evaluation and Comparison}
\label{sec:benchmarking}
As we have seen in this article, there has been a significant amount of research on efficient processing of DNNs.  We should consider several key metrics to compare the various strengths and weaknesses of different designs and proposed techniques.  These metrics should cover important attributes such as accuracy/robustness, power/energy consumption, throughput/latency and cost.  Reporting all these metrics is important in order to provide a complete picture of the trade-offs made by a proposed design or technique. We have prepared a website to collect these metrics from various publications~\cite{benchmarking}.

In terms of \emph{accuracy} and \emph{robustness}, it is important that the accuracy be reported on widely-accepted datasets as discussed in Section~\ref{sec:resources}. The difficulty of the dataset and/or task should be considered when measuring the accuracy.  For instance, the MNIST dataset for digit recognition is significantly easier than the ImageNet dataset. As a result, a DNN that performs well on MNIST may not necessarily perform well on ImageNet. Thus it is important that the same dataset and task is used when comparing the accuracy of different DNN models; currently ImageNet is preferred since it presents a challenge for DNNs, as opposed to MNIST, which can also be addressed with simple non-DNN techniques. To demonstrate primarily hardware innovations,  it would be desirable to report results for widely-used DNN models (e.g., AlexNet, GoogLeNet) whose accuracy and robustness have been well studied and tested. 

\emph{Energy} and \emph{power} are important when processing DNNs at the edge in embedded devices with limited battery capacity (e.g., smart phones, smart sensors, UAVs, and wearables), or in the cloud in data centers with stringent power ceilings due to cooling costs, respectively. Edge processing is preferred over the cloud for certain applications due to latency, privacy or communication bandwidth limitations.  When evaluating the power and energy consumption, it is important to account for all aspects of the system including the chip and external memory accesses. %In today's data centers, for each Watt that is consumed by the processors, an additional Watt is spent on cooling the processors.   

%\begin{itemize}
%    \item Latency:  The transmission delay to the cloud and back is not suitable for many real-time applications that require millisecond reaction time (e.g. navigation).
%\item Privacy: Local processing ensures that sensitive visual data is not shared, which is critical as cameras become more prevalent in our day-to-day lives. Imagine always-on vision for assisting the blind or as a user interface: a user might be ok if the phone is always watching as long as nobody else is watching as well.
%\item Limited bandwidth: In situations where limited connectivity is available, reliance on cloud compute would be harmful (e.g. search and rescue in disaster zones).
%\item Energy costs: The transmission of data itself requires significant energy. For instance, the energy cost for downlink over wireless networks (e.g. WiFi, LTE) is ~100nJ/bit. 
%\end{itemize}

\emph{High throughput} is necessary to deliver real-time performance for interactive applications such as navigation and robotics. For data analytics, high throughput means that more data can be analyzed in a given amount of time. As the amount of visual data is growing exponentially, high-throughput big data analytics becomes important, particularly if an action needs to be taken based on the analysis (e.g., security or terrorist prevention; medical diagnosis). 

\emph{Low latency} is necessary for real-time interactive applications.  Latency measures the time between when the pixel arrives to a system and when the result is generated.  Latency is measured in terms of seconds, while throughput is measured in operations/second. Often high throughput is obtained by batching multiple images/frames together for processing; this results in multiple frame latency (e.g., at 30 frames per second, a batch of 100 frames results in a 3 second delay). This delay is not acceptable for real-time applications, such as high-speed navigation where it would reduce the time available for course correction.  Thus achieving low latency and high throughput simultaneously can be a challenge.

\emph{Hardware cost} is in large part dictated by the amount of on-chip storage and the number of cores. Typical embedded processors have limited on-chip storage on the order of a few hundred kilobytes. Since there is a trade-off between the amount of on-chip memory and the external memory bandwidth, both metrics should be reported.  Similarly, there is a correlation between the number of cores and the throughput. In addition, while many cores can be built on a chip, the number of cores that can actually be used at a given time should be reported. It is often unrealistic to assume peak utilization and performance due to limitations of mapping and memory bandwidth.  Accordingly, the power and throughput should be reported for running actual DNNs as opposed to only reporting theoretical limits.

\subsection{Metrics for DNN Models}
To evaluate the properties of a given DNN model, we should consider the following metrics:
\begin{itemize}
    \item The \emph{accuracy} of the model in terms of the top-5 error on datasets such as ImageNet.  Also, the type of data augmentation used (e.g., multiple crops, ensemble models) should be reported.
    \item The \emph{network architecture} of the model should be reported, including number of layers, filter sizes, number of filters and number of channels.  
    \item The \emph{number of weights} impact the storage requirement of the model and should be reported.  If possible, the number of non-zero weights should be reported since this reflects the theoretical minimum storage requirements. 
    \item The \emph{number of MACs} that needs to be performed should be reported as it is somewhat indicative of the number of operations and potential throughput of the given DNN. If possible, the number of non-zero MACs should also be reported since this reflects the theoretical minimum compute requirements.  
\end{itemize}

Table~\ref{tab:sparse_dnns_NZ} shows how these metrics are reported for various well known DNNs. The accuracy is reported for the case where only a single crop for a single model is used for classification, such that the number of weights and MACs in the table are consistent.\footnote{Data augmentation is often used to increase accuracy.  This includes using multiple crops of an image to account for misalignment; in addition, an ensemble of multiple models can be used where each model has different weights due to different training settings, such as using different initializations or datasets, or even different network architectures. If multiple crops and models are used, then the number of MACs and weights required would increase.}  Note that accounting for the number of non-zero (NZ) operations significantly reduces the number of MACs and weights. Since the number of NZ MACs depends on the input data, we propose using the publicly available 50,000 validation images from ImageNet for the computation.  Finally, there are various methods to reduce the weights in a DNN (e.g., network pruning in Section~\ref{ssec:pruning}).  Table~\ref{tab:sparse_dnns_NZ} shows another example of these DNN model metrics, by comparing sparse DNNs pruned using~\cite{yang2016} to dense DNNs.

\begin{table}
\centering
\begin{tabular}{|c|c|c|c|c|}
\hline
\multirow{2}{*}{\textbf{Metrics}}  & \multicolumn{2}{|c|}{\textbf{AlexNet}} & \multicolumn{2}{|c|}{\textbf{GoogLeNet v1}}\\\cline{2-5}
&\textbf{dense}  & \textbf{sparse} &\textbf{dense}  & \textbf{sparse}\\\noalign{\hrule height 2pt}
\textbf{Top-5 error} & {19.6} & {20.4}& {11.7}& {12.7}\\\noalign{\hrule height 1.5pt}
\textbf{Number of CONV Layers} & {5} & {5}& {57}& {57}\\\hline
\begin{tabular}[c]{@{}c@{}}\textbf{Depth in}\\ \textbf{(Number of CONV Layers)}\end{tabular} & {5} & {5}& {21}& {21}\\\hline
\textbf{Filter Sizes} & \multicolumn{2}{|c|}{3,5,11} & \multicolumn{2}{|c|} {1,3,5,7}\\\hline
\textbf{Number of Channels} & \multicolumn{2}{|c|} {3-256} & \multicolumn{2}{|c|} {3-832}\\\hline
\textbf{Number of Filters} & \multicolumn{2}{|c|} {96-384} & \multicolumn{2}{|c|} {16-384}\\\hline
\textbf{Stride} & \multicolumn{2}{|c|} {1,4} & \multicolumn{2}{|c|} {1,2}\\\hline
\textbf{NZ Weights} & {2.3M} & {351k}& {6.0M}& {1.5M}\\\hline
\textbf{NZ MACs} & {395M} & {56.4M}& {806M}& {220M}\\\noalign{\hrule height 1.5pt}
\textbf{FC Layers} & {3} & {3}& {1}& {1}\\\hline
\textbf{Filter Sizes} & \multicolumn{2}{|c|}{1,6} & \multicolumn{2}{|c|} {1}\\\hline
\textbf{Number of Channels} & \multicolumn{2}{|c|} {256-4096} & \multicolumn{2}{|c|} {1024}\\\hline
\textbf{Number of Filters} & \multicolumn{2}{|c|} {1000-4096} & \multicolumn{2}{|c|} {1000}\\\hline
\textbf{NZ Weights} & {58.6M} & {5.4M}& {1M}& {870k}\\\hline
\textbf{NZ MACs} & {14.5M} & {1.9M}& {635k}& {663k}\\\noalign{\hrule height 1.5pt}
\textbf{Total NZ Weights} & {61M} & {5.7M}& {7M}& {2.4M}\\\hline
\textbf{Total NZ MACs} & {410M} & {58.3M}& {806M}& {221M}\\\hline
\end{tabular}
\caption{Metrics for Popular DNN Models. Sparsity is account for by reporting non-zero (NZ) weights and MACs.}
\label{tab:sparse_dnns_NZ}
\end{table}

\subsection{Metrics for DNN Hardware}
To measure the efficiency of the DNN hardware, we should consider the following additional metrics:
\begin{itemize}
    \item The \emph{power and energy} consumption of the design should be reported for various DNN models; the DNN model specifications should be provided including which layers and bit precision are supported by the hardware during measurement. In addition, the amount of off-chip accesses (e.g., DRAM accesses) should be included since it accounts for a significant portion of the system power; it can be reported in terms of the total amount of data that is read and written off-chip per inference. 
    \item The \emph{latency and throughput} should be reported in terms of the batch size and the actual run time for various DNN models, which accounts for mapping and memory bandwidth effects.  This provides a more useful and informative metric than peak throughput.    
    \item  The \emph{cost} of the chip depends on the area efficiency, which accounts for the size and type of memory (e.g., registers or SRAM) and the amount of control logic. It should be reported in terms of the core area in squared millimeters per multiplier along with process technology.    
\end{itemize}

In terms of cost, different platforms will have different implementation-specific metrics.  For instance, for an FPGA, the specific device should be reported, along with the utilization of resources such as DSP, BRAM, LUT and FF; performance density such as GOPs/slice can also be reported.

Each processor should report various specifications for each metric as shown in Table~\ref{tab:eyeriss_benchmark}, using the Eyeriss chip as an example.  It is important that all metrics and specifications are accounted for in order fairly evaluate all the design trade-offs. For instance, without the accuracy given for a specific dataset and task, one could run a simple DNN and easily claim low power, high throughput, and low cost -- however, the processor might not be usable for a meaningful task; alternatively, without reporting the off-chip bandwidth, one could build a processor with only multipliers and easily claim low cost, high throughput,  high accuracy, and low \emph{chip} power -- however, when evaluating \emph{system} power, the off-chip memory access would be substantial. Finally, the test setup should also be reported, including whether the results are measured or obtained from simulation\footnote{If obtained from simulation, it should be clarified whether it is from synthesis or post place-and-route and what library corner was used.} and how many images were tested.

%In addition, for each DNN model used to benchmark the processor, the following should be reported: the batch size (which affects off-chip access), the number of bits per operand, the energy per non-zero MAC, the off-chip access per non-zero MAC, the run time and the power consumption. 

\begin{table*}
\centering
\begin{tabular}{|l|l|c|c|}
\hline
\textbf{Metrics}  & \textbf{Specifications}  & \multicolumn{2}{|c|}{\textbf{Eyeriss}} \\\noalign{\hrule height 2pt}
\multirow{8}{*}{Cost}&\textbf{Process Technology} & \multicolumn{2}{|c|}{65nm LP TSMC (1.0V)}\\\cline{2-4}
&\textbf{Total Core Area (mm$^2$)} & \multicolumn{2}{|c|}{12.25}\\\cline{2-4}
&\textbf{Total On-chip Memory (kB)} & \multicolumn{2}{|c|}{192}\\\cline{2-4}
&\textbf{Number of Multipliers} & \multicolumn{2}{|c|}{168}\\\cline{2-4}
&\textbf{Core area} & \multicolumn{2}{|c|}{\multirow{2}{*}{0.073}} \\
&\textbf{per Multiplier (mm$^2$)} &  \multicolumn{2}{|c|}{}\\\cline{2-4}
&\textbf{On-chip Memory} & \multicolumn{2}{|c|}{\multirow{2}{*}{1.14}} \\
&\textbf{per Multiplier (kB)} &  \multicolumn{2}{|c|}{}\\\hline
\multirow{2}{*}{Test Setup} &\textbf{Measured or Simulated} & \multicolumn{2}{|c|}{Measured}\\\cline{2-4}
&\textbf{If Simulated, Syn or PnR} & \multicolumn{2}{|c|}{n/a}\\\noalign{\hrule height 1.5pt}
\multirow{6}{*}{Accuracy}&\textbf{DNN Model} &{AlexNet} & {VGG-16}\\\cline{2-4}
&\textbf{Top-5 error on ImageNet} &{19.8} & {8.8}\\\cline{2-4}
&\textbf{Dense/Sparse} & {Dense}& {Dense}\\\cline{2-4}
&\textbf{Supported Layers} & { All CONV layers }& {All CONV layers}\\\cline{2-4}
&\textbf{Bits per Weight} & {16}& {16}\\\cline{2-4}
&\textbf{Bits per Input Activation} & {16} & {16}\\\hline
\multirow{2}{*}{Latency and Throughput}&\textbf{Batch Size} & {4}& {3}\\\cline{2-4}
&\textbf{Run Time (msec)} & {115.3}& {4309.4}\\\hline
\multirow{3}{*}{Power and Energy}&\textbf{Power (mW)} & {278}& {236}\\\cline{2-4}
&\textbf{Off-chip Accesses} & {\multirow{2}{*}{3.85}}& {\multirow{2}{*}{107.03}}\\
&\textbf{per Image Inference (MBytes)} & &\\\hline
Test Setup & \textbf{Number of Images Tested} & {100}& {100}\\\hline

\end{tabular}
\caption{Example Benchmark Metrics for Eyeriss~\cite{isscc2016-chen}.}
\label{tab:eyeriss_benchmark}
\end{table*}

In summary, the evaluation process for whether a DNN system is a viable solution for a given application might go as follows: (1) the accuracy determines if it can perform the given task; (2) the latency and throughput determine if it can run fast enough and in real-time; (3) the energy and power consumption will primarily dictate the form factor of the device where the processing can operate; (4) the cost, which is primarily dictated by the chip area, determines how much one would pay for this solution.

%% file: summary.tex
\section{Summary} 

The use of deep neural networks (DNNs) has seen explosive growth in the past few years. They are currently widely used for many artificial intelligence (AI) applications including computer vision, speech recognition and robotics and are often delivering better than human accuracy.  However, while DNNs can deliver this outstanding accuracy, it comes at the cost of high computational complexity. Consequently, techniques that enable efficient processing of deep neural network to improve \emph{energy-efficiency} and \emph{throughput} without sacrificing \emph{accuracy} with cost-effective hardware are critical to expanding the deployment of DNNs in both existing and new domains.

Creating a system for efficient DNN processing should begin with understanding the current and future applications and the specific computations required both now and the potential evolution of those computations. This article surveys a number of the current applications, focusing on computer vision applications, the associated algorithms, and the data being used to drive the algorithms. These applications, algorithms and input data are experiencing rapid change. So extrapolating these trends to determine the degree of flexibility desired to handle next generation computations, becomes an important ingredient of any design project. 

During the design-space exploration process, it is critical to understand and balance the important system metrics. For DNN computation these include the accuracy, energy, throughput and hardware cost. Evaluating these metrics is, of course, key, so this article surveys the important components of a DNN workload. In specific, a DNN workload has two major components. First, the workload is the form of each DNN network including the `shape' of each layer and the interconnections between layers. These can vary both within and between applications. Second, the workload consists of the specific the data input to the DNN. This data will vary with the input set used for training or the data input during operation for inference.

This article also surveys a number of avenues that prior work have taken to optimize DNN processing. Since data movement dominates energy consumption, a primary focus of some recent research has been to reduce data movement while maintaining accuracy, throughput and cost.  This means selecting architectures with favorable memory hierarchies like a spatial array, and developing dataflows that increase data reuse at the low-cost levels of the memory hierarchy. We have included a taxonomy of dataflows and an analysis of their characteristics. Other work is presented that aims to save space and energy by changing the representation of data values in the DNN. Still other work saves energy and sometimes increases throughput by exploiting the sparsity of weights and/or activations. 

The DNN domain also affords an excellent opportunity for joint hardware/software co-design. For example, various efforts have noted that efficiency can be improved by increasing sparsity (increasing the number of zero values) or optimizing the representation of data by reducing the precision of values or using more complex mappings of the stored value to the actual value used for computation. However, to avoid losing accuracy it is often useful to modify the network or fine-tune the network's weights to accommodate these changes. Thus, this article both reviews a variety of these techniques and discusses the frameworks that are available for describing, running and training networks. 

Finally, DNNs afford the opportunity to use mixed-signal circuit design and advanced technologies to improve efficiency. These include using memristors for analog computation and 3-D stacked memory. Advanced technologies can also can facilitate moving computation closer to the source by embedding computation near or within the sensor and the memories.  Of course, all of these techniques should also be considered in combination, while being careful to understand their interactions and looking for opportunities for joint hardware/algorithm co-optimization. 

In conclusion, although much work has been done, deep neural networks remain an important area of research with many promising applications and opportunities for innovation at various levels of hardware design.

\section*{Acknowledgments}
Funding provided by DARPA YFA, MIT CICS, and gifts from Nvidia and Intel. The authors thank the anonymous reviewers as well as James Noraky, Mehul Tikekar and Zhengdong Zhang for providing valuable feedback on this paper.